\documentclass{clv3}\pdfoutput=1

\usepackage[T2A,T1]{fontenc}
\usepackage[utf8]{inputenc}
\usepackage[russian,english]{babel}

\usepackage{hyperref}

\usepackage{amsfonts,amsmath,amssymb,amsthm}

\usepackage{xcolor}
\newcommand\legend[1]{\fcolorbox{white}{#1}{\rule{0pt}{4pt}\rule{4pt}{0pt}}}
\definecolor{comment}{HTML}{8856a7}
\definecolor{ggplotsequential}{HTML}{b2abd2}
\definecolor{ggplotparallel}{HTML}{5e3c99}
\definecolor{ggplotones}{HTML}{b3cde3}
\definecolor{ggplotcount}{HTML}{8c96c6}
\definecolor{ggplotsim}{HTML}{8856a7}
\definecolor{ggplotverb}{HTML}{b2abd2}
\definecolor{ggplotsubject}{HTML}{e66101}
\definecolor{ggplotobject}{HTML}{fdb863}
\definecolor{ggplotframe}{HTML}{5e3c99}

\usepackage{algorithm}
\usepackage[noend]{algorithmic}

\renewcommand{\algorithmicrequire}{\textbf{Input:}}
\newcommand{\REQUIREP}{\item[\hphantom{\algorithmicrequire}]}

\newcommand{\watset}{\texorpdfstring{\textsc{Watset}}{Watset}}

\newcommand\sense[2]{\textit{#1}\ensuremath{^{#2}}}

\DeclareMathOperator{\tf}{\mathrm{tf}}
\DeclareMathOperator{\idf}{\mathrm{idf}}
\DeclareMathOperator{\tfidf}{\tf\!\textrm{--}\!\idf}

\DeclareMathOperator{\ctx}{\mathrm{ctx}}

\DeclareMathOperator{\senses}{\mathrm{senses}}

\DeclareMathOperator{\ssim}{\mathrm{sim}}
\DeclareMathOperator{\NN}{\mathrm{NN}}
\DeclareMathOperator{\nmpu}{\mathrm{nmPU}}
\DeclareMathOperator{\nipu}{\mathrm{niPU}}

\usepackage{multirow}
\usepackage{booktabs}
\title{Watset: Local-Global Graph Clustering with Applications in Sense and Frame Induction}

\runningtitle{Watset: Local-Global Graph Clustering with Applications}

\runningauthor{Ustalov et~al.}

\author{Dmitry Ustalov\thanks{B~6, 26, Mannheim, D-68159 Germany. E-mail: dmitry@informatik.uni-mannheim.de.}}
\affil{University of Mannheim}

\author{Alexander Panchenko\thanks{Vogt-K\"olln-Stra{\ss}e, 30, Hamburg, D-22527 Germany. E-mail: panchenko@informatik.uni-hamburg.de. 
This work was primarily done while the author was with University of Hamburg.
}}
\affil{University of Hamburg,
Skolkovo Institute of Science and Technology}

\author{Chris Biemann}
\affil{University of Hamburg}

\author{Simone Paolo Ponzetto}
\affil{University of Mannheim}

\historydates{Submission received: 20 August 2018; revised version received: 22 February 2019; accepted for publication: 15 March 2019.}

\begin{document}

\maketitle

\begin{abstract}
We present a detailed theoretical and computational analysis of the Watset meta-algorithm for fuzzy graph clustering, which has been found to be widely applicable in a variety of domains. This algorithm creates an intermediate representation of the input graph that reflects the ``ambiguity'' of its nodes. Then, it uses hard clustering to discover clusters in this ``disambiguated'' intermediate graph. After outlining the approach and analyzing its computational complexity, we demonstrate that Watset shows competitive results in three applications: unsupervised synset induction from a synonymy graph, unsupervised semantic frame induction from dependency triples, and unsupervised semantic class induction from a distributional thesaurus. Our algorithm is generic and can be also applied to other networks of linguistic data. 
\end{abstract}

\section{Introduction}

Language can be conceived as a system of interrelated symbols, such as words, senses, part-of-speeches, letters, etc. Ambiguity is a fundamental inherent property of language. Namely, each symbol can refer to several meanings mapping the space of objects to the space of communicative signs~\citep{deSaussure:16}. For language processing applications, these symbols need to be represented in a computational format. The structure discovery paradigm~\citep{Biemann:12} aims at \textit{inducing a system of linguistic symbols and relationships between them} in an unsupervised way to enable processing of a wide variety of languages. Clustering algorithms are central and ubiquitous tools for such kinds of unsupervised structure discovery processes applied to natural language data. In this article, we present a new clustering algorithm,\footnote{This article builds upon and expands on \citet{Ustalov:17:watset} and \citet{Ustalov:18:triframes}.} which is especially suitable for processing of graphs of linguistic data, since it performs disambiguation of symbols in the local context in order to subsequently globally cluster those disambiguated symbols.

At the heart of our method lies the pre-processing of a graph on the basis of local pre-clustering. Breaking nodes that connect to several communities, a.k.a. hubs, into several local senses, helps to better reach the goal of clustering, no matter which clustering algorithm is used. This results in a sparser sense-aware graphical representation of the input data. Such a representation allows the use of efficient hard clustering algorithms for performing fuzzy clustering.

The contribution presented in this article is four-fold:
\begin{enumerate}
  \item A \textbf{meta-algorithm for graph clustering}, called {\watset}, performing a fuzzy clustering of the input graph using hard clustering methods in two subsequent steps (Section~\ref{sec:watset}).
  \item A \textbf{method for synset induction} based on the {\watset} algorithm applied to synonymy graphs weighted by word embeddings (Section~\ref{sec:synsets}).
  \item A \textbf{method for semantic frame induction} based on the {\watset} algorithm applied as a triclustering algorithm to syntactic triples (Section~\ref{sec:triframes}).
  \item A \textbf{method for semantic class induction} based on the {\watset} algorithm applied to a distributional thesaurus (Section~\ref{sec:classes}).
\end{enumerate}

The article is organized as follows. Section~\ref{sec:related} discusses the related work. Section~\ref{sec:watset} presents the {\watset} algorithm in a more general fashion than previously introduced by \citet{Ustalov:17:watset}, including an analysis of its computational complexity and run-time. We also describe a simplified version of {\watset} that does not use the context similarity measure for propagating links in the original graph to the appropriate senses in the disambiguated graph. Three subsequent sections present different applications of the algorithm. Section~\ref{sec:synsets} applies {\watset} for unsupervised synset induction, referencing results by \citet{Ustalov:17:watset}. Section~\ref{sec:triframes} shows frame induction with {\watset} on the basis of a triclustering approach, as previously described by \citet{Ustalov:18:triframes}. Section~\ref{sec:classes} presents new experiments on semantic class induction with {\watset}. Section~\ref{sec:conclusion} concludes with the final remarks and pointers for future work.

Table~\ref{tab:structures} shows several examples of linguistic structures on which we conduct experiments described in this article. With the exception of the type of input graph and the hyper-parameters of the {\watset} algorithm, the overall pipeline remains similar in every described application. For instance, in Section~\ref{sec:synsets} the input of the clustering algorithm is a graph of ambiguous synonyms and the output is an induced linguistic structure that represents synsets. Thus, varying the input graphs we show how using the same methodology various types of linguistic structures can be induced in an unsupervised manner. This opens avenues for extraction of various meaningful structures from linguistic graphs in natural language processing (NLP) and other fields using the method presented in this article.

\begin{table}[t]
\centering\hyphenpenalty=10000
\caption{\label{tab:structures}Various types of input linguistic graphs clustered by the {\watset} algorithm and the corresponding induced output symbolic linguistic structures.}
\begin{tabular}{p{31mm}p{33mm}p{45mm}c}\toprule
\textbf{Input Nodes} & \textbf{Input Edges} & \textbf{Output Linguistic Structure} & \textbf{See} \\\midrule
Polysemous words & Synonymy relationships\raggedright & Synsets composed of disambiguated words\raggedright & \S~\ref{sec:synsets} \\\hline
Subject-Verb-Object (SVO) triples & Most distributionally similar SVO triples\raggedright & Lexical semantic frames\raggedright & \S~\ref{sec:triframes} \\\hline
Polysemous words & Most distributionally similar words\raggedright & Semantic classes composed of disambiguated words\raggedright & \S~\ref{sec:classes} \\\bottomrule
\end{tabular}
\end{table}

\section{Related Work}\label{sec:related}

We present surveys on graph clustering (Section~\ref{sub:clustering}), word sense induction (Section~\ref{sub:wsi}), lexical semantic frame induction (Section~\ref{sub:frames}), and semantic class induction (Section~\ref{sub:classes}), giving detailed explanations of algorithms used in our experiments and discussing related work on these topics.

\subsection{Graph Clustering}\label{sub:clustering}

Graph clustering is a process of finding groups of strongly related vertices in a graph, which is a field of research on its own with a large number of proposed approaches, see~\citet{Schaeffer:07} for a survey. Graph clustering methods are strongly related to the methods for finding communities in networks~\citep{Newman:04,Fortunato:10}. In our work, we focus mostly on the algorithms, which have proven to be useful for processing of networks of \textit{linguistic data}, such as word co-occurrence graphs, especially those that were used for induction of linguistic structures such as word senses.

\textbf{Markov Clustering}~\citep{vanDongen:00}, a.k.a. MCL, is a \textit{hard} clustering algorithm, i.e., a method which partions nodes of the graph in a set of disjoint clusters. This method is  based on simulation of stochastic flow in graphs. MCL simulates random walks within a graph by the alternation of two operators called expansion and inflation, which recompute the class labels. Notably, it has been successfully used for the word sense induction task~\citep{Dorow:03}.

\textbf{Chinese Whispers}~\citep{Biemann:06,Biemann:12}, a.k.a. CW, is a \textit{hard} clustering algorithm for weighted graphs that can be considered as a special case of MCL with a simplified class update step. At each iteration, the labels of all the nodes are updated according to the majority labels among the neighboring nodes. The algorithm has a hyper-parameter that controls graph weights that can be set to three values: (1) CW\textsubscript{top} sums over the neighborhood's classes; (2) CW\textsubscript{lin} downgrades the influence of a neighboring node by its degree; or (3) CW\textsubscript{log} by the logarithm of its degree.

\textbf{MaxMax}~\citep{Hope:13:maxmax} is a \textit{fuzzy} clustering algorithm particularly designed for the word sense induction task. In a nutshell, pairs of nodes are grouped if they have a maximal mutual affinity. The algorithm starts by converting the undirected input graph into a directed graph by keeping the maximal affinity nodes of each node. Next, all nodes are marked as root nodes. Finally, for each root node, the following procedure is repeated: all transitive children of this root form a cluster and the root are marked as non-root nodes; a root node together with all its transitive children form a fuzzy cluster.

\textbf{Clique Percolation Method} (CPM) by \citet{Palla:05} is a \textit{fuzzy} clustering algorithm, i.e., a method that partitions nodes of a graph in a set of potentially overlapping clusters. The method is designed for unweighted graphs and builds up clusters from $k$-cliques corresponding to fully connected sub-graphs of $k$ nodes. While this method is only commonly used in social network analysis for clique detection, we decided to add it to the comparison as synsets are essentially cliques of synonyms.

\textbf{Louvain} method~\citep{Blondel:08} is a \textit{hard} graph clustering method developed for identification of communities in large networks. The algorithm finds hierarchies of clusters in a recursive fashion. It is based on a greedy method that optimizes modularity of a partition of the network. First, it looks for small communities by optimizing modularity locally. Second, it aggregates nodes belonging to the same community and builds a new network whose nodes are the communities. These steps are repeated to maximize modularity of the clustering result.

\subsection{\label{sub:wsi}Word Sense Induction}

Word Sense Induction is an unsupervised knowledge-free approach to Word Sense Disambiguation (WSD): it uses neither handcrafted lexical resources nor hand-annotated sense-labeled corpora. Instead, it induces word sense inventories automatically from corpora. Unsupervised WSD methods fall into two main categories: context clustering and word ego network clustering.

\textbf{Context clustering approaches}, such as~\citet{Pedersen:97,Schutze:98}, represent an instance usually by a vector that characterizes its context, where the definition of context can vary greatly. These vectors of each instance are then clustered.

\citet{Schutze:98} induced sparse sense vectors by clustering context vectors using the expectation-maximization (EM) algorithm. This approach is fitted with a similarity-based WSD mechanism. \citet{Pantel:02} used a two-staged \textit{Clustering by Committee} algorithm. In a first stage, it uses average-link clustering to find small and tight clusters which are used to iteratively identify committees from these clusters. \citet{Reisinger:10} presented a multi-prototype vector space. Sparse $\tfidf$ vectors are clustered using a parametric method fixing the same number of senses for all words. Sense vectors are centroids of the clusters.

While most dense word vector models represent a word with a single vector and thus conflate senses~\citep{Mikolov:13,Pennington:14}, there are several approaches that produce word sense embeddings. Multi-prototype extensions of the Skip-Gram model~\citep{Mikolov:13} that use no predefined sense inventory learn one embedding word vector per one word sense and are commonly fitted with a disambiguation mechanism~\citep{Huang:12,Apidianaki:14,Tian:14,Neelakantan:14,Bartunov:16,Li:15,Cocos:16,Pelevina:16,Thomason:17}.

\citet{Huang:12} introduced multiple word prototypes for dense vector representations (embeddings). Their approach is based on a  neural network architecture; during training, all contexts of the word are clustered. 

\citet{Apidianaki:14} use an aligned parallel corpus and WordNet for English to perform cross-lingual word sense disambiguation to produce French synsets. However, \citet{Cocos:16} showed that it is possible to successfully perform a monolingual word sense induction using only such a paraphrase corpus as PPDB \citep{Pavlick:15}.

\citet{Tian:14} introduced a probabilistic extension of the Skip-Gram model~\citep{Mikolov:13} that learns multiple sense-aware prototypes weighted by their prior probability. These models use parametric clustering algorithms that produce a fixed number of senses per word. 

\citet{Neelakantan:14} proposed a multi-sense extension of the Skip-Gram model that was the first one to learn the number of senses by itself. During training, a new sense vector is allocated if the current context's similarity to existing senses is below some threshold. All mentioned above sense embeddings were evaluated on the contextual word similarity task, each one improving upon previous models. 

\citet{NietoPina:15} presented another multi-prototype modification of the Skip-Gram model. Their approach outperforms that of \citet{Neelakantan:14}, but requires the number of senses for each word to be set manually.

\citet{Bartunov:16} introduced AdaGram, a non-parametric method for learning sense embeddings based on a Bayesian extension of the Skip-Gram model. The granularity of learned sense embeddings is controlled by the $\alpha$ parameter.

\citet{Li:15} proposed an approach for learning of sense embeddings based on the Chinese Restaurant Process. A new sense is allocated if a new word context is significantly different from existing senses. The approach was tested on multiple NLP tasks, showing that sense embeddings can significantly improve the performance of part-of-speech tagging, semantic relationship identification and semantic relatedness tasks, but yield no improvement for named entity recognition and sentiment analysis. 

\citet{Thomason:17} performed multi-modal word sense induction by combining both language and vision signals. In this approach, word embeddings are learned from the ImageNet corpus~\citep{Deng:09} and visual features are obtained from a deep neural network. Running a $k$-Means algorithm on the joint feature set produces WordNet-like synsets.

\textbf{Word ego network clustering methods} cluster graphs of words semantically related to the ambiguous word~\citep{Lin:98,Pantel:02,Widdows:02,Biemann:06,Hope:13:maxmax}. An ego network consists of a single node (ego) together with the nodes they are connected to (alters) and all the edges among those alters~\citep{Everett:05}. In our case, such a network is a local neighborhood of one word. Nodes of the ego network can be (1) words semantically similar to the target word, as in our approach, or (2) context words relevant to the target, as in the \textit{UoS} system ~\citep{Hope:13:uos}. Graph edges represent semantic relationships between words derived using corpus-based methods (e.g., distributional semantics) or gathered from dictionaries. The sense induction process using word graphs is explored by \citet{Widdows:02,Biemann:06,Hope:13:maxmax}. Disambiguation of instances is performed by assigning the sense with the highest overlap between the instance's context words and the words of the sense cluster. \citet{Veronis:04} compiles a corpus with contexts of polysemous nouns using a search engine. A word graph is built by drawing edges between co-occurring words in the gathered corpus, where edges below a certain similarity threshold were discarded. His HyperLex algorithm detects hubs of this graph, which are interpreted as word senses. Disambiguation in this experiment is performed by computing the distance between context words and hubs in this graph.

\citet{DiMarco:13} presents a comprehensive study of several graph-based WSI methods including Chinese Whispers, HyperLex, and curvature clustering~\citep{Dorow:04}. Besides, the authors propose two novel algorithms: Balanced Maximum Spanning Tree Clustering and Squares (B-MST), Triangles and Diamonds (SquaT++). To construct graphs, authors use first-order and second-order relationships extracted from a background corpus as well as keywords from snippets. This research goes beyond intrinsic evaluations of induced senses and measures impact of the WSI in the context of an information retrieval via clustering and diversifying Web search results. Depending on the dataset, HyperLex, B-MST or Chinese Whispers provided the best results. For a comparative study of graph clustering algorithms for word sense induction in a pseudo-word evaluation confirming the effectiveness of CW, see \citet{Cecchini:18}.

\textbf{Methods based on clustering of synonyms}, such as our approach and MaxMax~\citep{Hope:13:maxmax}, induce the resource from an ambiguous graph of synonyms where edges a extracted from manually-created resources. To the best of our knowledge, most experiments either employed graph-based word sense induction applied to text-derived graphs or relied on a linking-based method that already assumes the availability of a WordNet-like resource. A notable exception is the ECO (Extraction, Clustering, Ontologisation) approach by~\citet{GoncaloOliveira:14}, which was applied to induce a WordNet of the Portuguese language called Onto.PT.\footnote{\url{http://ontopt.dei.uc.pt}} ECO is a \textit{fuzzy} clustering algorithm that was used to induce synsets for a Portuguese WordNet from several available synonymy dictionaries. The algorithm starts by adding random noise to edge weights. Then, the approach applies Markov Clustering (Section~\ref{sub:clustering}) of this graph several times to estimate the probability of each word pair being in the same synset. Finally, candidate pairs over a certain threshold are added to output synsets. We compare to this approach and to five other state-of-the-art graph clustering algorithms described in Section~\ref{sub:clustering} as the baselines.

\subsection{\label{sub:frames}Semantic Frame Induction}

Frame Semantics was originally introduced by \citet{Fillmore:82} and further developed in the FrameNet project \citep{Baker:98}. FrameNet is a lexical resource composed of a collection of semantic frames, relationships between them and a corpus of frame occurrences in text. This annotated corpus gave rise to the development of frame parsers using supervised learning \cite[\emph{inter alia}]{Gildea:02,Erk:06,Das:14}, as well as its application to a wide range of tasks, ranging from  answer extraction in Question Answering \citep{Shen:07} and Textual Entailment \citep{Burchardt:09,BenAharon:10}.

However, frame-semantic resources are arguably expensive and time-consuming to build due to difficulties in defining the frames, their granularity and domain, as well as the complexity of the construction and annotation tasks. Consequently, such resources exist only for a few languages~\citep{Boas:09} and even English is lacking domain-specific frame-based resources. Possible inroads are cross-lingual semantic annotation transfer \citep{Pado:09,Hartmann:16} or linking FrameNet to other lexical-semantic or ontological resources \citep[\emph{inter alia}]{Narayanan:03,Tonelli:09,Laparra:10,Gurevych:12}. One inroad for overcoming these issues  is automatizing the process of FrameNet construction through unsupervised frame induction techniques, as investigated by the systems described below.

\textbf{LDA-Frames}~\citep{Materna:12,Materna:13} is an approach to inducing semantic frames using a latent Dirichlet allocation (LDA) by \citet{Blei:03} for generating semantic frames and their respective frame-specific semantic roles at the same time. The authors evaluated their approach against the CPA corpus~\citep{Hanks:05}. Although \citet{Ritter:10} have applied LDA for inducing structures similar to frames, their study is focused on the extraction of mutually-related frame arguments.

\textbf{ProFinder}~\citep{Cheung:13} is another generative approach that also models both frames and roles as latent topics. The evaluation was performed on the in-domain information extraction task MUC-4~\citep{Sundheim:92} and on the text summarization task TAC-2010.\footnote{\url{https://tac.nist.gov/2010/Summarization}}

\citet{Modi:12} build on top of an unsupervised semantic role labeling model~\citep{Titov:12}. The raw text of sentences from the FrameNet data is used for training. The FrameNet gold annotations are then used to evaluate the labeling of the obtained frames and roles, effectively clustering instances known during induction.

\citet{Kawahara:14} harvest a huge collection of verbal predicates along with their argument instances and then apply the Chinese Restaurant Process clustering algorithm to group predicates with similar arguments. The approach was evaluated on the verb cluster dataset of \citet{Korhonen:03}.

These and some other related approaches, e.g., the one by \citet{OConnor:13}, were all evaluated in completely different incomparable settings, and used different input corpora, making it difficult to judge their relative performance.

\subsection{\label{sub:classes}Semantic Class Induction}

The problem of inducing semantic classes from text, also known as semantic lexicon induction, has been also extensively explored in previous works. This is because inducing semantic classes directly from text has the potential to avoid the limited coverage problems of knowledge bases like Freebase, DBpedia~\citep{Bizer:09} or BabelNet~\citep{Navigli:12:babelnet} that rely on Wikipedia~\citep{Hovy:13}, as well as to allow for resource induction across domains~\citep{Hovy:11}. Information about semantic classes, in turn, has been shown to benefit such high-level NLP tasks as coreference~\citep{Ng:07}.

Induction of semantic classes as a research direction in field of NLP starts, to the best of our knowledge, with \citet{Lin:01}, where sets of similar words are clustered into concepts. While this approach performs a hard clustering and does not label clusters, these drawbacks are addressed by \citet{Pantel:02}, where words can belong to several clusters, thus representing senses.

\citet{Pantel:04} aggregate hypernyms per cluster, which come from Hearst ~(\citeyear{Hearst:92}) patterns. Pattern-based approaches were further developed using graph-based methods using a PageRank-based weighting~\citep{Kozareva:08}, random walks~\citep{Talukdar:08}, or heuristic scoring~\citep{Qadir:15}. Other approaches use probabilistic graphical models, such as the ones proposed by \citet{Ritter:10} and \citet{Hovy:11}. To ensure the overall quality of extraction pattern with minimal supervision, \citet{Thelen:02} explored a bootstrapping approach, later extended by \citet{McIntosh:09} with bagging and distributional similarity to minimise the semantic drift problem of iterative bootstrapping algorithms.

As an alternative to pattern-based methods, \citet{Panchenko:18:mangosteen} show how to apply semantic classes to improve hypernymy extraction and taxonomy induction. Like in our experiments in Section~\ref{sec:classes}, it uses a distributional thesaurus as input, as well as multiple pre- and post-processing stages to filter the input graph and disambiguate individual nodes. In contrast to \citet{Panchenko:18:mangosteen}, here we directly apply the {\watset} algorithm to obtain the resulting distributional semantic classes instead of using a sophisticated parametric pipeline that performs a sequence of clustering and pruning steps.

Another related strain of research to semantic class induction is dedicated to the automatic \textit{set expansion} task~\citep{Sarmento:07,Wang:08,Pantel:09,Rong:16,Shen:17}. In this task, a set of input lexical entries, such as words or entities, is provided, e.g., ``apple, mango, pear, banana''. The system is expected to extend this initial set with relevant entries, such as other fruits in this case, e.g., ``peach'' and ``lemon''. Beside the academic publications listed above, Google Sets was an industrial system for providing similar functionality.\footnote{\url{http://web.archive.org/web/20110327090414/http://labs.google.com/sets}}

\section{\label{sec:watset}\watset, an Algorithm for Fuzzy Graph Clustering}

In this section, we present {\watset}, a meta-algorithm for fuzzy graph clustering. Given a graph connecting potentially ambiguous objects, e.g., words, {\watset} induces a set of unambiguous overlapping clusters (communities) by disambiguating and grouping the ambiguous objects. {\watset} is a meta-algorithm that uses existing \textit{hard} clustering algorithms for graphs to obtain a \textit{fuzzy} clustering, a.k.a. \textit{soft} clustering.

In computational linguistics, graph clustering is used for addressing problems such as word sense induction~\citep{Biemann:06}, lexical chain computing~\citep{Medelyan:07}, Web search results diversification~\citep{DiMarco:13}, sentiment analysis~\citep{Pang:04}, cross-lingual semantic relationship induction~\citep{Lewis:13:relations}; more applications can be found in the book by  \citet{Mihalcea:11}.

\paragraph{Definitions} Let $G = (V, E)$ be an undirected simple graph,\footnote{A simple graph has no loops, i.e., $u \neq v$, $\forall \{u, v\} \in E$. We use this property for context disambiguation in Section~\ref{sub:watset:wsd}.} where $V$ is a set of nodes and $E \subseteq V^2$ is a set of undirected edges. We denote a subset of nodes $C^i \subseteq V$ as a cluster. A graph clustering algorithm then is a function $\textsc{Cluster} : (V, E) \to C$ such that ${V = \bigcup_{C^i \in C} C^i}$. We distinguish two classes of graph clustering algorithms: \textit{hard} clustering algorithms (partitionings) produce non-overlapping clusters, i.e., ${C^i \cap C^j = \emptyset} \iff i \neq j$, $\forall C^i, C^j \in C$, while \textit{fuzzy} clustering algorithms permit cluster overlapping, i.e., a node can be a member of several clusters in $C$.

\subsection{\label{sub:outline}Outline of {\watset}, a Fuzzy Method for Local-Global Graph Clustering}

\begin{figure}[t]
  \centering
  \includegraphics[width=\linewidth]{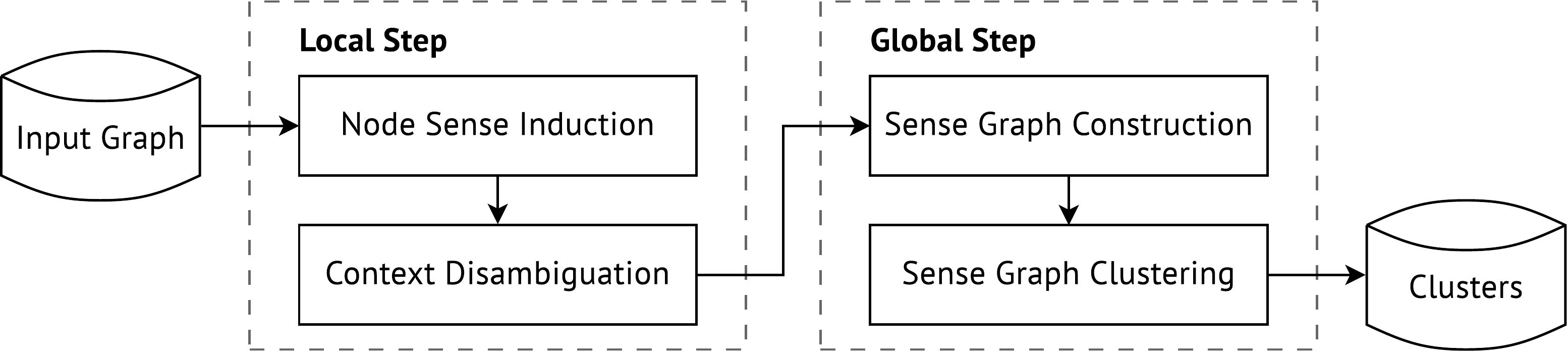}
  \caption{\label{fig:watset:outline}The outline of the {\watset} algorithm showing the \textit{local} step of word sense induction and context disambiguation, and the \textit{global} step of sense graph constructing and clustering.}
\end{figure}

{\watset} constructs an intermediate representation of the input graph called a \textit{sense graph}, which has been sketched as a ``disambiguated word graph'' in~\citet{Biemann:12}. This is achieved by node sense induction based on hard clustering of the input graph node neighborhoods. The sense graph has the edges established between the different \textit{senses} of the input graph nodes. The global clusters of the input graph are obtained by applying a hard clustering algorithm to the sense graph; removal of the sense labels yields overlapping clusters.

An outline of our algorithm is depicted in \figurename~\ref{fig:watset:outline}. {\watset} takes an undirected graph $G = (V, E)$ as the input and outputs a set of clusters $C$. The algorithm has two steps: local and global. The \textit{local} step, as described in Section~\ref{sub:local}, disambiguates the potentially ambiguous nodes in $G$. The \textit{global} step, as described in Section~\ref{sub:global}, uses these disambiguated nodes to construct an intermediate sense graph $\mathcal{G} = (\mathcal{V}, \mathcal{E})$ and produce the overlapping clustering $C$. {\watset} is parameterized by two graph partitioning algorithms $\mathrm{Cluster\textsubscript{Local}}$ and $\mathrm{Cluster\textsubscript{Global}}$, and a context similarity measure $\ssim$. The complete pseudocode of {\watset} is presented in Algorithm~\ref{alg:watset}. For the sake of illustration, while describing the approach, we will provide examples with words and their synonyms. However, {\watset} is not bound only to the lexical units and relationships, so our examples are given \textit{without loss of generality}. Note also that {\watset} can be applied for both unweighted and weighted graphs as soon as the underlying hard clustering algorithms $\mathrm{Cluster\textsubscript{Local}}$ and $\mathrm{Cluster\textsubscript{Global}}$ take edge weights into account.

\begin{algorithm}[t]
\caption{\label{alg:watset}\watset, a Local-Global Meta-Algorithm for Fuzzy Graph Clustering.}
\begin{algorithmic}[1]
\REQUIRE{graph $G = (V, E)$,}
\REQUIREP{hard clustering algorithms $\mathrm{Cluster\textsubscript{Local}}$ and $\mathrm{Cluster\textsubscript{Global}}$,}
\REQUIREP{context similarity measure $\ssim : (\ctx(a), \ctx(b)) \to \mathbb{R}$, $\forall \ctx(a), \ctx(b) \subseteq V$.}
\ENSURE{clusters $C$.}
\FORALL[Local Step: Sense Induction]{\label{alg:watset:wsi:begin}$u \in V$}
\STATE{$\senses(u) \gets \emptyset$}
\STATE{\label{alg:watset:ego:begin}$V_u \gets \{v \in V : \{u, v\} \in E\}$}\COMMENT{Note that $u \notin V_u$}
\STATE{$E_u \gets \{\{v, w\} \in E : v, w \in V_u\}$}
\STATE{\label{alg:watset:ego:end}$G_u \gets (V_u, E_u)$}
\STATE{\label{alg:watset:local}$C_u \gets \mathrm{Cluster\textsubscript{Local}}(G_u)$}\COMMENT{Cluster the open neighborhood of $u$}
\FORALL{\label{alg:watset:ctx:begin}$C^i_u \in C_u$}
\STATE{$\ctx(\sense{u}{i}) \gets C^i_u$}
\STATE{\label{alg:watset:ctx:end}\label{alg:watset:wsi:end}$\senses(u) \gets \senses(u) \cup \{\sense{u}{i}\}$}
\ENDFOR
\ENDFOR
\STATE{\label{alg:watset:senses}$\mathcal{V} \gets \bigcup_{u \in V} \senses(u)$}\COMMENT{Global Step: Sense Graph Nodes}
\FORALL[Local Step: Context Disambiguation]{\label{alg:watset:wsd:begin}$\hat{u} \in \mathcal{V}$}
\STATE{$\widehat{\ctx}(\hat{u}) \gets \emptyset$}
\FORALL{\label{alg:watset:dctx:begin}$v \in \ctx(\hat{u})$}
\STATE{$\hat{v} \gets {\arg\max}_{v' \in \senses(v)} \ssim(\ctx(\hat{u}) \cup \{u\}, \ctx(v'))$}\COMMENT{$\hat{u}$ is a sense of $u \in V$}
\STATE{\label{alg:watset:dctx:end}\label{alg:watset:wsd:end}$\widehat{\ctx}(\hat{u}) \gets \widehat{\ctx}(\hat{u}) \cup \{\hat{v}\}$}
\ENDFOR
\ENDFOR
\STATE{\label{alg:watset:dgraph:begin}$\mathcal{E} \gets \{\{\hat{u}, \hat{v}\} \in \mathcal{V}^2 : \hat{v} \in \widehat{\ctx}(\hat{u})\}$}\COMMENT{Global Step: Sense Graph Edges}
\STATE{\label{alg:watset:dgraph:end}$\mathcal{G} \gets (\mathcal{V}, \mathcal{E})$}\COMMENT{Global Step: Sense Graph Construction}
\STATE{\label{alg:watset:global}$\mathcal{C} \gets \mathrm{Cluster\textsubscript{Global}}(\mathcal{G})$}\COMMENT{Global Step: Sense Graph Clustering}
\STATE{\label{alg:watset:delabel}$C \gets \{\{u \in V : \hat{u} \in \mathcal{C}^i\} \subseteq V : \mathcal{C}^i \in \mathcal{C}\}$}\COMMENT{Remove the sense labels}
\RETURN{$C$}
\end{algorithmic}
\end{algorithm}

\subsection{\label{sub:local}Local Step: Node Sense Induction and Disambiguation}

The \textit{local} step of {\watset} discovers the node senses in the input graph and uses this information to discover which particular senses of the nodes were connected via the edges of the input graph $G$.

\begin{figure}[htbp]
  \centering
  \includegraphics[scale=.5]{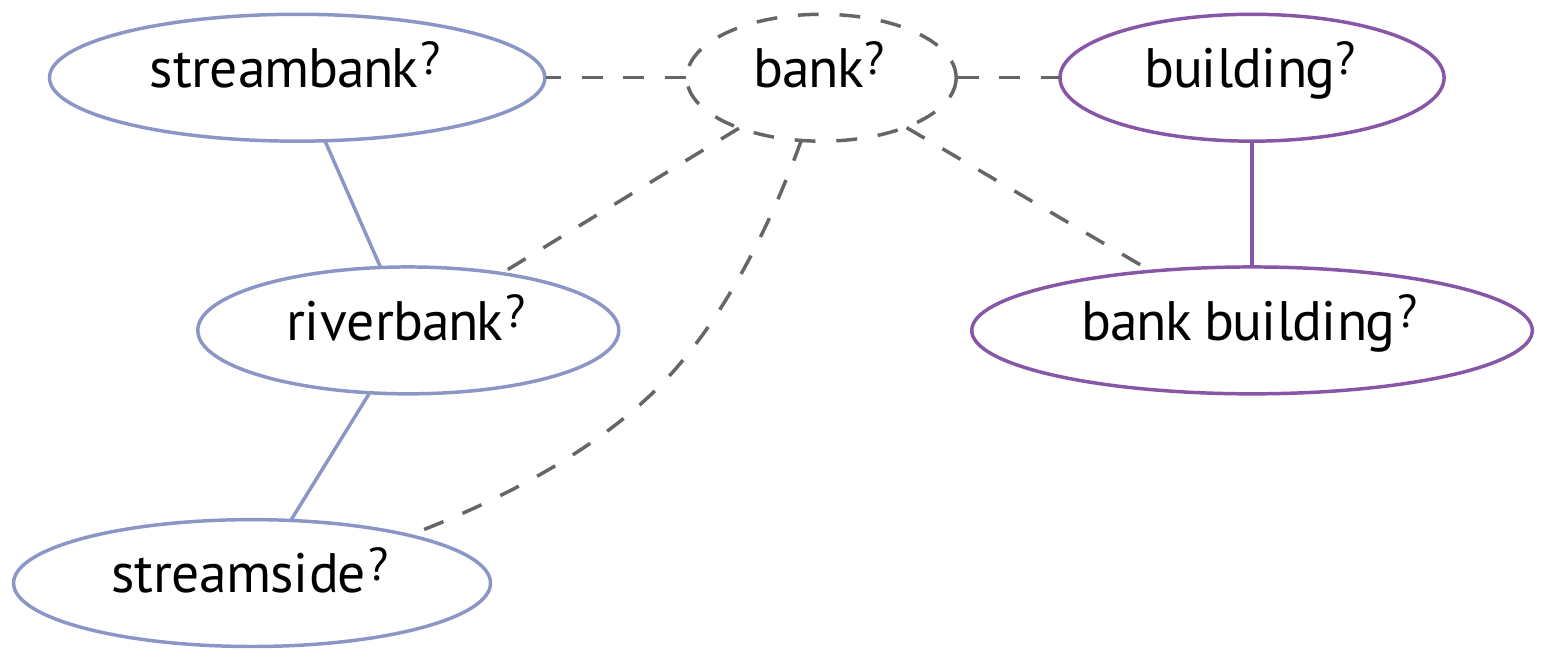}
  \caption{\label{fig:watset:wsi}Clustering the neighborhood of the node ``bank'' of the input graph results in two clusters treated as the non-disambiguated sense contexts: \textcolor{ggplotcount}{$\sense{bank}{1} = \{\textit{streambank}, \textit{riverbank}, \dots\}$} and \textcolor{ggplotsim}{$\{\sense{bank}{2} = \textit{bank building}, \textit{building}, \dots\}$}.}
\end{figure}

\subsubsection{Node Sense Induction} We induce node senses using the word neighborhood clustering approach by~\citet{Dorow:03}. In particular, we assume that the removal of the nodes participating in many triangles separates a graph into several connected components. Each component corresponds to the sense of the target node, so this procedure is executed for every node independently. \figurename~\ref{fig:watset:wsi} illustrates this approach for sense induction. For related work on word sense induction approaches, see the survey in Section~\ref{sub:wsi}.

Given a node $u \in V$, we extract its open neighborhood $G_u = (V_u, E_u)$ from the input graph $G$, such that the target node $u$ is not included into $V_u$ (lines~\ref{alg:watset:ego:begin}--\ref{alg:watset:ego:end}):
\begin{align}
V_u &= \{v \in V : \{u, v\} \in E\}\text{,} \\
E_u &= \{\{v, w\} \in E : v, w \in V_u\}\text{.}
\end{align}

Then, we run a hard graph clustering algorithm on $G_u$ that assigns one node to one and only one cluster, yielding a clustering $C_u$ (line~\ref{alg:watset:local}). We treat each obtained cluster $C^i_u \in C_u \subset V_u$ as representing a context for a different sense of the node $u \in V$ (lines~\ref{alg:watset:ctx:begin}--\ref{alg:watset:ctx:end}). We denote, e.g., \sense{bank}{1}, \sense{bank}{2} and other labels as the node \textit{senses} referred to as $\senses(\text{bank})$. In the example in Table~\ref{tab:watset:wsi}, $\left|\senses(\text{bank})\right| = 4$. Given a sense $u^i \in \senses(u)$, we denote ${\ctx(u^i) = C^i_u}$ as a \textit{context} of this sense of the node $u \in V$. Execution of this procedure for all the words in $V$ results in the set of senses for the global step (line~\ref{alg:watset:senses}):
\begin{equation}
  \mathcal{V} = \bigcup_{u \in V} \senses(u)\text{.}
  \label{eq:senses}
\end{equation}

\begin{table}[t]
\centering
\caption{\label{tab:watset:wsi}Example of induced senses for the node ``bank'' and the corresponding clusters (contexts).}
\begin{tabular}{p{83mm}r}\toprule
\textbf{Sense}  & \textbf{Context} \\\midrule
\sense{bank}{1} & $\{\textit{streambank}, \textit{riverbank}, \dots\}$\\
\sense{bank}{2} & $\{\textit{bank building}, \textit{building}, \dots\}$\\
\sense{bank}{3} & $\{\textit{bank company}, \dots\}$\\
\sense{bank}{4} & $\{\textit{coin bank}, \textit{penny bank}, \dots\}$\\\bottomrule
\end{tabular}
\end{table}

\begin{figure}[t]
  \centering
  \includegraphics[scale=.5]{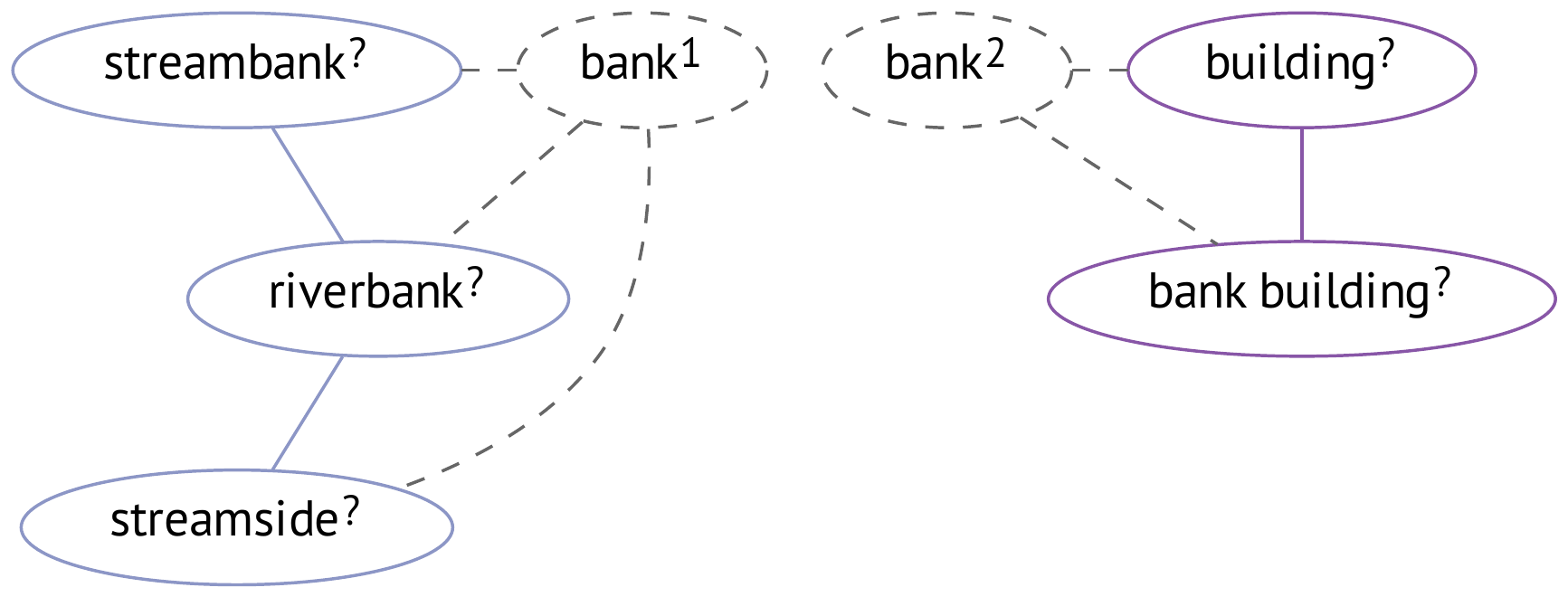}
  \caption{\label{fig:watset:ctx}Contexts for two different senses of the node ``bank'': only its senses \textcolor{ggplotcount}{\sense{bank}{1}} and \textcolor{ggplotsim}{\sense{bank}{2}} are currently known, while the other nodes in contexts need to be disambiguated.}
\end{figure}

\subsubsection{\label{sub:watset:wsd}Disambiguation of Neighbors} Although at the previous step we have induced node senses and mapped them to the corresponding contexts~(Table~\ref{tab:watset:wsi}), the elements of these contexts do not contain sense information. For example, the context of $\sense{bank}{2}$ in \figurename~\ref{fig:watset:ctx} has two elements $\{\sense{bank building}{?}, \sense{building}{?}\}$, the sense labels of which are currently not known. We recover the sense labels of nodes in a context using the sense disambiguated approach proposed by \citet{Faralli:16} as follows.

We represent each context as a vector in a vector space model~\citep{Salton:75} constructed for all the contexts. Since the graph $G$ is simple (Section~\ref{sec:watset}) and the context of any sense $\hat{u} \in \mathcal{V}$ does not include the corresponding node $u \in V$ (Table~\ref{tab:watset:wsi}), we \textit{temporarily} put it into the context during disambiguation. This prevents the situation of non-matching when the context of a candidate sense $v' \in \senses(v)$ has only one element and that element is $u$, i.e., $\ctx(v') = \{u\}$. We intentionally perform this insertion temporarily only during matching to prevent self-referencing. When a context $\ctx(\hat{u}) \subset V$ is transformed into a vector, we assign to each element $v \in \ctx(\hat{u})$ of this vector a weight equal to the weight of the edge $\{u, v\} \in E$ of the input graph $G$. If $G$ in unweighted, we assign $1$ if and only if $\{u, v\} \in E$, otherwise $0$ is assigned. Table~\ref{tab:watset:vsm} shows an example of the context vectors used for disambiguating the word \textit{building} in the context of the sense $\sense{bank}{2}$ in Figure~\ref{fig:watset:ctx}. In this example the vectors essentially represent one-hot encoding as the example input graph is unweighted.

\begin{table}[t]
\centering
\caption{\label{tab:watset:vsm}An example of context vectors for the node senses demonstrated in Figures~\ref{fig:watset:ctx} and \ref{fig:watset:wsd}. Since the graph is unweighted, one-hot encoding has been used. For matching purposes, the word \textcolor{ggplotsim}{``bank''} is temporarily added into \textcolor{ggplotsim}{$\ctx(\sense{bank}{2})$}.}
\begin{tabular}{lccccc}\toprule
\textbf{Sense} & \textbf{bank} & \textbf{bank building} & \textbf{building} & \textbf{construction} & \textbf{edifice} \\\midrule
\sense{bank}{2}     & \textcolor{ggplotsim}{$1$} & $1$ & $1$ & $0$ & $0$ \\
\sense{building}{1} & $1$ & $1$ & $0$ & $1$ & $0$ \\
\sense{building}{2} & $0$ & $0$ & $0$ & $0$ & $1$ \\\bottomrule
\end{tabular}
\end{table}

Then, given a sense $\hat{u} \in \mathcal{V}$ of a node $u \in V$ and the context of this sense $\ctx(\hat{u}) \subset V$, we \textit{disambiguate} each node $v \in \ctx(\hat{u})$. For that, we find the sense $\hat{v} \in \senses(v)$ the context $\ctx(\hat{v}) \subset V$ of which maximizes the similarity to the target context $\ctx(\hat{u})$. We compute the similarity using a context similarity measure $\ssim : (\ctx(a), \ctx(b)) \to \mathbb{R}$, $\forall \ctx(a), \ctx(b) \subseteq V$.\footnote{For the sake of brevity, by \textit{context similarity} we mean \textit{similarity between context vectors in a sparse vector space model} \citep{Salton:75}.} Typical choices for the similarity measure are dot product, cosine similarity, Jaccard index, etc. Hence, we \textit{disambiguate} each context element $v \in \ctx(\hat{u})$:
\begin{equation}
  \hat{v} = {\arg\max}_{v' \in \senses(v)} \ssim(\ctx(\hat{u}) \cup \{u\}, \ctx(v'))\text{.}
  \label{eq:dctx}
\end{equation}

\begin{figure}[t]
  \centering
  \includegraphics[scale=.45]{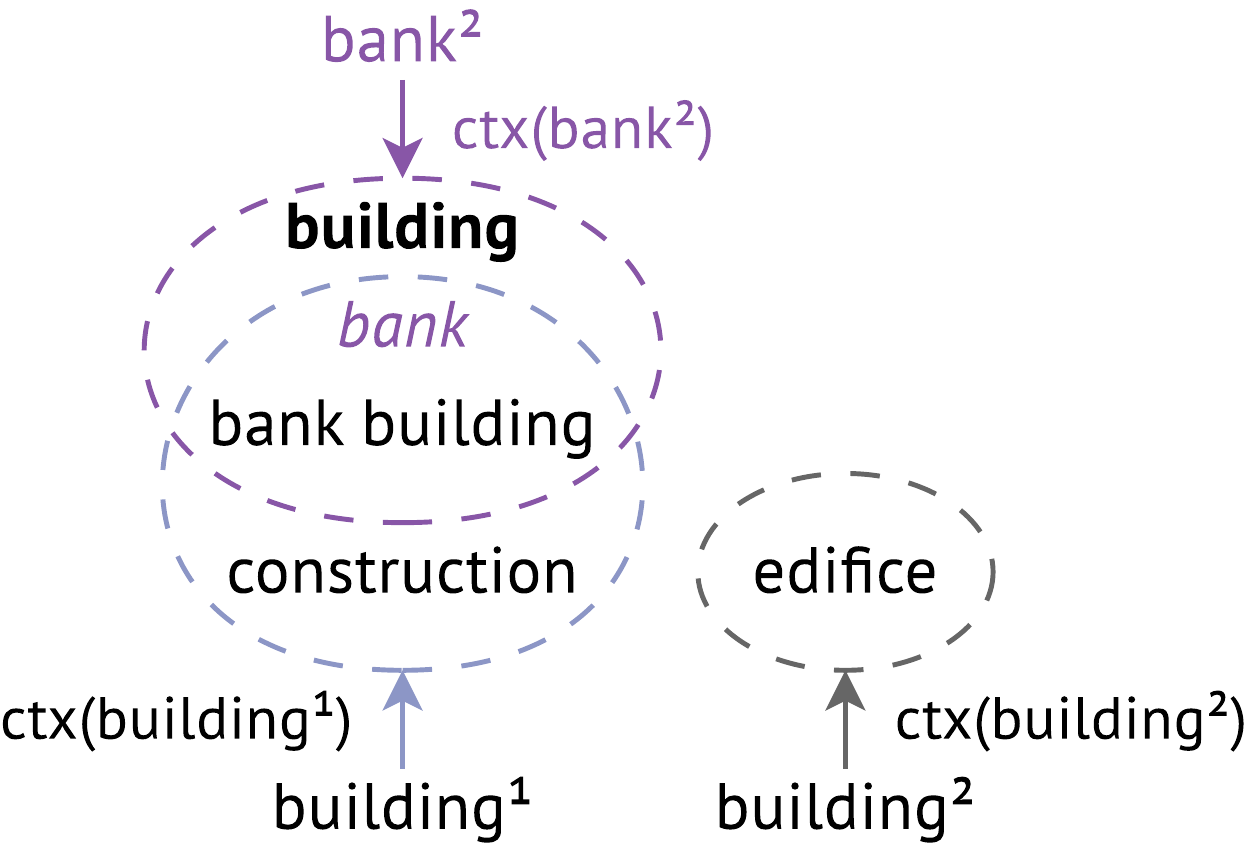}
  \caption{\label{fig:watset:wsd}Matching the meaning of the ambiguous node ``building'' in the context of the sense \textcolor{ggplotsim}{\sense{bank}{2}}. For matching purposes, the word \textcolor{ggplotsim}{``bank''} is temporarily added into \textcolor{ggplotsim}{$\ctx(\sense{bank}{2})$}.}
\end{figure}

An example in Figure~\ref{fig:watset:wsd} illustrates the node sense disambiguation process. The context of the sense $\sense{bank}{2}$ is $\ctx(\sense{bank}{2}) = \{\text{building}, \text{bank building}\}$ and the disambiguation target is \textit{building}. Having chosen cosine similarity as the context similarity measure, we compute the similarity between $\ctx(\sense{bank}{2} \cup \{\text{bank}\})$ and the context of every sense of \textit{building} in Table~\ref{tab:watset:vsm}: ${\cos(\ctx(\sense{bank}{2}) \cup \{\text{bank}\}, \ctx(\sense{building}{1}))} = \frac{2}{3}$ and ${\cos(\ctx(\sense{bank}{2}) \cup \{\text{bank}\}, \ctx(\sense{building}{2})) = 0}$. Therefore, for the word \textit{building} in the context of \sense{bank}{2}, its first sense, \sense{building}{1}, should be used because its similarity value is higher.

Finally, we construct a disambiguated context $\widehat{\ctx}(\hat{u}) \subset \mathcal{V}$ which is a sense-aware representation of $\ctx(\hat{u})$. This disambiguated context indicates which node senses were connected to $\hat{u} \in \mathcal{V}$ in the input graph $G$. For that, in lines~\ref{alg:watset:dctx:begin}--\ref{alg:watset:dctx:end}, we apply the disambiguation procedure defined in Equation~\eqref{eq:dctx} for every node $v \in \ctx(\hat{u})$:
\begin{equation}
  \widehat{\ctx}(\hat{u}) = \{\hat{v} \in \mathcal{V} : v \in \ctx(\hat{u})\}\text{.}
\end{equation}

As the result of the \textit{local} step, for each node $u \in V$ in the input graph, we induce the $\senses(u) \subset \mathcal{V}$ of nodes and provide each sense $\hat{u} \in \mathcal{V}$ with a disambiguated context $\widehat{\ctx}(\hat{u}) \subseteq \mathcal{V}$.

\subsection{\label{sub:global}Global Step: Sense Graph Construction and Clustering}

The \textit{global} step of {\watset} constructs an intermediate \textit{sense graph} expressing the connections between the node senses discovered at the \textit{local} step. We assume that the nodes $\mathcal{V}$ of the sense graph are non-ambiguous, so running a hard clustering algorithm on this graph outputs clusters $C$ covering the set of nodes $V$ of the input graph $G$.

\subsubsection{Sense Graph Construction} Using the set of node senses defined in Equation~\eqref{eq:senses}, we construct the sense graph $\mathcal{G} = (\mathcal{V}, \mathcal{E})$ by establishing undirected edges between the senses connected through the disambiguated contexts (lines~\ref{alg:watset:dgraph:begin}--\ref{alg:watset:dgraph:end}):
\begin{equation}
  \mathcal{E} = \{\{\hat{u}, \hat{v}\} \in \mathcal{V}^2 : \hat{v} \in \widehat{\ctx}(\hat{u})\}\text{.}
\end{equation}

Note that this edge construction approach disambiguates the edges $E$ such that if a pair of nodes was connected in the input graph $G$, then the corresponding sense nodes will be connected in the sense graph $\mathcal{G}$. As the result, the constructed sense graph $\mathcal{G}$ is a sense-aware representation of the input graph $G$. In case $G$ is weighted, we assign each edge $\{\hat{u}, \hat{v}\} \in \mathcal{E}$ the same weight as the edge $\{u, v\} \in E$ has in the input graph.

\subsubsection{Sense Graph Clustering} Running a hard clustering algorithm on $\mathcal{G}$ produces the set of sense-aware clusters $\mathcal{C}$, each sense-aware cluster $\mathcal{C}^i \in \mathcal{C}$ is a subset of $\mathcal{V}$ (line~\ref{alg:watset:global}). In order to obtain the set of clusters $C$ that covers the set of nodes $V$ of the input graph $G$, we simply remove the sense labels from the elements of clusters $\mathcal{C}$ (line~\ref{alg:watset:delabel}):
\begin{equation}
  C = \big\{\{u \in V : \hat{u} \in \mathcal{C}^i\} \subseteq V : \mathcal{C}^i \in \mathcal{C}\big\}\text{.}
\end{equation}

\begin{figure}[t]
  \centering
  \includegraphics[scale=.5]{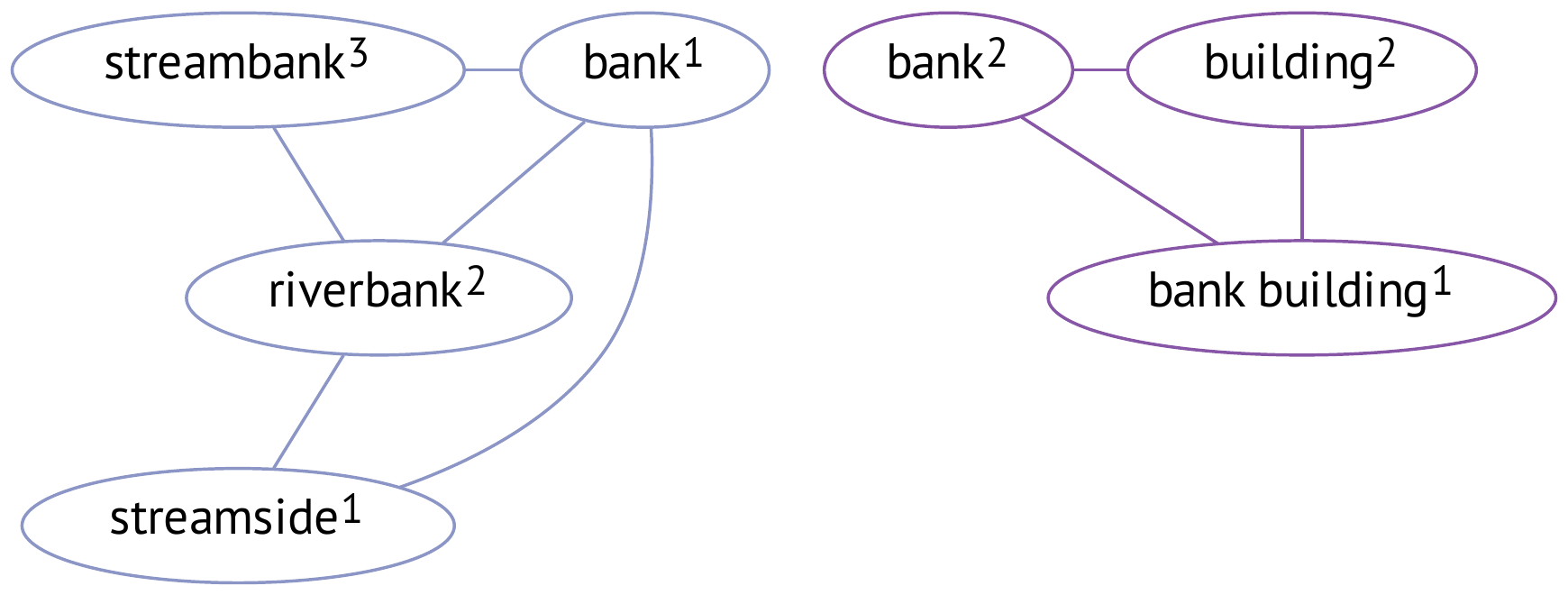}
  \caption{\label{fig:watset:global}Clustering of the \textit{sense graph} $\mathcal{G}$ yields two clusters, \textcolor{ggplotcount}{$\{\sense{bank}{1}, \sense{streambank}{3}, \sense{riverbank}{2}, \dots\}$} and \textcolor{ggplotsim}{$\{\sense{bank}{2}, \sense{bank building}{1}, \sense{building}{2}, \dots\}$}; if one removes the sense labels, the clusters will overlap resulting in a \textit{soft} clustering of the input graph $G$.}
\end{figure}

\figurename~\ref{fig:watset:global} illustrates the sense graph and its clustering on the example of the node ``bank''. The construction of a sense graph requires disambiguation of the input graph nodes. Note that traditional approaches to graph-based sense induction, such as the ones proposed by \citet{Veronis:04,Biemann:06,Hope:13:maxmax}, do not perform this step, but perform only local clustering of the graph since they do not aim at a global representation of clusters.

As the result of the \textit{global} step, a set of clusters $C$ of the input graph $G$ is obtained using an intermediate sense-aware graph $\mathcal{G}$. The presented local-global graph clustering approach, {\watset}, makes it possible to naturally achieve a \textit{soft} clustering of a graph using \textit{hard} clustering algorithms only.

\subsection{\label{sub:simplified}Simplified {\watset}}

The original {\watset} algorithm, as previously published \citep{Ustalov:17:watset} and described in Section~\ref{sub:outline}, has context construction and disambiguation steps. These steps involve computation of a context similarity measure, which needs to be chosen as a hyper-parameter of the algorithm (Section~\ref{sub:watset:wsd}). In this section, we propose a simplified version of {\watset} (Algorithm~\ref{alg:watset:simplified}) that requires no context similarity measure, which leads to faster computation in practice with less hyper-parameter tuning. As our experiments throughout the article show, this simplified version demonstrates similar performance to the original {\watset} algorithm.

\begin{algorithm}[t]
\caption{\label{alg:watset:simplified}Simplified \watset.}
\begin{algorithmic}[1]
\REQUIRE{graph $G = (V, E)$, hard clustering algorithms $\mathrm{Cluster\textsubscript{Local}}$ and $\mathrm{Cluster\textsubscript{Global}}$.}
\ENSURE{clusters $C$.}
\STATE{$\mathcal{V} \gets \emptyset$}
\FORALL[Local Step: Sense Induction]{$u \in V$}
\STATE{$V_u \gets \{v \in V : \{u, v\} \in E\}$}\COMMENT{Note that $u \notin V_u$}
\STATE{$E_u \gets \{\{v, w\} \in E : v, w \in V_u\}$}
\STATE{$G_u \gets (V_u, E_u)$}
\STATE{$C_u \gets \mathrm{Cluster\textsubscript{Local}}(G_u)$}\COMMENT{Cluster the open neighborhood of $u$}
\FORALL{$C^i_u \in C_u$}
\FORALL{$v \in C^i_u$}
\STATE{$\senses[u][v] \gets i$}\COMMENT{Node $v$ is connected to the $i$-th sense of $u$}
\ENDFOR
\STATE{$\mathcal{V} \gets \mathcal{V} \cup \{u^i\}$}
\ENDFOR
\ENDFOR
\STATE{$\mathcal{E} \gets \{\{u^{\senses[u][v]}, v^{\senses[v][u]}\} \in \mathcal{V}^2 : \{u, v\} \in E\}$}\COMMENT{Global Step: Sense Graph Edges}
\STATE{$\mathcal{G} \gets (\mathcal{V}, \mathcal{E})$}\COMMENT{Global Step: Sense Graph Construction}
\STATE{$\mathcal{C} \gets \mathrm{Cluster\textsubscript{Global}}(\mathcal{G})$}\COMMENT{Global Step: Sense Graph Clustering}
\STATE{$C \gets \{\{u \in V : \hat{u} \in \mathcal{C}^i\} \subseteq V : \mathcal{C}^i \in \mathcal{C}\}$}\COMMENT{Remove the sense labels}
\RETURN{$C$}
\end{algorithmic}
\end{algorithm}

In the input graph $G$ a pair of nodes $\{u, v\} \in V^2$ can be incident to one and only one edge. Otherwise these nodes are not connected. Due to the use of a \textit{hard} clustering algorithm for node sense induction (Section~\ref{sub:wsi}), in any pair of nodes $\{u, v\} \in E$, the node $v$ can appear in the context of only one sense of $u$ and vice versa. Therefore, we can omit the context disambiguation step (Section~\ref{sub:watset:wsd}) by tracking the node sense identifiers produced during sense induction.

Given a pair $\{u, v\} \in E$, we reuse the sense information from Table~\ref{tab:watset:wsi} to determine which context of a sense $\hat{u} \in \mathcal{V}$ contains $v$. We denote this as $\senses[u][v] \in \mathbb{N}$, which indicates $v \in \ctx(u^{\senses[u][v]})$, i.e., the fact that node $v$ is connected to the node $u$ in the specified sense $u^{\senses[u][v]}$. Following the example in Figure~\ref{fig:watset:wsi}, if the context of \sense{bank}{1} contains the word \textit{streambank} then the context of one of the senses of \textit{streambank} must contain the word \textit{bank}, e.g., \sense{streambank}{3}. This information allows us to create Table~\ref{tab:watset:simplified} that allows producing the set of sense-aware edges by simultaneously retrieving the corresponding sense identifiers:
\begin{equation}
  \mathcal{E} = \left\{\{u^{\senses[u][v]}, v^{\senses[v][u]}\} \in \mathcal{V}^2 : \{u, v\} \in E\right\}\text{.}
\end{equation}

\begin{table}[t]
\centering
\caption{\label{tab:watset:simplified}Node sense identifier tracking in Simplified {\watset} as according to Figure~\ref{fig:watset:wsi}.}
\begin{tabular}{p{55mm}p{55mm}c}\toprule
\textbf{Source} & \textbf{Target} & \textbf{Index} \\\midrule
bank & streambank & 1 \\
& riverbank & 1 \\
& streamside & 1 \\\cmidrule{2-3}
& building & 2 \\
& bank building & 2 \\\midrule
streambank & bank & 3 \\
& riverbank & 3 \\\midrule
\dots \\\bottomrule
\end{tabular}
\end{table}

This allows us to construct the sense graph $\mathcal{G}$ in linear time $O(|E|)$ by querying the node sense index to disambiguate the input edges $E$ in a deterministic way. Other steps are identical to the original {\watset} algorithm (Section~\ref{sub:outline}). Simplified {\watset} is presented in Algorithm~\ref{alg:watset:simplified}.

\subsection{\label{sub:complexity}Algorithmic Complexity}

We analyze the computational complexity of the separate routines of {\watset} and then present the overall complexity compared to other hard and soft clustering algorithms. Our analysis is based on the assumption that the context similarity measure in Equation~\eqref{eq:dctx} can be computed in linear time with respect to the number of dimensions $d \in \mathbb{N}$. For instance, such measures as cosine and Jaccard satisfy this requirement. In all our experiments throughout the paper we use the cosine similarity measure: $\ssim(\ctx(a), \ctx(b)) = \cos(\ctx(a), \ctx(b))$, $\forall \ctx(a), \ctx(b) \subseteq V$. Provided that the context vectors are normalized, the complexity of such a measure is bound by the complexity of an inner product of two vectors, which is $O(\left|\ctx(a) \cup \ctx(b)\right|)$.

Since the running time of our algorithm depends on the task-specific choice of two hard clustering algorithms, $\mathrm{Cluster\textsubscript{Local}}$ and $\mathrm{Cluster\textsubscript{Global}}$, we report algorithm-specific analysis on two hard clustering algorithms that are popular in computational linguistics: Chinese Whispers (CW) by \citet{Biemann:06} and Markov Clustering (MCL) by \citet{vanDongen:00}. Given a graph $G = (V, E)$, the computational complexity is $O(\left|E\right|)$ for CW and $O(\left|V\right|^3)$ for MCL.\footnote{Although MCL can be implemented more efficiently than $O(\left|V\right|^3)$, cf.~\citet[p.~125]{vanDongen:00}, we would like to use the consistent worst case scenario notation for all the mentioned clustering algorithms.} Additionally, we denote $\deg_{\max}$ as the maximum degree of $G$. Note that while in general, $\deg_{\max}$ is bound by $|V|$, in the real natural language-derived graphs this variable is distributed according to a power law. It is small for the majority of the nodes in a graph, making average running times acceptable in practice as presented in Section~\ref{sec:runningtime}.

\subsubsection{Node Sense Induction} This operation is executed for every node of the input graph $G$, i.e., $\left|V\right|$ times. By definition of an undirected graph, the maximum number of neighbors of a node in $G$ is $\deg_{\max}$ and the maximum number of edges in a neighborhood is $\frac{\deg_{\max} (\deg_{\max} - 1)}{2}$. Thus, this operation takes $O(\left|V\right|\deg^2_{\max})$ steps with CW and $O(\left|V\right|\deg^3_{\max})$ steps with MCL.

\subsubsection{Disambiguation of Neighbors} Let $\senses_{\max}$ be the maximum number of senses for a node and $\ctx_{\max}$ be the maximum size of the node sense context. Thus, this operation takes $O(\left|V\right| \times \senses_{\max} \times \ctx_{\max})$ steps to iterate over all the node sense contexts. At each iteration, it scans all the senses of the ambiguous node in context and computes a similarity between its context and the candidate sense context in a linear time (Section~\ref{sub:complexity}). This requires $O(\senses_{\max} \times \ctx_{\max})$ steps per each node in context. Therefore, the whole operation takes $O(\left|V\right| \times \senses_{\max}^2 \times \ctx_{\max}^2)$ steps. Since the maximum number of node senses is observed in a special case when the neighborhood is an unconnected graph, $\senses_{\max} \leq \deg_{\max}$. Given the fact that the maximum context size is observed in a special case when the neighborhood is a fully connected graph, $\ctx_{\max} \leq \deg_{\max}$. Thus, disambiguation of all the node sense contexts takes $O(\left|V\right|\deg^4_{\max})$ steps. Note that since the simplified version of {\watset}, as described in Section~\ref{sub:simplified}, does not perform context disambiguation, this term should be taken into account only for the original version of {\watset} (Algorithm~\ref{alg:watset}).

\subsubsection{Sense Graph Clustering} Like the input graph $G$, the sense graph $\mathcal{G}$ is undirected, so it has at most $\left|V\right|\deg_{\max}$ nodes and $\frac{\left|V\right|\deg_{\max}(\left|V\right|\deg_{\max} - 1)}{2}$ edges. Thus, this operation takes $O(\left|V\right|^2\deg^2_{\max})$ steps with CW and $O(\left|V\right|^3\deg^3_{\max})$ steps with MCL.

\subsubsection{Overall Complexity} Table~\ref{tab:watset:complexity} presents comparison of {\watset} to other hard and soft graph clustering algorithms popular in computational linguistics,\footnote{Our survey was based on \citet{Mihalcea:11,DiMarco:13,Lewis:13:semantics}.} such as Chinese Whispers (CW) by \citet{Biemann:06}, Markov Clustering (MCL) by \citet{vanDongen:00}, and MaxMax by \citet{Hope:13:maxmax}. Additionally, we compare {\watset} to several graph clustering algorithms that are popular in network science, such as the Louvain method by \citet{Blondel:08} and Clique Percolation (CPM) by \citet{Palla:05}. The notation \watset{[MCL, CW]} means using MCL for local clustering and CW for global clustering, cf.\ the discussion on graph clustering algorithms in Section~\ref{sub:clustering}.

\begin{table}[t]
\centering
\caption{\label{tab:watset:complexity}Computational complexity of graph clustering algorithms, where $|V|$ is the number of vertices, $|E|$ is the number of edges, and $\deg_{\max}$ is the maximum degree of a vertex. For brevity, we do not insert rows corresponding to Simplified {\watset} (Algorithm~\ref{alg:watset:simplified}) that does not require the $\textcolor{comment}{O(\left|V\right|\deg^4_{\max})}$ term related to context disambiguation.}
\resizebox{1.0\linewidth}{!}{
\begin{tabular}{lcl}\toprule
\textbf{Algorithm} & \textbf{Hard or Soft} & \textbf{Computational Complexity} \\\midrule
Chinese Whispers~\citep{Biemann:06}    & hard  & $O(\left|E\right|)$            \\
Markov Clustering~\citep{vanDongen:00} & hard  & $O(\left|V\right|^3)$          \\
MaxMax~\citep{Hope:13:maxmax}          & soft  & $O(\left|E\right|)$            \\\midrule
Louvain method~\citep{Blondel:08}      & hard  & $O(\left|V\right| \log(\left|V\right|))$  \\
Clique Percolation~\citep{Palla:05}    & soft  & $2^{\left|V\right|}$ \\\midrule
\watset{[CW, CW]}                     & soft  & $O(\left|V\right|^2\deg^2_{\max} \textcolor{comment}{+ \left|V\right|\deg^4_{\max}})$ \\
\watset{[CW, MCL]}                    & soft  & $O(\left|V\right|^3\deg^3_{\max} \textcolor{comment}{+ \left|V\right|\deg^4_{\max}})$ \\
\watset{[MCL, CW]}                    & soft  & $O(\left|V\right|^2\deg^2_{\max} \textcolor{comment}{+ \left|V\right|\deg^4_{\max}})$ \\
\watset{[MCL, MCL]}                   & soft  & $O(\left|V\right|^3\deg^3_{\max} \textcolor{comment}{+ \left|V\right|\deg^4_{\max}})$ \\\bottomrule
\end{tabular}
}
\end{table}

The analysis shows that the most time-consuming operations in {\watset} are sense graph clustering and context disambiguation. Although the overall computational complexity of our meta-algorithm is higher than of the other methods, its compute-intensive operations, such as node sense induction and context disambiguation, are executed for every node independently, so the algorithm can easily be run in a parallel or a distributed way to reduce the running time.

\subsubsection{\label{sec:runningtime}An Empirical Evaluation of Average Running Times} In order to evaluate the running time of {\watset} on a real-world scenario, we applied it to the clustering of co-occurrence graphs. Word clusters discovered from co-occurrence graphs are the sets of semantically related polysemous words, so we ran our sense-aware clustering algorithm to obtain overlapping word clusters.

We used the English word co-occurrence graphs from the Leipzig Corpora Collection by~\citet{Goldhahn:12} since it is partitioned into corpora of different sizes.\footnote{\url{http://wortschatz.uni-leipzig.de/en/download}} We evaluated on the graphs corresponding to five different English corpus sizes: \texttt{10K}, \texttt{30K}, \texttt{100K}, \texttt{300K}, and \texttt{1M} tokens (\tablename~\ref{tab:watset:leipzig}). The measurements were made independently among the graphs using the \watset{[CW, CW]} algorithm with the lowest complexity bound by ${O(\left|V\right|^2\deg^2_{\max} + \left|V\right|\deg^4_{\max})}$.

\begin{table}[t]
\centering
\caption{\label{tab:watset:leipzig}Parameters of the co-occurrence graphs for different corpus sizes in the Leipzig Corpora Collection, where $|V|$ is the number of vertices, $|E|$ is the number of edges, and $\deg_{\max}$ is the maximum degree of a vertex; time is measured in minutes.}
\begin{tabular}{rrrrrr}\toprule
\textbf{Size} & $\mathbf{|V|}$ & $\mathbf{|E|}$ & $\mathbf{\deg_{\max}}$ & \textbf{Sequential Time, min.} & \textbf{Parallel Time, min.} \\\midrule
\texttt{10K}  & $4{,}907$   & $16{,}057$   & $547$    & $0.13 \pm 0.01$ & $0.04 \pm 0.00$ \\
\texttt{30K}  & $11{,}627$  & $55{,}181$   & $1{,}307$ & $0.91 \pm 0.05$ & $0.36 \pm 0.02$ \\
\texttt{100K} & $27{,}200$  & $203{,}946$  & $3{,}319$ & $9.33 \pm 0.13$ & $3.78 \pm 0.08$ \\
\texttt{300K} & $55{,}359$  & $630{,}138$  & $7{,}467$ & $53.34 \pm 0.16$ & $24.44 \pm 0.18$ \\
\texttt{1M}   & $117{,}141$ & $2{,}031{,}283$ & $18{,}081$ & $347.16 \pm 1.97$ & $158.00 \pm 1.88$ \\\bottomrule
\end{tabular}
\end{table}

Since our implementation of {\watset} in the Java programming language, as described in Section~\ref{sub:implementation}, is multi-threaded and runs node sense induction and context disambiguation steps in parallel, we study the benefit of multiple available central processing unit (CPU) cores to the overall running time. The single-threaded setup that uses only one CPU core will be referred to as \textit{sequential}, while the multi-threaded setup that uses all the CPU cores available on the machine will be referred to as \textit{parallel}.

For each graph, we ran {\watset} for five times. Following \citet{Horky:15}, the first three runs were used off-record to \textit{warm-up} the Java virtual machine. The next two runs were used for actual measurement. We used the following computational node for this experiment: two Intel Xeon E5-2630~v4 CPUs, 256~GB of ECC RAM, Ubuntu~16.04.4~LTS (Linux 4.13.0, x86\_64), Oracle~Java 8b121; 40 logical cores were available in total. \tablename~\ref{tab:watset:leipzig} reports the running time mean and the standard deviation for both setups, sequential and parallel.

\begin{figure}[t]
  \centering
  \includegraphics[scale=.75]{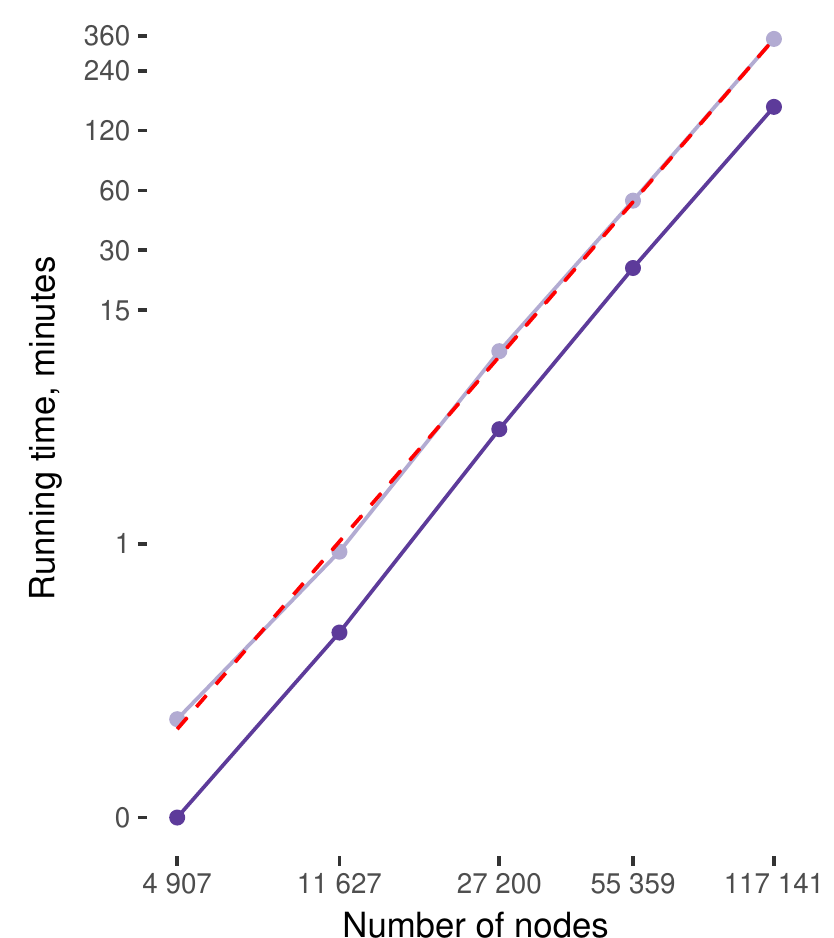} 
  \includegraphics[scale=.75]{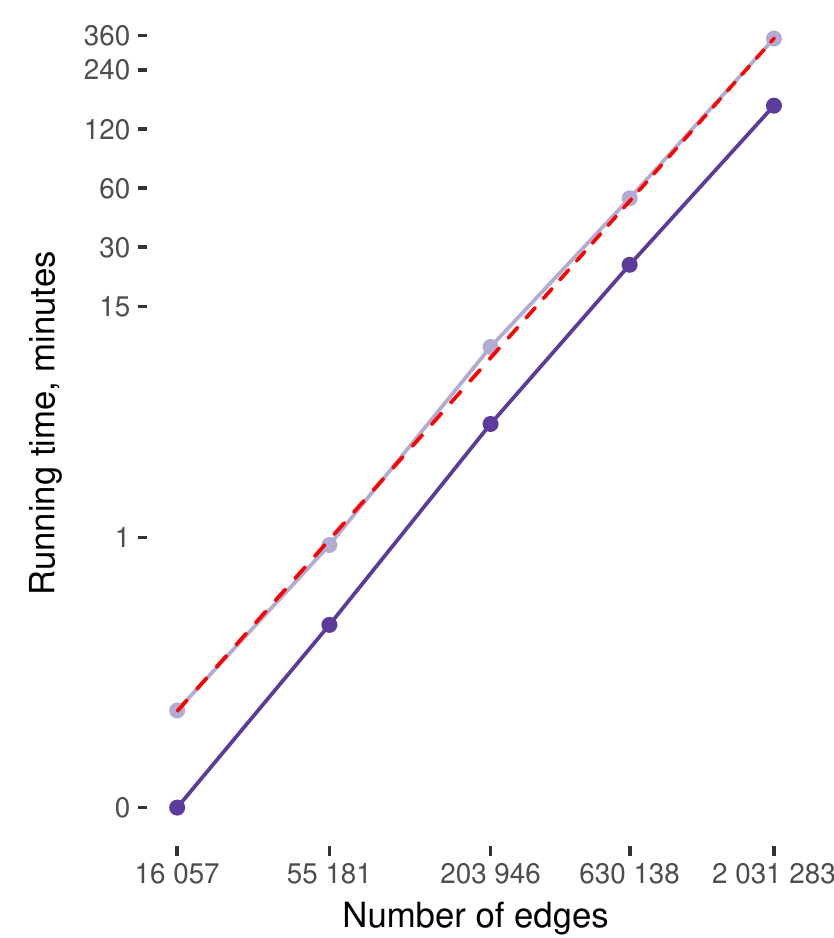} 
  \\{\footnotesize\textsf{Algorithm: \legend{ggplotsequential}\,sequential, \legend{ggplotparallel}\,parallel}.}
  \caption{\label{fig:watset:performance:both}\textit{Log-log} plots showing growth of the empirical average running time in number of nodes (left) and number of edges (right) of two \watset{[CW\textsubscript{top}, CW\textsubscript{top}]} setups: \textcolor{ggplotsequential}{sequential} and \textcolor{ggplotparallel}{parallel}. The \textcolor{red}{dashed} line is fitted to the running time data of the sequential version of {\watset}, showing polynomial growth in $O(|V|^{2.52})$ and $O(|E|^{1.63})$, respectively.}
\end{figure}

Figure~\ref{fig:watset:performance:both} shows the polynomial growth of $O(|V|^{2.52})$, which is smaller than the worst case of ${O(\left|V\right|^2\deg^2_{\max} + \left|V\right|\deg^4_{\max})}$. This is because in co-occurrence graphs, as well as in many other real-world graphs that also exhibit scale-free small world properties~\citep{Steyvers:05}, the degree distribution among nodes is strongly right-skewed. This makes {\watset} useful for processing real-world graphs. Both \tablename~\ref{tab:watset:leipzig} and Figure~\ref{fig:watset:performance:both} clearly confirm that {\watset} scales well and can be parallelized on multiple CPU cores, which makes it possible to process very large graphs.

\section{\label{sec:synsets}Application to Unsupervised Synset Induction}

A \textit{synset} is a set of mutual synonyms, which can be represented as a clique graph where nodes are words and edges are synonymy relationships. Synsets represent word senses and are building blocks of such such as thesauri and lexical ontologies as WordNet~\citep{Fellbaum:98}. These resources are crucial for many natural language processing applications that require common sense reasoning, such as information retrieval~\citep{Gong:05}, sentiment analysis~\citep{MontejoRaez:14}, and question answering~\citep{Kwok:01,Zhou:13}.

For most languages, no manually-constructed resource is available that is comparable to the English WordNet in terms of coverage and quality~\citep{Braslavski:16}. For instance, \citet{Kiselev:15} present a comparative analysis of lexical resources available for the Russian language concluding that there is no resource compared to WordNet in terms of completeness and availability for Russian. This lack of linguistic resources for many languages strongly motivates the development of new methods for automatic construction of WordNet-like resources. In this section, we apply {\watset} for unsupervised synset induction from a synonymy graph and compare it to state-of-the-art graph clustering algorithms ran on the same task.

\subsection{Synonymy Graph Construction and Clustering}

Wikipedia,\footnote{\url{http://www.wikipedia.org}} Wiktionary,\footnote{\url{http://www.wiktionary.org}} OmegaWiki\footnote{\url{http://www.omegawiki.org}} and other collaboratively-created resources contain a large amount of lexical semantic information---yet designed to be human-readable and not formally structured. While semantic relationships can be automatically extracted using tools such as DKPro JWKTL\footnote{\url{https://dkpro.github.io/dkpro-jwktl}} by \citet{Zesch:08} and Wikokit\footnote{\url{https://github.com/componavt/wikokit}} by \citet{Krizhanovsky:13}, words in these relationships are not disambiguated. For instance, the synonymy pairs $\{\textit{bank}, \textit{streambank}\}$ and $\{\textit{bank}, \textit{banking company}\}$ will be connected via the word ``bank'', while they refer to the different senses. This problem stems from the fact that articles in Wiktionary and similar resources list `undisambiguated' synonyms. They are easy to disambiguate for humans while reading a dictionary article but can be a source of errors for language processing systems.

Although large-scale automatically constructed lexical semantic resources like BabelNet~\citep{Navigli:12:babelnet} are available, they contain synsets with relationships other than synonymity. For instance, in BabelNet~4.0, the synset for \textit{bank} as an institution contains among other things non-synonyms like \textit{Monetary intermediation} and \textit{Money-lenders}.\footnote{\url{https://babelnet.org/synset?word=bn:00008364n}}

A synonymy dictionary can be perceived as a graph, where the nodes correspond to lexical units (words) and the edges connect pairs of the nodes when the synonymy relationship between them holds. Since such a graph can easily be obtained for arbitrary language, we expect that constructing and clustering a sense-aware representation of a synonymy graph yields plausible synsets covering polysemous words.

\subsubsection{\label{sub:graph}Synonymy Graph Construction} Given a synonymy dictionary, we construct the synonymy graph $G = (V, E)$ as follows. The set of nodes $V$ includes every lexical unit appearing in the input dictionary. An edge in the set of edges $E \subseteq V^2$ is established if and only if a pair of words are distinguished synonyms as according to the input synonymy dictionary. To enhance our representation with the contextual semantic similarity between synonyms, we assigned every edge $\{u, v\} \in E$ a weight equal to the cosine similarity of Skip-Gram word embeddings~\citep{Mikolov:13}. As the result, we obtained a weighted synonymy graph $G$.

\subsubsection{Synonymy Graph Clustering} Since the graph $G$ contains both monosemeous and polysemous words without indication of the particular senses, we run {\watset} to obtain a soft clustering $C$ of the synonymy graph $G$. Since our algorithm explicitly induces and clusters the word senses, the elements of the clusters $C$ are by definition synsets, i.e., sets of words that are synonymous with each other.

\subsection{Evaluation}

We conduct our experiments on resources from two different languages. We evaluate our approach on two datasets for English to demonstrate its performance in a resource-rich language. Additionally, we evaluate it on two Russian datasets since Russian is a good example of an under-resourced language with a clear need for synset induction~\citep{Kiselev:15}.

\subsubsection{\label{sub:synsets:setup}Experimental Setup} We compare {\watset} with five popular graph clustering methods presented in Section~\ref{sub:clustering}: Chinese Whispers (CW), Markov Clustering (MCL), MaxMax, ECO, and the Clique Percolation Method (CPM). The first two algorithms perform \textit{hard} clustering algorithms, while the last three are \textit{soft} clustering methods just like our method. Although the hard clustering algorithms are able to discover clusters that correspond to synsets composed of unambiguous words, they can produce wrong results in the presence of lexical ambiguity when a node should belong to several synsets. In our experiments, we use CW and MCL also as the underlying algorithms for local and global clustering in {\watset}, so our comparison will show the difference between the ``plain'' underlying algorithms and their utilization in {\watset}. We also report the performance of Simplified {\watset} (Section~\ref{sub:simplified}).

In our experiments, we rely on our own implementation of MaxMax and ECO as reference implementations are not available. For CW,\footnote{\url{https://github.com/uhh-lt/chinese-whispers}} MCL,\footnote{\url{https://micans.org/mcl/}} and CPM,\footnote{\url{https://networkx.github.io}} available implementations have been used. During the evaluation, we delete clusters equal to or larger than the threshold of 150 words as they can hardly represent any meaningful synset. Only the clusters produced by the MaxMax algorithm were actually affected by this threshold.

\paragraph{Quality Measure} To evaluate the quality of the induced synsets, we transform them into synonymy pairs and computed precision, recall, and F\textsubscript{1}-score on the basis of the overlap of these synonymy pairs with the synonymy pairs from the gold standard datasets. The F\textsubscript{1}-score calculated this way is known as \textit{paired F-score}~\citep{Manandhar:10,Hope:13:maxmax}. Let $C$ be the set of obtained synsets and $C_G$ be the set of gold synsets. Given a synset containing $n > 1$ words, we generate $\frac{n(n-1)}{2}$ pairs of synonyms, so we transform $C$ into a set of pairs $P$ and $C_G$ into a set of gold pairs $P_G$. We then compute the numbers of positive and negative answers as follows:
\begin{align}
  \mathrm{TP} &= |P \cup P_G|\text{,}\\
  \mathrm{FP} &= |P \setminus P_G|\text{,}\\
  \mathrm{FN} &= |P_G \setminus P|\text{,}
\end{align}
where $\mathrm{TP}$ is the number of true positives, $\mathrm{FP}$ is the number of false positives, and $\mathrm{FN}$ is the number of false negatives. As the result, we use the standard definitions of precision as ${\mathrm{Pr} = \frac{\mathrm{TP}}{\mathrm{TP} + \mathrm{FP}}}$, recall as ${\mathrm{Re} = \frac{\mathrm{TP}}{\mathrm{TP} + \mathrm{FN}}}$, and F\textsubscript{1}-score as ${\mathrm{F}_1 = \frac{2 \cdot \mathrm{Pr} \cdot \mathrm{Re}}{\mathrm{Pr} + \mathrm{Re}}}$. The advantage of this measure compared to other cluster evaluation measures, such as \textit{fuzzy B-Cubed}~\citep{Jurgens:13} and \textit{normalized modified purity}~\citep{Kawahara:14}, is its straightforward interpretability.

\paragraph{Statistical Testing} We evaluate the statistical significance of the experimental results using a McNemar's test~(\citeyear{McNemar:47}). Given the results of two algorithms, we build a $2 \times 2$ contingency table and compute the $p$-value of the test using the Statsmodels toolkit~\citep{Seabold:10}.\footnote{\url{https://www.statsmodels.org/}} Since the hypothesis tested by the McNemar's test is whether the results from both algorithms are similar against the alternative that they are not, we use the $p$-value of this test to assess the significance in the difference between F\textsubscript{1}-scores~\citep{Dror:18}. We consider the performance of one algorithm to be higher than the performance of another if its F\textsubscript{1}-score is larger and the corresponding $p$-value is smaller than a significance level of $0.01$.

\paragraph{Gold Standards} We conduct our evaluation on four lexical semantic resources for two different natural languages. Statistics of the gold standard datasets are present in Table~\ref{tab:watset:gold}. We report the number of lexical units (\# words), synsets (\# synsets), and the generated synonymy pairs (\# pairs).

\begin{table}[t]
\caption{\label{tab:watset:gold}Statistics of the gold standard datasets used in our experiments.}
\centering
\begin{tabular}{p{50mm}crrr}\toprule
\textbf{Resource} & \textbf{Language} & \textbf{\#~words} & \textbf{\#~synsets} & \textbf{\#~pairs} \\\midrule
WordNet   & \multirow{2}{*}{English} & $148{,}730$ & $117{,}659$ & $152{,}254$\\
BabelNet  & & $11{,}710{,}137$ & $6{,}667{,}855$ & $28{,}822{,}400$\\\midrule
RuWordNet & \multirow{2}{*}{Russian} & $110{,}242$ & $49{,}492$ & $278{,}381$\\
YARN      & & $9{,}141$ & $2{,}210$ & $48{,}291$\\\bottomrule
\end{tabular}
\end{table}

We use WordNet,\footnote{\url{https://wordnet.princeton.edu}} a popular \textit{English} lexical database constructed by expert lexicographers~\citep{Fellbaum:98}. WordNet contains general vocabulary and appears to be the \textit{de facto} gold standard in similar tasks~\citep{Hope:13:maxmax}. We used WordNet~3.1 to derive the synonymy pairs from synsets. Additionally, to compare to an automatically constructed lexical resource, we use BabelNet,\footnote{\url{https://www.babelnet.org}} a large-scale multilingual semantic network based on WordNet, Wikipedia and other resources~\citep{Navigli:12:babelnet}. We retrieved all the synonymy pairs from the BabelNet~3.7 synsets marked as English using the BabelNet Extract tool~\citep{Ustalov:17:babelnet}.

As a lexical ontology for \textit{Russian}, we use RuWordNet\footnote{\url{https://ruwordnet.ru/en}} by \citet{Loukachevitch:16}, containing both general vocabulary and domain-specific synsets related to sport, finance, economics, etc. Up to a half of the words in this resource are multi-word expressions~\citep{Kiselev:15}, which is due to the coverage of domain-specific vocabulary. RuWordNet is a WordNet-like version of the RuThes thesaurus that is constructed in the traditional way, namely by a small group of expert lexicographers \citep{Loukachevitch:11}. In addition, we use Yet Another RussNet\footnote{\url{https://russianword.net/en}} (YARN) by \citet{Braslavski:16} as another gold standard for Russian. The resource is constructed using crowdsourcing and mostly covers general vocabulary. In particular, non-expert users are allowed to edit synsets in a collaborative way, loosely supervised by a team of project curators. Due to the ongoing development of the resource, we selected as the silver standard only those synsets that were edited at least eight times in order to filter out noisy incomplete synsets.\footnote{In YARN, an edit operation can be an addition or a removal of a synset element; an average synset in our dataset contains $6.77 \pm 3.54$ words.} We do not use BabelNet for evaluating the Russian synsets as our manual inspection during prototyping showed, on average,  a much lower quality  than its English subset.

\paragraph{Input Data} For each language, we constructed a synonymy graph using openly available synonymy dictionaries. The statistics of the graphs used as the input in the further experiments are shown in \tablename~\ref{tab:watset:raw}.

\begin{table}[t]
\centering
\caption{\label{tab:watset:raw}Statistics of the input datasets used in our experiments.}
\begin{tabular}{p{95mm}rr}\toprule
\textbf{Language} & \textbf{\#~words} & \textbf{\#~pairs} \\\midrule
English & $243{,}840$ & $212{,}163$\\
Russian &  $83{,}092$ & $211{,}986$\\\bottomrule
\end{tabular}
\end{table}

For \textit{English}, synonyms were extracted from the English Wiktionary,\footnote{We used the Wiktionary dumps of February 1, 2017.} which is the largest Wiktionary at the present moment in terms of the lexical coverage, using the DKPro~JWKTL tool by~\citet{Zesch:08}. English words have been extracted from the dump.

For \textit{Russian}, synonyms from three sources were combined to improve lexical coverage of the input dictionary and to enforce confidence in jointly observed synonyms: (1) synonyms listed in the Russian Wiktionary extracted using the Wikokit tool by~\citet{Krizhanovsky:13}; (2) the dictionary of~\citet{Abramov:99}; and (3) the Universal Dictionary of Concepts~\citep{Dikonov:13}. While the two latter resources are specific to Russian, Wiktionary is available for most languages. Note that the same input synonymy dictionaries were used by authors of YARN to construct synsets using crowdsourcing. The results on the YARN dataset show how close an automatic synset induction method can approximate manually created synsets provided the same starting material.\footnote{We used the YARN dumps of February 7, 2017.}

Due to the vocabulary differences between the input data and the gold standard datasets, we use the intersection between the lexicon of the gold standard and the united lexicon of all the compared configurations of the algorithms during all the experiments in this section.

\subsubsection{\label{sub:parameters}Parameter Tuning} We tuned the hyper-parameters for such methods as CPM~\citep{Palla:05} and ECO~\citep{GoncaloOliveira:14} on the evaluation dataset. We do not perform any tuning of {\watset} because the underlying local and global clustering algorithms, CW and MCL, are parameter-free, so we use default configurations of them and their variations. As CPM\textsubscript{$k\!=\!3$} we denote that this method shown the best performance using the threshold value of $k=3$. For ECO, we found the threshold value of $\theta = 0.05$ yielding the best results, as opposed to the value of $\theta = 0.2$ suggested by \citet{GoncaloOliveira:14}.

We also study the performance impact of different edge weighting approaches for the same input graph. For that, we present the results of running the same algorithms in three different setups: \textsf{ones} that assigns every edge the constant weight of $1$, \textsf{count} that weights the edge $\{u, v\} \in E$ with the number of times a synonymy pair appeared in the input dictionary, and \textsf{sim} that uses cosine similarity between word embeddings as described in Section~\ref{sub:graph}. For English, we use the commonly used 300-dimensional word embeddings trained on the 100 billion tokens Google News corpus.\footnote{\url{https://code.google.com/archive/p/word2vec/}} For Russian, we use the 500-dimensional embeddings from the Russian Distributional Thesaurus (RDT) trained on a 12.9 billion tokens corpus of books, that yielded the state-of-art performance on a shared task on Russian semantic similarity \citep{Panchenko:17:rdt}.\footnote{\url{https://doi.org/10.5281/zenodo.163857}}

\subsubsection{\label{sub:watset:results}Results and Discussion} 
\begin{figure}[t]
  \centering
  \includegraphics[width=.4875\textwidth]{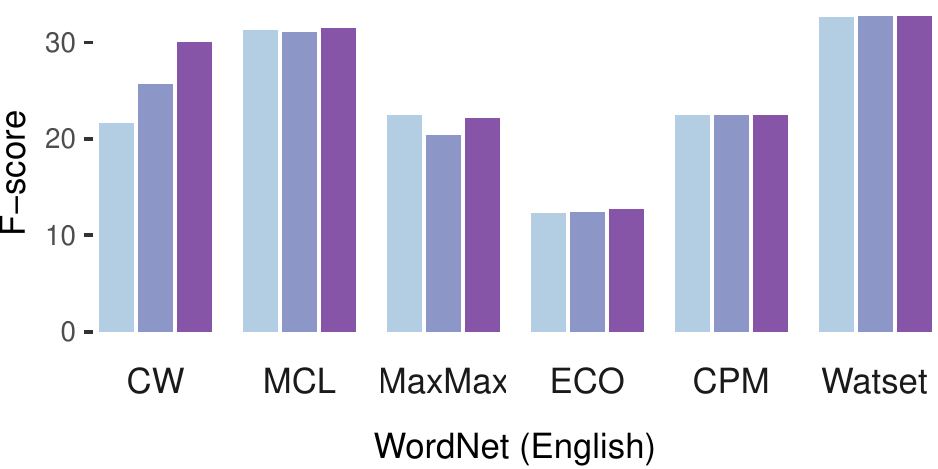} 
  \hfill
  \includegraphics[width=.4875\textwidth]{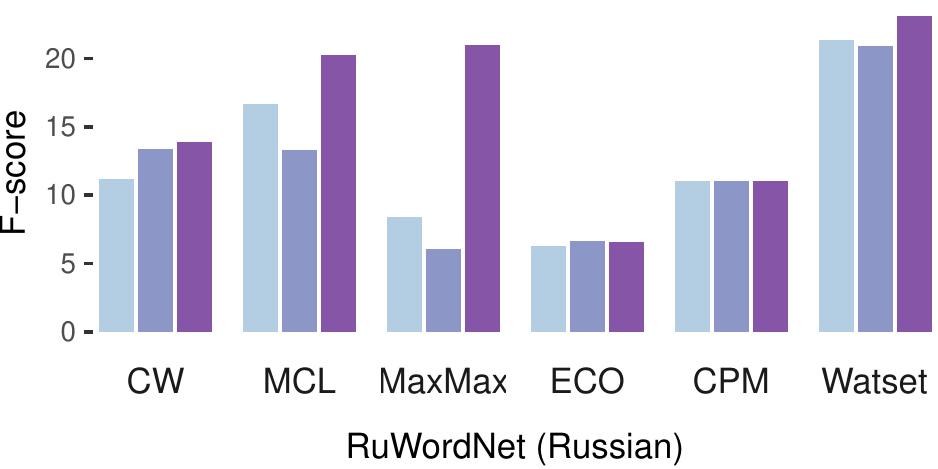} 
  \\
  \includegraphics[width=.4875\textwidth]{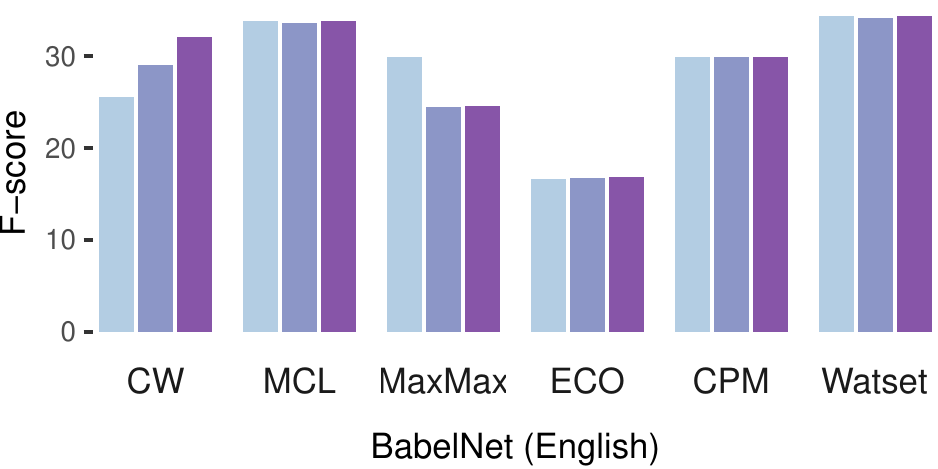} 
  \hfill
  \includegraphics[width=.4875\textwidth]{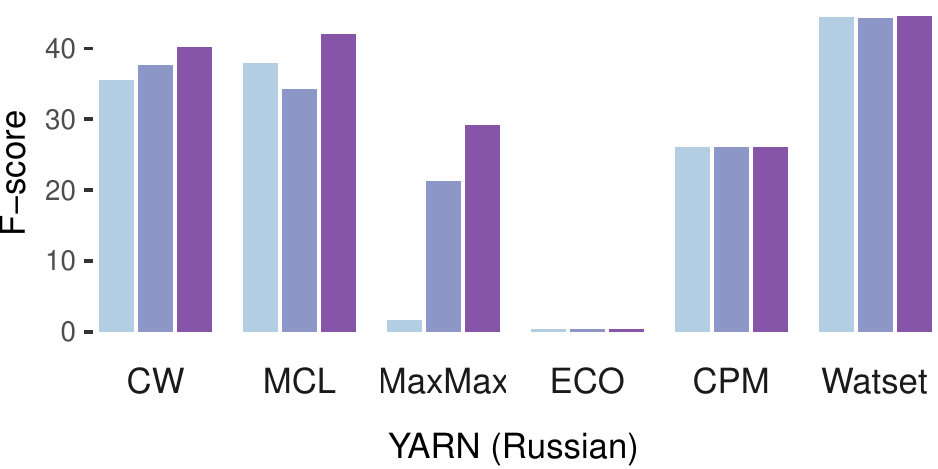} 
  \\{\scriptsize\textsf{Weighting: \legend{ggplotones}\,ones, \legend{ggplotcount}\,count, \legend{ggplotsim}\,sim.}}
  \caption{\label{fig:watset:weights}Impact of the different graph weighting schemas on the performance of synset induction. Each bar corresponds to the top performance of a method in Tables~\ref{tab:watset:english} and~\ref{tab:watset:russian}.}
\end{figure}

\figurename~\ref{fig:watset:weights} presents an overview of the evaluation results on both datasets. Since the synonymy graph construction step is the same for all the experiments, we start our analysis with the comparison of different edge weighting approaches introduced in Section~\ref{sub:parameters}: constant values (\textsf{ones}), frequencies (\textsf{count}), and semantic similarity scores (\textsf{sim}) based on word vector similarity. Results across various configurations and methods indicate that using the weights based on the similarity scores provided by word embeddings is the best strategy for all methods except MaxMax on the English datasets. However, its performance using the \textsf{ones} weighting does not exceed the other methods using the \textsf{sim} weighting. Therefore, we report all further results on the basis of the \textsf{sim} weights. The edge weighting scheme impacts Russian more for most algorithms. The CW algorithm, however, remains sensitive to the weighting also for the English dataset due to its randomized nature.

{\tablename}s~\ref{tab:watset:english} and~\ref{tab:watset:russian} present evaluation results for both languages. For each method, we show the best configurations in terms of F\textsubscript{1}-score. One may note that the granularity of the resulting synsets, especially for Russian, is very different, ranging from 4,000 synsets for the CPM{\textsubscript{$k\!=\!3$}} method to 67,645 induced by the ECO method. Both tables report the number of words, synsets, and synonyms after pruning huge clusters larger than 150 words. Without this pruning, the MaxMax and CPM methods tend to discover giant components obtaining almost zero precision as we generate all possible pairs of nodes in such clusters. The other methods did not exhibit such behavior.

\begin{table}[t]
\centering
\caption{\label{tab:watset:english}Comparison of the synset induction methods on datasets for English. All methods rely on the similarity edge weighting (\textsf{sim}); best configurations of each method in terms of F\textsubscript{1}-scores are shown for each dataset. Results are sorted by F\textsubscript{1}-score on BabelNet, top three values of each measure are boldfaced and statistically significant results are marked with an asterisk ($^\ast$). Simplified {\watset} is denoted as {\watset\S}.}
\resizebox{1.0\linewidth}{!}{
\begin{tabular}{l*{3}{r}|*{3}{c}|*{3}{c}}\toprule
\multirow{4}{*}{\textbf{Method}} & \multicolumn{1}{c}{\multirow{4}{*}{\rotatebox[origin=c]{90}{\textbf{\#~words}}}} & \multicolumn{1}{c}{\multirow{4}{*}{\rotatebox[origin=c]{90}{\textbf{\#~synsets}}}} & \multicolumn{1}{c|}{\multirow{4}{*}{\rotatebox[origin=c]{90}{\textbf{\#~pairs}}}} & \multicolumn{3}{c|}{} \\
\multicolumn{4}{c|}{} & \multicolumn{3}{c|}{\textbf{WordNet}} & \multicolumn{3}{c}{\textbf{BabelNet}} \\
\multicolumn{4}{c|}{} & \multicolumn{3}{c|}{} \\
\multicolumn{4}{c|}{} & \small\textbf{Pr} & \small\textbf{Re} & \small\textbf{F\textsubscript{1}} & \small\textbf{Pr} & \small\textbf{Re} & \small\textbf{F\textsubscript{1}}\\\midrule
\watset[MCL, MCL] & $243{,}840$ & $112{,}267$ & $345{,}883$ &
$34.48$ & $\mathbf{30.82}$ & $\mathbf{32.54}^\ast$ &
$40.01$ & $\mathbf{30.06}$ & $\underline{\mathbf{34.33}}^\ast$ \\
MCL & $243{,}840$ & $84{,}679$ & $387{,}315$ &
$34.21$ & $29.10$ & $31.45^\ast$ &
$38.98$ & $29.97$ & $\mathbf{33.89}^\ast$ \\
CW\textsubscript{top} & $243{,}840$ & $77{,}879$ & $539{,}753$ &
$28.54$ & $\underline{\mathbf{31.67}}$ & $30.02^\ast$ &
$32.57$ & $\underline{\mathbf{31.71}}$ & $\mathbf{32.14}^\ast$ \\
\watset[CW\textsubscript{log}, MCL] & $243{,}840$ & $164{,}689$ & $227{,}906$ &
$39.35$ & $27.99$ & $\underline{\mathbf{32.71}}^\ast$ &
$43.94$ & $24.47$ & $31.44^\ast$ \\
\watset\S[CW\textsubscript{top}, MCL] & $243{,}840$ & $164{,}683$ & $227{,}872$ &
$39.17$ & $27.83$ & $\mathbf{32.54}^\ast$ &
$43.87$ & $24.40$ & $31.36^\ast$ \\
\watset\S[CW\textsubscript{log}, MCL] & $243{,}840$ & $165{,}406$ & $222{,}554$ &
$\mathbf{40.20}$ & $27.44$ & $\mathbf{32.62}^\ast$ &
$\mathbf{44.63}$ & $24.09$ & $31.29^\ast$ \\
CPM{\textsubscript{$k\!=\!2$}} & $186{,}896$ & $67{,}109$ & $317{,}293$ &
$\mathbf{56.06}$ & $14.06$ & $22.48^\ast$ &
$\mathbf{49.23}$ & $21.44$ & $29.87^\ast$ \\
MaxMax & $219{,}892$ & $73{,}929$ & $797{,}743$ &
$17.59$ & $\mathbf{29.97}$ & $22.17^\ast$ &
$20.16$ & $\mathbf{31.34}$ & $24.53^\ast$ \\
ECO & $243{,}840$ & $171{,}773$ & $84{,}372$ &
$\underline{\mathbf{78.41}}$ & $6.95$ & $12.77\hphantom{^\ast}$ &
$\underline{\mathbf{69.91}}$ & $9.59$ & $16.87\hphantom{^\ast}$ \\\bottomrule
\end{tabular}
}
\end{table}

\begin{table}[t]
\centering
\caption{\label{tab:watset:russian}Results on datasets for Russian sorted by F\textsubscript{1}-score on Yet Another RussNet (YARN), top three values of each measure are boldfaced and statistically significant results are marked with an asterisk ($^\ast$). Simplified {\watset} is denoted as {\watset\S}.}
\resizebox{1.0\linewidth}{!}{
\begin{tabular}{l*{3}{r}|*{3}{c}|*{3}{c}}\toprule
\multirow{4}{*}{\textbf{Method}} & \multicolumn{1}{c}{\multirow{4}{*}{\rotatebox[origin=c]{90}{\textbf{\#~words}}}} & \multicolumn{1}{c}{\multirow{4}{*}{\rotatebox[origin=c]{90}{\textbf{\#~synsets}}}} & \multicolumn{1}{c|}{\multirow{4}{*}{\rotatebox[origin=c]{90}{\textbf{\#~pairs}}}} & \multicolumn{3}{c|}{} \\
\multicolumn{4}{c|}{} & \multicolumn{3}{c|}{\textbf{RuWordNet}} & \multicolumn{3}{c}{\textbf{YARN}} \\
\multicolumn{4}{c|}{} & \multicolumn{3}{c|}{} \\
\multicolumn{4}{c|}{} & \small\textbf{Pr} & \small\textbf{Re} & \small\textbf{F\textsubscript{1}} & \small\textbf{Pr} & \small\textbf{Re} & \small\textbf{F\textsubscript{1}}\\\midrule
\watset\S[CW\textsubscript{lin}, MCL] & $83{,}092$ & $58{,}353$ & $242{,}615$ &
$15.01$ & $\mathbf{32.55}$ & $\mathbf{20.55}^\ast$ &
$46.70$ & $\mathbf{42.69}$ & $\underline{\mathbf{44.61}}^\ast$ \\
\watset[CW\textsubscript{lin}, MCL] & $83{,}092$ & $55{,}369$ & $332{,}727$ &
$11.95$ & $\underline{\mathbf{34.91}}$ & $17.81^\ast$ &
$40.10$ & $\underline{\mathbf{46.32}}$ & $\mathbf{42.99}^\ast$ \\
MCL & $83{,}092$ & $21{,}973$ & $353{,}848$ &
$15.54$ & $29.10$ & $20.26^\ast$ &
$54.95$ & $33.94$ & $\mathbf{41.97}^\ast$ \\
CW\textsubscript{lin} & $83{,}092$ & $19{,}124$ & $672{,}076$ &
$8.73$ & $\mathbf{34.20}$ & $13.91^\ast$ &
$36.33$ & $\mathbf{45.13}$ & $40.25^\ast$ \\
\watset\S[MCL, CW\textsubscript{lin}] & $83{,}092$ & $62{,}700$ & $175{,}643$ &
$\mathbf{19.46}$ & $28.48$ & $\underline{\mathbf{23.12}}^\ast$ &
$52.28$ & $29.41$ & $37.65^\ast$ \\
MaxMax & $83{,}092$ & $27{,}011$ & $461{,}748$ &
$17.58$ & $26.09$ & $\mathbf{21.01}^\ast$ &
$\mathbf{58.24}$ & $19.49$ & $29.20^\ast$ \\
CPM{\textsubscript{$k\!=\!3$}} & $15{,}555$ & $4{,}000$ & $45{,}231$ &
$\mathbf{23.44}$ & $7.23$ & $11.05^\ast$ &
$\mathbf{62.51}$ & $6.04$ & $11.02^\ast$ \\
ECO & $83{,}092$ & $67{,}645$ & $18{,}362$ &
$\underline{\mathbf{72.41}}$ & $3.45$ & $6.58\hphantom{^\ast}$ &
$\underline{\mathbf{90.36}}$ & $0.18$ & $0.36\hphantom{^\ast}$ \\\bottomrule
\end{tabular}
}
\end{table}

The disambiguation of the input graph performed by the {\watset} method splits nodes belonging to several local communities to several nodes, significantly facilitating the clustering task otherwise complicated by the presence of the hubs that wrongly link semantically unrelated nodes. {\watset} robustly outperformed all other methods according to F\textsubscript{1}-score on all the datasets for English (\tablename~\ref{tab:watset:english}) and Russian (\tablename~\ref{tab:watset:russian}). In particular, on WordNet for English, \watset{[CW\textsubscript{log}, MCL]} has statistically significantly outperformed all other methods ($p \ll 0.01$), including different configurations of our algorithm. On BabelNet for English, \watset{[MCL, MCL]} showed a similar behavior ($p \ll 0.01$). On RuWordNet for Russian, Simplified \watset[MCL, CW\textsubscript{lin}] statistically significantly outperformed all other algorithms, including highly competitive MCL and MaxMax ($p \ll 0.01$). Similarly, on YARN for Russian, Simplified \watset[CW\textsubscript{lin}, MCL] has significantly outperformed all the other algorithms ($p \ll 0.01$).

Interestingly, in all the cases, the toughest competitor was a hard clustering algorithm---MCL \citep{vanDongen:00}. We observed that the ``plain'' MCL successfully groups monosemous words, but isolates the neighborhood of polysemous words, which results in the recall drop in comparison to {\watset}. CW operates faster due to a simplified update step. On the same graph, CW tends to produce larger clusters than MCL. This leads to a higher recall of ``plain'' CW as compared to the ``plain'' MCL, at the cost of lower precision. Although that MCL demonstrated highly competitive results, the best configuration of {\watset} has statistically significantly outperformed it on all the datasets.

Using MCL instead of CW for sense induction in {\watset} expectedly produced more fine-grained senses. However, at the global clustering step, these senses erroneously tend to form coarse-grained synsets connecting unrelated senses of the ambiguous words. This explains the generally higher recall of \watset{[MCL, $\cdot$]}. Despite the randomized nature of CW, variance across runs do not affect the overall ranking. The rank of different weighting schemes on the node degree of CW\textsubscript{top/lin/log} can change, while the rank of the best CW configuration compared to other methods remains the same.

The MaxMax algorithm showed mixed results. On the one hand, it outputs large clusters uniting more than a hundred nodes. This inevitably leads to a high recall, as it is clearly seen in the results for Russian because such synsets still pass under our cluster size threshold of 150 words. Its synsets on the English datasets are even larger and have been pruned, which resulted in the low recall. On the other hand, smaller synsets having at most 10--15 words were identified correctly. MaxMax appears to be extremely sensitive to edge weighting, which also complicates its application in practice.

The CPM algorithm showed unsatisfactory results, emitting giant components encompassing thousands of words. Such clusters were automatically pruned, but the remaining clusters are quite correct synsets, which is confirmed by the high precision values. When increasing the minimal number of elements in the clique $k$, recall improves, but at the cost of a dramatic precision drop. We suppose that the network structure assumptions exploited by CPM do not accurately model the structure of our synonymy graphs.

Finally, the ECO method yielded the worst results because most of the cluster candidates failed to pass through the constant threshold used for estimating whether a pair of words should be included in the same cluster. Most synsets produced by this method were trivial, i.e., containing only a single word. The remaining synsets for both languages have at most three words that have been connected by a chance due to the edge noising procedure used in this method, resulting in a low recall.

The results obtained on all gold standards (\figurename~\ref{fig:watset:weights}) show similar trends in terms of relative ranking of the methods. Yet absolute scores of YARN and RuWordNet are substantially different due to the inherent difference of these datasets. RuWordNet is more domain-specific in terms of vocabulary, so our input set of generic synonymy dictionaries has a limited coverage on this dataset. On the other hand, recall calculated on YARN is substantially higher as this resource was manually built on the basis of synonymy dictionaries used in our experiments.

\begin{table}[t]
\caption{\label{tab:watset:sample}Sample synsets induced by the \watset{[MCL, MCL]} method for English using the \textsf{sim} weighting approach.}
\centering
\begin{tabular}{cp{120mm}}\toprule
\textbf{Size} & \textbf{Synset}\\\midrule
2 & decimal point, dot \\
2 & wall socket, power point \\
3 & gullet, throat, food pipe \\
3 & CAT, computed axial tomography, CT \\
4 & microwave meal, ready meal, TV dinner, frozen dinner \\
4 & mock strawberry, false strawberry, gurbir, Indian strawberry \\
5 & objective case, accusative case, oblique case, object case, accusative \\
5 & discipline, sphere, area, domain, sector \\
6 & radio theater, dramatized audiobook, audio theater, radio play, radio drama, audio play \\
6 & integrator, reconciler, consolidator, mediator, harmonizer, uniter \\
7 & invite, motivate, entreat, ask for, incentify, ask out, encourage \\
7 & curtail, craw, yield, riding crop, harvest, crop, hunting crop \\
\bottomrule
\end{tabular}
\end{table}

\tablename~\ref{tab:watset:sample} presents examples of the obtained synsets of various sizes for the top {\watset} configuration on English. As one might observe, the quality of the results is highly plausible. Since in this configuration we assigned edge weights based on the cosine of the angle between Skip-Gram word vectors~\citep{Mikolov:13}, we should note that such an approach assigns high values of similarity not just to synonymous words, but to antonymous and generally any lexically related words. This is a common problem with lexical embeddings spaces which we tried to evade by explicitly using a synonymy dictionary as an input. For example, ``audio play'' and ``radio play'', or ``accusative'' and ``oblique'', are semantically related expressions, but really not synonyms. Such a problem can be addressed using techniques such as retrofitting~\citep{Faruqui:15} and contextualization~\citep{Peters:18}.

However, one limitation of all the approaches considered in this section is the dependence on the completeness of the input dictionary of synonyms. In some parts of the input synonymy graph, important bridges between words can be missing, leading to smaller-than-desired synsets. A promising extension of the present methodology is using distributional models to enhance connectivity of the graph by cautiously adding extra relationships~\citep{Ustalov:17:fighting}.

\paragraph{Cross-Resource Evaluation} In order to estimate the upper bound of precision, recall, and F\textsubscript{1}-score in our synset induction experiments, we conducted a cross-resource evaluation between the used gold standard datasets (\tablename~\ref{tab:watset:xres}). Similarly to the experimental setup described in Section~\ref{sub:synsets:setup}, we transformed synsets from every dataset into sets of synonymy pairs. Then, for every pair of gold standard datasets, we computed the pairwise precision, recall and F\textsubscript{1}-score by assessing synset-induced synonymy pairs of one dataset on the pairs of another dataset. As the result, we see that the low absolute numbers in evaluation are due to an inherent vocabulary mismatch between the input dictionaries of synonyms and the gold datasets since no single resource for Russian can obtain high recall scores on another one. Surprisingly, even BabelNet, which integrates most of the available lexical resources, still does not reach a recall substantially larger than 50\%.\footnote{We used BabelNet~3.7 extracting all 3,497,327 synsets that were marked as Russian.} Note that the results of this cross-dataset evaluation are not directly comparable to results in \tablename~\ref{tab:watset:russian} since in our experiments we use much smaller input dictionaries than those used by BabelNet. Our cross-resource evaluation demonstrates that unlike WordNet and BabelNet, which are built on a similar conceptual basis, RuWordNet and YARN have a very different structure, so an algorithm that shows good results on one will likely not perform very well on another.

\begin{table}[t]
\centering
\caption{\label{tab:watset:xres}Performance of lexical resources cross-evaluated against each other.}
\begin{tabular}{p{35mm}p{35mm}*{4}{c}}\toprule
\textbf{Input Synsets} & \textbf{Gold Synsets}  & \textbf{Language} & \textbf{Pr} & \textbf{Re} & \textbf{F\textsubscript{1}} \\\midrule
BabelNet  & WordNet    & \multirow{2}{*}{English} & $72.93$ & $99.76$ & $84.26$ \\
WordNet   & BabelNet   & & $99.79$ & $69.86$ & $82.18$ \\\midrule
YARN      & RuWordNet & \multirow{2}{*}{Russian} & $16.36$ & $16.21$ & $16.28$ \\
BabelNet  & RuWordNet & & $34.84$ & $40.87$ & $37.61$ \\\midrule
RuWordNet & YARN     & \multirow{2}{*}{Russian} & $66.96$ & $12.13$ & $20.54$ \\
BabelNet  & YARN     & & $51.53$ & $10.89$ & $17.98$ \\\bottomrule
\end{tabular}
\end{table}

\section{\label{sec:triframes}Application to Unsupervised Semantic Frame Induction}

\begin{figure}[t]
  \centering
  \includegraphics[width=\linewidth]{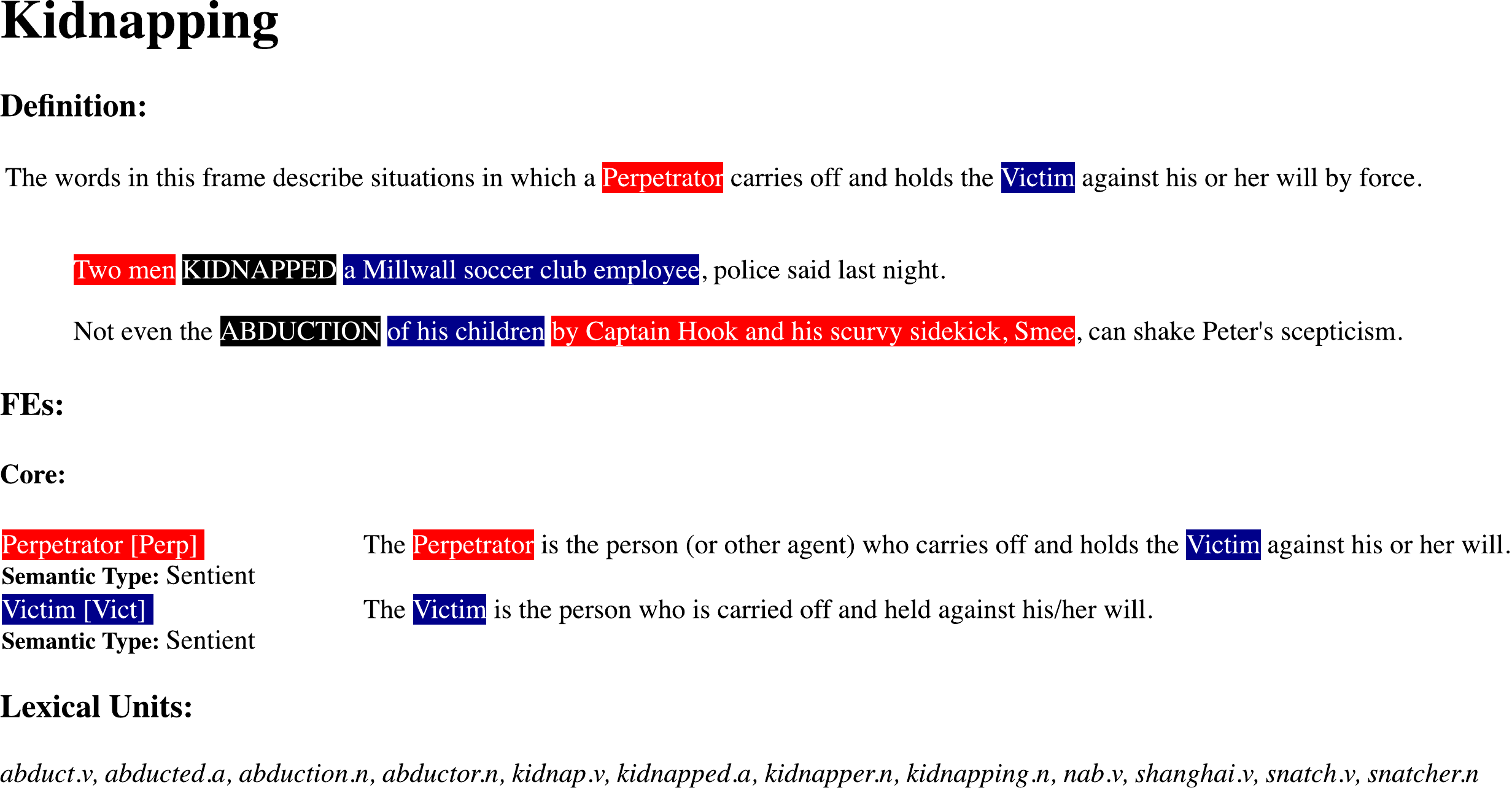}
  \caption{\label{fig:triframes:framenet} Definition, examples, core semantic roles, and frame invoking lexical units of the semantic frame ``Kidnapping'' from the FrameNet resource. }
\end{figure}

In this section, our goal is to investigate the applicability of our graph clustering technique in a different task. Namely, we explore how \textit{semantic frames}---more complex linguistic structures than synsets---can be induced from text using \watset. A \textit{semantic frame} is a central concept of the Frame Semantics theory~\citep{Fillmore:82}. A frame is a structure that describes certain situation or action, e.g., ``Dining'' or ``Kidnapping'', in terms of participants involved in these actions which fill semantic roles of this frame and words commonly describing such situations. \figurename~\ref{fig:triframes:framenet} illustrates a part of the ``Kidnapping'' semantic frame from the FrameNet resource.\footnote{\url{https://framenet.icsi.berkeley.edu/fndrupal/luIndex}}

Recent years have seen much work on Frame Semantics, enabled by the availability of a large set of frame definitions, as well as a manually annotated text corpus provided by the FrameNet project~\citep{Baker:98}. FrameNet data enabled the development of wide-coverage frame parsers using supervised learning \cite[\emph{inter alia}]{Gildea:02,Erk:06,Das:14}, as well as its application to a wide range of tasks, ranging from answer extraction in Question Answering~\citep{Shen:07} and Textual Entailment~\citep{Burchardt:09,BenAharon:10} to event-based predictions of stock markets~\citep{Xie:13}.

However, frame-semantic resources are arguably expensive and time-consuming to build due to difficulties in defining the frames, their granularity and domain. The complexity of the frame construction and annotation tasks requiring expertise in the underlying knowledge. Consequently, such resources exist only for a few languages~\citep{Boas:09} and even English is lacking domain-specific frame-based resources. Possible inroads are cross-lingual semantic annotation transfer~\citep{Pado:09,Hartmann:16} or linking FrameNet to other lexical-semantic or ontological resources~\citep[\emph{inter alia}]{Narayanan:03,Tonelli:09,Laparra:10,Gurevych:12}. But while the arguably simpler task of PropBank-based Semantic Role Labeling has been successfully addressed by unsupervised approaches~\citep{Lang:10,Titov:11}, fully unsupervised frame-based semantic annotation exhibits far more challenges, starting with the preliminary step of automatically inducing a set of semantic frame definitions that would drive a subsequent text annotation. We aim at overcoming these issues by automatizing the process of FrameNet construction through unsupervised frame induction techniques using {\watset}.

According to our statistics on the dependency-parsed FrameNet corpus of over 150 thousand sentences~\citep{Bauer:12}, the \texttt{SUBJ} and \texttt{OBJ} relationships are the two most common shortest paths between frame evoking elements (\textit{FEEs}) and their roles, accounting for 13.5\% of instances of a heavy-tail distribution of over 11 thousand different paths that occur three times or more in the FrameNet data. While this might seem a simplification that does not cover prepositional phrases and frames filling the roles of other frames in a nested fashion, we argue that the overall frame inventory can be induced on the basis of this restricted set of constructions, leaving other paths and more complex instances for further work. Thus, we expect the triples obtained from such a Web-scale corpus as DepCC~\citep{Panchenko:18:depcc} to cover most core arguments sufficiently. In contrast to the recent approaches like the one by \citet{Jauhar:17}, the approach we describe in this section induces semantic frames without any supervision, yet capturing only two core roles: the \textit{subject} and the \textit{object} of a frame triggered by \textit{verbal} predicates. Note that it is not generally correct to expect that the SVO triples obtained by a dependency parser are necessarily the core arguments of a predicate. Such roles can be implicit, i.e., unexpressed in a given context \citep{Schenk:16}, so additional syntactic relationships between frame elements could be taken into account~\citep{Kallmeyer:18}.

We cast the frame induction problem as a \textit{triclustering} task~\citep{Zhao:05,Ignatov:15}. Triclustering is a generalization of traditional clustering and biclustering problems~\citep[p.~144]{Mirkin:96}, aiming at simultaneously clustering objects along three dimensions, i.e., subject, verb and object in our case (cf.\ Table~\ref{tab:tricluster}). First, triclustering allows to avoid the prevalent pipelined architecture of frame induction approaches, e.g., the one by \citet{Kawahara:14}, where two independent clusterings are needed. Second, benchmarking frame induction as triclustering against other methods on dependency triples makes it possible to abstract away the evaluation of frame induction algorithms from other factors, e.g., the input corpus or pre-processing steps, thus allowing a fair comparison of different induction models.

\begin{table}[t]
\centering
\caption{\label{tab:tricluster}Example of a tricluster of lexical units corresponding to the \textcolor{ggplotframe}{``Kidnapping''} frame from FrameNet.}
\begin{tabular}{p{40mm}lp{70mm}}\toprule
\textbf{FrameNet} & \textbf{Role} & \textbf{Lexical Units (LU)} \\\midrule
Perpetrator & \textcolor{ggplotsubject}{Subject} & kidnapper, alien, militant \\
FEE         & \textcolor{ggplotverb}{Verb} & snatch, kidnap, abduct \\
Victim      & \textcolor{ggplotobject}{Object} & son, people, soldier, child \\\bottomrule
\end{tabular}
\end{table}

\subsection{Frame Induction as a Triclustering Task}

We focused on a simple setup for semantic frame induction using two roles and SVO triples, arguing that it still can be useful as frame roles are primarily expressed by subjects and objects, giving rise to semantic structures extracted in an unsupervised way with high coverage. Thus, given a vocabulary $V$ and a set of SVO triples $T \subseteq V^3$ from a syntactically analyzed corpus, our approach for frame induction, called Triframes, constructs a triple graph and clusters it using the {\watset} algorithm described in Section~\ref{sec:watset}.

Triframes reduces the frame induction problem to a simpler graph clustering problem. The algorithm has three steps: construction, clustering, and extraction. The triple graph \textit{construction} step, as described in Section~\ref{sub:triframes:graph}, uses a $d$-dimensional word embedding model $v \in V \to \vec{v} \in \mathbb{R}^d$ to embed triples in a dense vector space for establishing edges between them. The graph \textit{clustering} step, as described in Section~\ref{sub:triframes:clustering}, uses a clustering algorithm like {\watset} to obtain sets of triples corresponding to the instances of the semantic frames. The final, \textit{aggregation} step, as described in Section~\ref{sub:triframes:aggregation}, transforms the discovered triple clusters into frame-semantic representations. Triframes is parameterized by the number of nearest neighbors $k \in \mathbb{N}$ for establishing edges and a graph clustering algorithm $\mathrm{Cluster}$. The complete pseudocode of Triframes is presented in Algorithm~\ref{alg:triframes}.

\begin{algorithm}[t]
\caption{Unsupervised Semantic Frame Induction from Subject-Verb-Object Triples.}
\label{alg:triframes}
\begin{algorithmic}[1]
\REQUIRE{a set of SVO triples $T \subseteq V^3$,}
\REQUIREP{an embedding model $v \in V \to \vec{v} \in \mathbb{R}^d$,}
\REQUIREP{the number of nearest neighbors $k \in \mathbb{N}$,}
\REQUIREP{a graph clustering algorithm $\mathrm{Cluster}$.}
\ENSURE{a set of triframes $F$.}
\FORALL[Embed the triples]{\label{alg:triframes:triples:begin}$t = (s, p, o) \in T$}
\STATE{\label{alg:triframes:triples:end}$\vec{t} \gets \vec{s} \oplus \vec{p} \oplus \vec{o}$}
\ENDFOR
\STATE{\label{alg:triframes:graph:begin}$E \gets \{(t, t') \in T^2 : t' \in \NN_k(t), t \neq t'\}$}\COMMENT{Construct edges using nearest neighbors}
\STATE{\label{alg:triframes:graph:end}$G \gets (T, E)$}
\STATE{$F \gets \emptyset$}
\FORALL[Cluster the graph]{\label{alg:triframes:cluster}$C^i \in \mathrm{Cluster}(G)$}
\STATE{\label{alg:triframes:aggregate:begin}$f_s \gets \{s \in V : (s, v, o) \in C^i\}$}\COMMENT{Aggregate \textcolor{ggplotsubject}{subjects}}
\STATE{$f_v \gets \{v \in V : (s, v, o) \in C^i\}$}\COMMENT{Aggregate \textcolor{ggplotverb}{verbs}}
\STATE{\label{alg:triframes:aggregate:end}$f_o \gets \{o \in V : (s, v, o) \in C^i\}$}\COMMENT{Aggregate \textcolor{ggplotobject}{objects}}
\STATE{$F \gets F \cup \{(f_s, f_v, f_o)\}$}
\ENDFOR
\RETURN{$F$}
\end{algorithmic}
\end{algorithm}

\subsubsection{\label{sub:triframes:graph}SVO Triple Similarity Graph Construction} 
\begin{figure}[t]
  \centering
  \includegraphics[scale=.475]{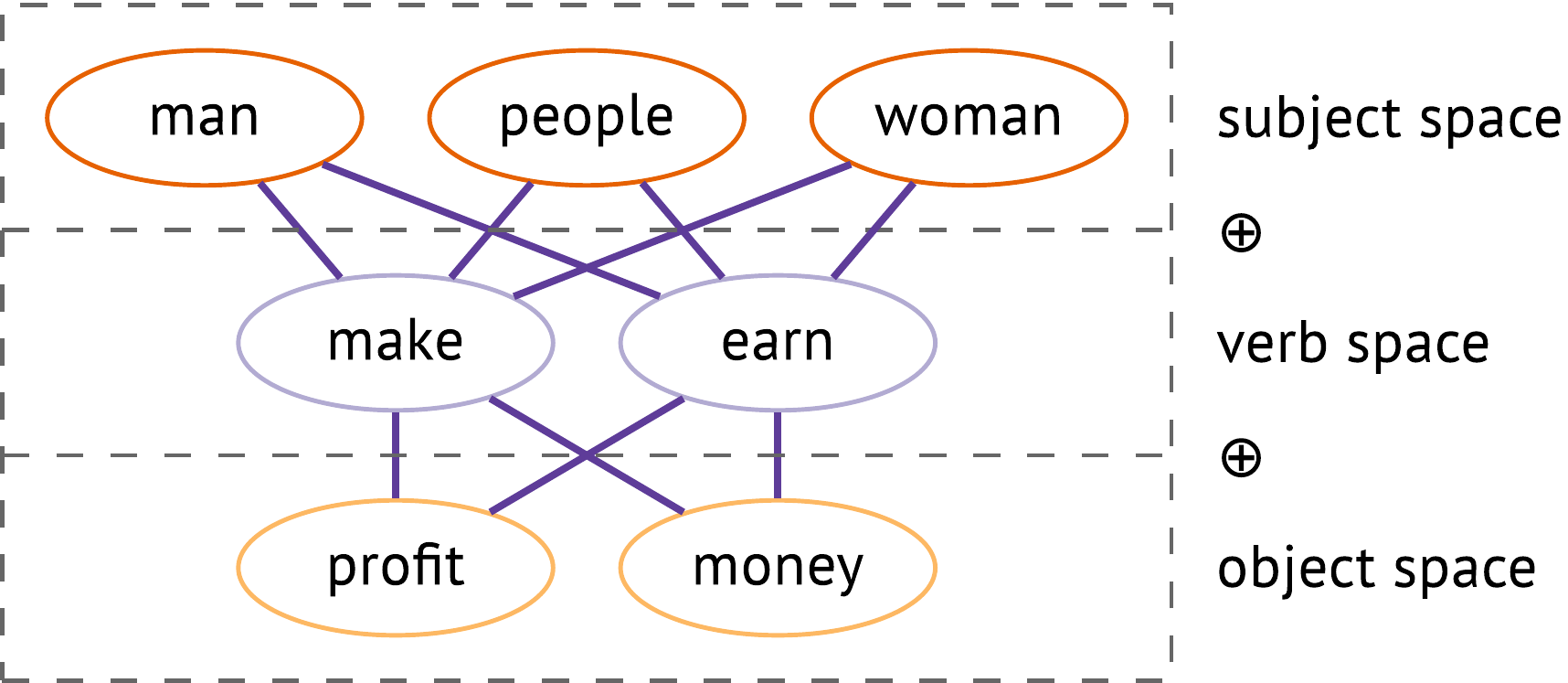}
  \caption{\label{fig:triframes:cat}Concatenation of the vectors corresponding to the triple elements, \textcolor{ggplotsubject}{subjects}, \textcolor{ggplotverb}{verbs}, and \textcolor{ggplotobject}{objects}, expresses the structural similarity of the triples.}
\end{figure}

We construct the triple graph ${G = (T, E)}$ in which the triples are connected to each other as according to the semantic similarity of their elements: subjects, verbs, objects. To express similarity, we embed the triples using distributional representations of words. In particular, we use a word embedding model to map every triple $t = (s, p, o) \in T$ to a $(3d)$-dimensional vector $\vec{t} = \vec{s} \oplus \vec{p} \oplus \vec{o}$ (lines~\ref{alg:triframes:triples:begin}--\ref{alg:triframes:triples:end}). Such a representation enables computing the distance between the triples in whole rather than between individual elements of them. The use of distributional models like Skip-Gram~\citep{Mikolov:13} makes it possible to take into account the contextual information of the whole triple. The concatenation of the vectors for words forming triples leads to the creation of a $(|T| \times 3d)$-dimensional vector space. \figurename~\ref{fig:triframes:cat} illustrates this idea: we expect structurally similar triples of different elements to be located in a dense vector space close to each other, while non-similar triples to be located far away to each other.

Given a triple $t \in T$, we denote the $k \in \mathbb{N}$ nearest neighbors extraction procedure of its concatenated embedding from the formed vector space as $\NN_k(t) \subseteq T \setminus \{t\}$. Then, we use the triple embeddings to generate the undirected graph $G = (T, E)$ by constructing the edge set $E \subseteq T^2$. For that, we retrieve $k$ nearest neighbors of each triple vector $\vec{t} \in \mathbb{R}^{3d}$ and establish cosine similarity-weighted edges between the corresponding triples. We establish edges only between the triples appearing in $k$ nearest neighbors (lines~\ref{alg:triframes:graph:begin}--\ref{alg:triframes:graph:end}):
\begin{equation}
  E = \{(t, t') \in T^2 : t' \in \NN_k(t)\}\text{.}
\end{equation}

As the result, the constructed triple graph $G$ has a clustered structure in which the clusters are sets of SVO triples representing the same frame.

\subsubsection{\label{sub:triframes:clustering}Similarity Graph Clustering} 
\begin{figure}[t]
  \centering
  \includegraphics[scale=.5]{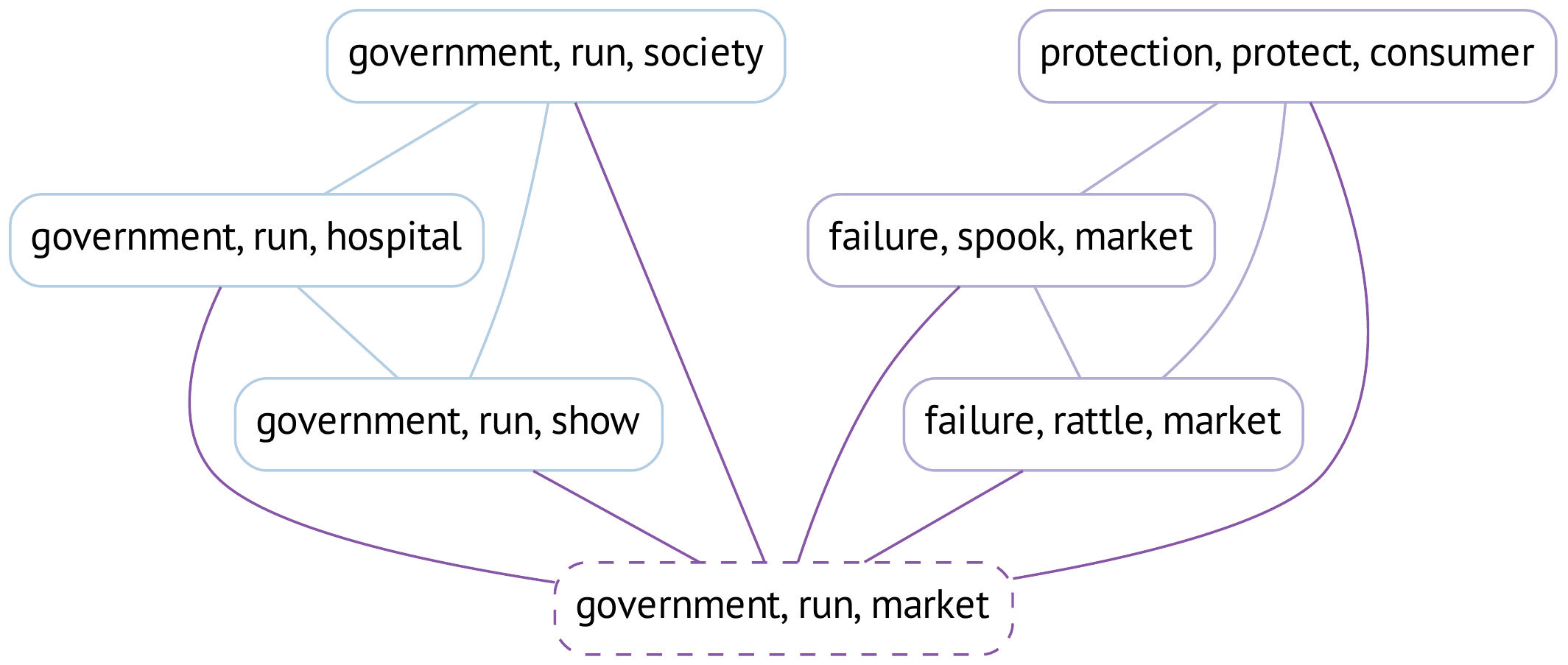}
  \caption{\label{fig:triframes:wsd}Example of two senses associated with a triple $(\textcolor{ggplotsubject}{\textit{government}}, \textcolor{ggplotverb}{\textit{run}}, \textcolor{ggplotobject}{\textit{market}})$.}
\end{figure}

We assume that the triples representing similar contexts fill similar roles, which is explicitly encoded by the concatenation of the corresponding vectors of the words constituting the triple (\figurename~\ref{fig:triframes:cat}). We use the {\watset} algorithm to obtain the clustering of the SVO triple graph $G$ (line~\ref{alg:triframes:cluster}). As described in Section~\ref{sec:watset}, our algorithm treats the SVO triples as the vertices $T$ of the input graph $G = (T, E)$, induces their senses (\figurename~\ref{fig:triframes:wsd}), and constructs an intermediate sense-aware representation that is clustered a hard clustering algorithm like Chinese Whispers~\citep{Biemann:06}. {\watset} is a suitable algorithm for this problem due to its performance on the related synset induction task (Section~\ref{sec:synsets}), its fuzzy nature, and the ability to find the number of frames automatically.

\subsubsection{\label{sub:triframes:aggregation}Aggregating Triframes} Finally, for each cluster $C^i \in C$, we aggregate the subjects, the verbs, and the objects of the contained triples into separate sets (lines~\ref{alg:triframes:aggregate:begin}--\ref{alg:triframes:aggregate:end}). As the result, each cluster is transformed into a \textit{triframe}, which is a triple that is composed of the subjects $f_s \subseteq V$, the verbs $f_v \subseteq V$, and the objects $f_o \subseteq V$. For example, the triples shown in \figurename~\ref{fig:triframes:cat} will form a triframe $(\{\textcolor{ggplotsubject}{\textit{man}}, \textcolor{ggplotsubject}{\textit{people}}, \textcolor{ggplotsubject}{\textit{woman}}\}, \{\textcolor{ggplotverb}{\textit{make}}, \textcolor{ggplotverb}{\textit{earn}}\}, \{\textcolor{ggplotobject}{\textit{profit}}, \textcolor{ggplotobject}{\textit{money}}\})$.

\subsection{Evaluation}

Currently, there is no universally accepted approach for evaluating unsupervised frame induction methods. All the previously developed methods were evaluated on completely different incomparable setups and used different input corpora~\cite[etc.]{Titov:12,Materna:13,OConnor:13}. We propose a unified methodology by treating the complex multi-stage frame induction task as a straightforward triple clustering task.

\subsubsection{\label{sub:triframes:setup}Experimental Setup} We compare our method, \textit{Triframes} {\watset}, to several available state-of-the-art baselines applicable to our dataset of triples (Section~\ref{sub:frames}). \textit{LDA-Frames} by \citet{Materna:12,Materna:13} is a frame induction method based on topic modeling. \textit{Higher-Order Skip-Gram (HOSG)} by~\citet{Cotterell:17} generalizes the Skip-Gram model \citep{Mikolov:13} by extending it from word-context co-occurrence matrices to tensors factorized with a polyadic decomposition. In our case, this tensor consisted of SVO triple counts. \textit{NOAC} by \citet{Egurnov:17} is an extension of the Object-Attribute-Condition (\textit{OAC}) triclustering algorithm by \citet{Ignatov:15} to numerically weighted triples. This incremental algorithm searches for dense regions in triadic data. Also, we use five simple baselines. In the \textit{Triadic} baselines, independent word embeddings of subject, object, and verb are concatenated and then clustered using $k$-Means~\citep{Hartigan:79} and spectral clustering~\citep{Shi:00}. In \textit{Triframes CW}, instead of {\watset}, we use Chinese Whispers (CW), a \textit{hard} graph clustering algorithm~\citep{Biemann:06}. We also evaluate the performance of Simplified {\watset} (Section~\ref{sub:simplified}). Finally, two trivial baselines are \textit{Singletons} that creates a single cluster per instance and \textit{Whole} that creates one cluster for all elements.

\paragraph{Quality Measure} Following the approach for verb class evaluation by \citet{Kawahara:14}, we employ \textit{normalized modified purity} ($\nmpu$) and \textit{normalized inverse purity} ($\nipu$) as the quality measures for overlapping clusterings. Given the clustering $C$ and the gold clustering $C_G$, normalized modified purity quantifies the clustering precision as the average of the weighted overlap $\delta_{C^i}(C^i \cap C^j_G)$ between each cluster $C^i \in C$ and the gold cluster $C^i_G \in C_G$ that maximizes the overlap with $C^i$:
\begin{equation}
  \nmpu = \frac{1}{|C|} \sum^{|C|}_{i \in \mathbb{N} : |C^i| > 1} \max_{1 \leq j \leq |C_G|} \delta_{C^i}(C^i \cap C^j_G)\text{,}
\end{equation}
where the weighted overlap is the sum of the weights $C^{i,v}$ for each word $v \in C^i$ in $i$-th cluster: $\delta_{C^i}(C^i \cap C^j_G) = \sum_{v \in C^i \cap C^j_G} C^{i,v}$. Note that $\nmpu$ counts all the singleton clusters as wrong. Similarly, normalized inverse purity (collocation) quantifies the clustering recall:
\begin{equation}
  \nipu = \frac{1}{|C_G|} \sum^{|G|}_{j = 1} \max_{1 \leq i \leq |C|} \delta_{C^j_G}(C^i \cap C^j_G)\text{.}
\end{equation}
Then, $\nmpu$ and $\nipu$ are combined together as the harmonic mean to yield the overall clustering F\textsubscript{1}-score computed as $\mathrm{F}_1 = 2\frac{\nmpu \cdot \nipu}{\nmpu + \nipu}$, which we use to rank the approaches.

Our framework can be extended to the evaluation of more than two roles by generating more roles per frame. Currently, given a set of gold triples generated from the FrameNet, each triple element has a role, e.g., ``\textit{Victim}'', ``\textit{Predator}'', and ``\textit{FEE}''. We use a fuzzy clustering evaluation measure that operates not on triples, but instead on a set of tuples. Consider for instance a gold triple {$(\text{Freddy}\colon \textit{Predator}, \text{kidnap}\colon \textit{FEE}, \text{kid}\colon \textit{Victim})$}. It will be converted to three pairs {$(\text{Freddy}, \textit{Predator})$}, {$(\text{kidnap}, \textit{FEE})$}, {$(\text{kid}, \textit{Victim})$}. Each cluster in both $C$ and $C_G$ is transformed into a union of all constituent typed pairs. The quality measures are finally calculated between these two sets of tuples corresponding to $C$ and $C_G$. Note that one can easily pull in more than two core roles by adding to this gold standard set of tuples other roles of the frame, e.g., $\{(\text{forest}, \textit{Location})\}$. In our experiments, we focused on two main roles as our contribution is related to the application of triclustering methods. However, if more advanced methods of clustering are used, yielding clusters of arbitrary modality ($n$-clustering), one could also use our evaluation scheme.

\paragraph{Statistical Testing} Since that the normalization term of the quality measures used in this experiment does not allow us to compute a contingency table, we cannot directly apply the McNemar's test or a location test to evaluate the statistical significance of the results as we did in our synset induction experiment (Section~\ref{sub:synsets:setup}). Thus, we have applied a bootstrapping approach for statistical significance evaluation as follows. Given a set of clusters $C$ and a set of gold standard clusters $C_G$, we bootstrap an $N$-sized distribution of F\textsubscript{1}-scores. On each iteration, we take a sample $C'$ with replacements of $|C|$ elements from $C$. Then, we compute $\nmpu$, $\nipu$ and F\textsubscript{1} on $C'$ against the gold standard clustering $C_G$. Finally, for each pair of compared algorithms we use a two-tailed $t$-test~\citep{Welch:47} from the Apache Commons Math library\footnote{\url{https://commons.apache.org/proper/commons-math/}} to assess the significance in the difference in means between the corresponding bootstrap F\textsubscript{1}-score distributions. Thus, we consider than performance of one algorithm to be higher than the performance of another if both the $p$-value of the $t$-test is smaller than the significance level of $0.01$ and the mean bootstrap F\textsubscript{1}-score of the first method is larger than of the second. Due to a high computational complexity of bootstrapping~\citep{Dror:18}, we had to limit the value of $N$ to $5000$ in the frame induction experiment and to $10{,}000$ in the verb clustering experiment.

\paragraph{Gold Standard Datasets} We constructed a gold standard set of triclusters. Each tricluster corresponds to a FrameNet frame, similarly to the one illustrated in Table~\ref{tab:tricluster}. We extracted frame annotations from the over 150 thousand sentences from FrameNet~1.7 \citep{Baker:98}. We used the frame, FEE, and arguments labels in this dataset to generate triples in the form $(\text{word}_i\colon \textit{role}_1, \text{word}_j\colon \textit{FEE}, \text{word}_k\colon \textit{role}_2)$, where $\text{word}_{i/j/k}$ correspond to the roles and FEE in the sentence. We omitted roles expressed by multiple words as we use dependency parses, where one node represents a single word only.

For the sentences where more than two roles are present, all possible triples were generated. For instance, consider the sentence ``Two \textit{men} \textit{kidnapped} a soccer club \textit{employee} at the train \textit{station}.'', where ``men'' has a semantic role of \textit{Perpetrator}, ``employee'' has a semantic role of \textit{Victim}, ``station'' has the semantic role of \textit{Place}, and the word ``kidnapped'' is a frame-evoking lexical element (see \figurename~\ref{fig:triframes:framenet}). In this sentence containing three semantic roles, the following triples will be generated: (\textcolor{ggplotsubject}{men}: \textit{Perpetrator}, \textcolor{ggplotverb}{kidnap}: \textit{FEE}, \textcolor{ggplotobject}{employee}: \textit{Victim}), (\textcolor{ggplotsubject}{men}: \textit{Perpetrator}, \textcolor{ggplotverb}{kidnap}: \textit{FEE}, \textcolor{ggplotobject}{station}: \textit{Place}), (\textcolor{ggplotsubject}{employee}: \textit{Victim}, \textcolor{ggplotverb}{kidnap}: \textit{FEE}, \textcolor{ggplotobject}{station}: \textit{Place}). Sentences with less than two semantic roles were not considered. Finally, for each frame, we selected only two roles, which are the most frequently co-occurring in the FrameNet annotated texts. This has left us with about $10^5$ instances for the evaluation. For the evaluation purposes, we operate on the intersection of triples from DepCC and FrameNet. Experimenting on the full set of DepCC triples is only possible for several methods that scale well (\watset, CW, $k$-Means), but is prohibitively expensive for other methods (LDA-Frames, NOAC) because of the input data size combined with the complexity of these algorithms. During prototyping, we found that removing the triples containing pronouns from both the input and the gold standard dataset dramatically reduces the number of instances without the change of the ranks in the evaluation results. Thus, we decided to perform our experiments on the whole dataset without such a filtering.

In addition to the frame induction evaluation, where subjects, objects, and verbs are evaluated together, we also used a dataset of polysemous verb classes introduced by~\citet{Korhonen:03} and employed by~\citet{Kawahara:14}. Statistics of both datasets are summarized in Table~\ref{tab:triframes:datasets}. Note that the polysemous verb dataset is rather small, whereas the FrameNet triples set is fairly large, enabling reliable comparisons.

\paragraph{Input Data} In our evaluation, we use subject-verb-object triples from the DepCC dataset~\citep{Panchenko:18:depcc},\footnote{\url{https://www.inf.uni-hamburg.de/en/inst/ab/lt/resources/data/depcc.html}} which is a dependency-parsed version of the Common Crawl corpus, and the standard 300-dimensional Skip-Gram word embedding model trained on Google News corpus~\citep{Mikolov:13}. All the evaluated algorithms are executed on the same set of triples, eliminating variations due to different corpora or pre-processing.

\begin{table}[t]
\centering
\caption{\label{tab:triframes:datasets}Statistics of the evaluation datasets.}
\resizebox{1.0\linewidth}{!}{
\begin{tabular}{lrrr}\toprule
\textbf{Dataset} & \textbf{\# instances} & \textbf{\# unique} & \textbf{\# clusters} \\\midrule
FrameNet Triples (Bauer et al. \citeyear{Bauer:12}) & $99{,}744$ & $94{,}170$ & $383$ \\
Polysemous Verb Classes (Korhonen et al.~\citeyear{Korhonen:03}) & $246$ &    $110$ & $62$ \\\bottomrule
\end{tabular}
}
\end{table}

\subsubsection{Parameter Tuning} We tested various hyper-parameters of each of these algorithms and report the best results overall per frame induction algorithm. We run $500$ iterations of the LDA-Frames model with the default parameters~\citep{Materna:13}. For Higher-Order Skip-Gram (HOSG) by \citet{Cotterell:17}, we trained three vector arrays (for subjects, verbs, and objects) on the $108{,}073$ SVO triples from the \textit{FrameNet} corpus, using the implementation provided by the authors.\footnote{\url{https://github.com/azpoliak/skip-gram-tensor}} Training was performed with $5$ negative samples, $300$-dimensional vectors, and $10$ epochs. We constructed an embedding of a triple by concatenating embeddings for subjects, verbs, and objects, and clustered them using $k$-Means with the number of clusters set to $10{,}000$ (this value provided the best performance). We tested several configurations of the NOAC method by \citet{Egurnov:17} varying the minimum density of the cluster: the density of 0.25 led to the best results. For our Triframes method, we tried different values of $k \in \{5, 10, 30, 100\}$, while the best results were obtained on $k=30$ for both Triframes {\watset} and CW. The both \textit{Triadic} baselines shown the best results on $k=500$.

\subsubsection{Results and Discussion} We perform two experiments to evaluate our approach: (1) a frame induction experiment on the FrameNet annotated corpus by~\citet{Bauer:12}; (2) the polysemous verb clustering experiment on the dataset by~\citet{Korhonen:03}. The first is based on the newly introduced frame induction evaluation scheme (cf. Section~\ref{sub:triframes:setup}). The second one evaluates the quality of verb clusters only on a standard dataset from prior work.

\begin{figure}[t]
  \centering
  \includegraphics[scale=.75]{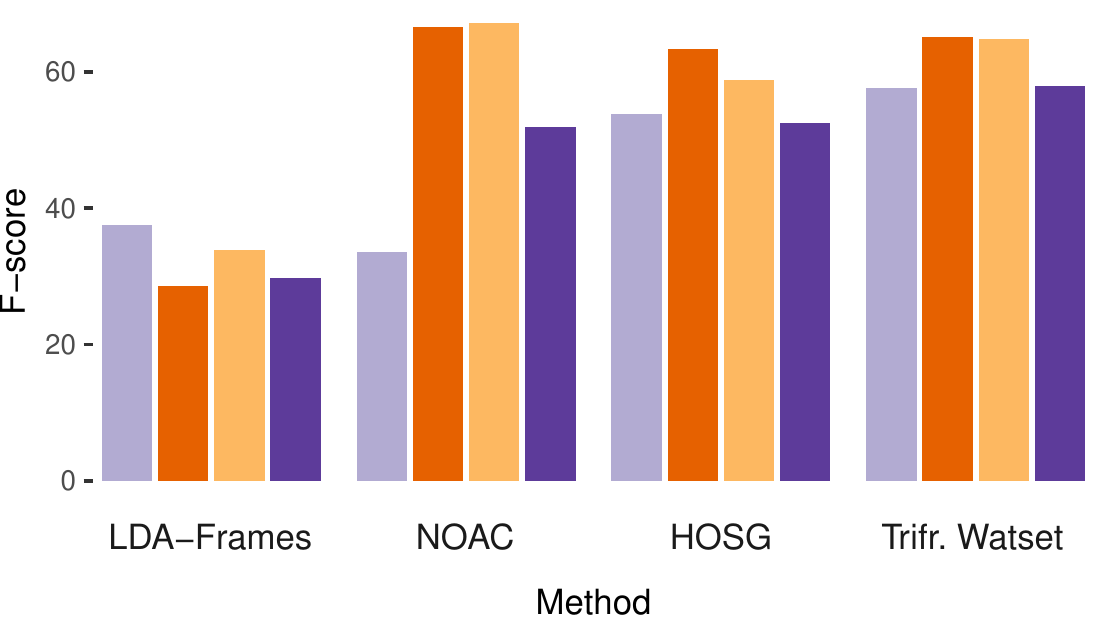} 
  \\{\footnotesize\textsf{Element: \legend{ggplotverb}\,verb, \legend{ggplotsubject}\,subject, \legend{ggplotobject}\,object, \legend{ggplotframe}\,frame}.}
  \caption{\label{fig:frames}F\textsubscript{1}-score values measured on the FrameNet Corpus~\citep{Bauer:12}. Each block corresponds to the top performance of the method in \tablename~\ref{tab:frames}.}
\end{figure}

\paragraph{Frame Induction Experiment} In \tablename~\ref{tab:frames} and \figurename~\ref{fig:frames}, the results of the experiment are presented. Triframes based on {\watset} clustering outperformed the other methods on both Verb F\textsubscript{1} and overall Frame F\textsubscript{1}. The \textit{HOSG}-based clustering proved to be the most competitive baseline, yielding decent scores according to all four measures. The \textit{NOAC} approach captured the frame grouping of slot fillers well but failed to establish good verb clusters. Note that \textit{NOAC} and \textit{HOSG} use only the graph of syntactic triples and do not rely on pre-trained word embeddings. This suggests a high complementarity of signals based on distributional similarity and global structure of the triple graph. Finally, the simpler \textit{Triadic} baselines relying on hard clustering algorithms showed low performance, similar to that of \textit{LDA-Frames}, justifying the more elaborate {\watset} method. Although we, due to the computational reasons (Section~\ref{sub:triframes:setup}), have statistically evaluated only Frame F\textsubscript{1} results, we found all the results but HOSG to be statistically significant ($p \ll 0.01$).

\begin{table}[t]
\caption{\label{tab:frames}Frame evaluation results on the triples from the FrameNet~1.7 corpus~\citep{Baker:98}. The results are sorted by the descending order of the Frame F\textsubscript{1}-score. Best results are boldfaced and statistically significant results are marked with an asterisk ($^\ast$). Simplified {\watset} is denoted as {\watset\S}.}
\resizebox{1.0\linewidth}{!}{
\centering
\begin{tabular}{l*{4}{|*{3}{c}}}\toprule
\multirow{2}{*}{\textbf{Method}} & \multicolumn{3}{c|}{\textbf{Verb}}
& \multicolumn{3}{c|}{\textbf{Subject}}
& \multicolumn{3}{c|}{\textbf{Object}}
& \multicolumn{3}{c}{\textbf{Frame}} \\
& \small\textbf{nmPU} & \small\textbf{niPU} & \small\textbf{F\textsubscript{1}}
& \small\textbf{nmPU} & \small\textbf{niPU} & \small\textbf{F\textsubscript{1}}
& \small\textbf{nmPU} & \small\textbf{niPU} & \small\textbf{F\textsubscript{1}}
& \small\textbf{nmPU} & \small\textbf{niPU} & \small\textbf{F\textsubscript{1}} \\\midrule
Triframes \watset[CW\textsubscript{top}, CW\textsubscript{top}] & $42.84$ & $88.35$ & $57.70$ & $54.22$ & $81.40$ & $65.09$ & $53.04$ & $83.25$ & $64.80$ & $55.19$ & $60.81$ & $\mathbf{57.87}^\ast$ \\
Triframes \watset\S[CW\textsubscript{top}, CW\textsubscript{top}] & $42.70$ & $87.41$ & $57.37$ & $54.29$ & $78.92$ & $64.33$ & $52.87$ & $83.47$ & $64.74$ & $55.12$ & $59.92$ & $57.42^\ast$ \\
Triframes \watset[MCL, MCL] & $52.60$ & $70.07$ & $\mathbf{60.09}$ & $55.70$ & $74.51$ & $63.74$ & $54.14$ & $78.70$ & $64.15$ & $60.93$ & $52.44$ & $56.37^\ast$ \\

Triframes \watset\S[MCL, MCL] & $55.13$ & $69.58$ & $61.51$ & $55.10$ & $76.02$ & $63.89$ & $54.27$ & $78.48$ & $64.17$ & $60.56$ & $52.16$ & $56.05^\ast$ \\
HOSG \citep{Cotterell:17} &    $44.41$ & $68.43$ & $53.86$ & $52.84$ & $74.53$ & $61.83$ & $54.73$ & $74.05$ &    $62.94$ & $55.74$ & $50.45$ & $52.96\hphantom{^\ast}$ \\
NOAC (Egurnov et al. \citeyear{Egurnov:17}) & $20.73$ & $88.38$ & $33.58$ & $57.00$ & $80.11$ & $\mathbf{66.61}$ & $57.32$ & $81.13$ & $\mathbf{67.18}$ & $44.01$ & $63.21$ & $51.89^\ast$ \\
Triadic Spectral    & $49.62$ & $24.90$ & $33.15$ & $50.07$ & $41.07$ & $45.13$ & $50.50$ & $41.82$ & $45.75$ & $52.05$ & $28.60$ & $36.91^\ast$ \\
Triadic $k$-Means   & $\mathbf{63.87}$ & $23.16$ & $33.99$ & $\mathbf{63.15}$ & $38.20$ & $47.60$ & $\mathbf{63.98}$ & $37.43$ & $47.23$ & $\mathbf{63.64}$ & $24.11$ & $34.97^\ast$ \\
LDA-Frames \citep{Materna:13} & $26.11$ & $66.92$ & $37.56$ & $17.28$ & $83.26$ & $28.62$ & $20.80$ & $90.33$ & $33.81$ & $18.80$ & $71.17$ & $29.75^\ast$ \\
Triframes CW        &  $7.75$ &  $6.48$ &  $7.06$ &  $3.70$ & $14.07$ &  $5.86$ & $51.91$ & $76.92$ & $61.99$ & $21.67$ & $26.50$ & $23.84\hphantom{^\ast}$ \\\midrule
Singletons          &     $0$ & $18.03$ &     $0$ &     $0$ & $20.56$ &     $0$ &     $0$ & $17.35$ &     $0$ & $81.44$ & $15.50$ & $26.04\hphantom{^\ast}$ \\
Whole               &  $7.35$ & $\mathbf{100.0}$ & $13.70$ &  $5.62$ & $\mathbf{97.40}$ & $10.63$ &  $4.24$ & $\mathbf{98.01}$ &  $8.14$ &  $5.07$ & $\mathbf{98.75}$ &  $9.65\hphantom{^\ast}$ \\\bottomrule
\end{tabular}
}
\end{table}

While triples are intuitively less ambiguous than words, still some frequent and generic triples like $(\textcolor{ggplotsubject}{\text{she}}, \textcolor{ggplotverb}{\text{make}}, \textcolor{ggplotobject}{\text{it}})$ can act as hubs in the graph, making it difficult to split it into semantically plausible clusters. The poor results of the Chinese Whispers hard clustering algorithm illustrate this. Since the hubs are ambiguous, i.e., can belong to multiple clusters, the use of the {\watset} fuzzy clustering algorithm that splits the hubs by disambiguating them leads to the best results (see Table~\ref{tab:frames}). We found that in average, {\watset} tends to create smaller clusters than its closest competitors, HOSG and NOAC. For instance, an average frame produced by Triframes~\watset{[CW\textsubscript{top}, CW\textsubscript{top}]} has $2.87 \pm 4.60$ subjects, $3.77 \pm 16.31$ verbs, and $3.27 \pm 6.31$ objects. NOAC produced in average $8.95 \pm 15.05$ subjects, $133.94 \pm 227.60$ verbs, and $15.17 \pm 18.37$ objects per frame. HOSG produced in average $3.00 \pm 4.20$ subjects, $6.49 \pm 12.15$ verbs, and $2.81 \pm 4.89$ objects per frame. We conclude that {\watset} was producing smaller clusters in general, which appear to be meaningful yet insufficiently coarse-grained as according to the used gold standard verb dataset.

\paragraph{Verb Clustering Experiment} \tablename~\ref{tab:verbs} presents the evaluation results on the second dataset for the best models identified on the first dataset. The \textit{LDA-Frames} yielded the best results with our approach performing comparably in terms of the F\textsubscript{1}-score. We attribute the low performance of the Triframes method based on CW clustering (Triframes~CW) to its hard partitioning output, whereas the evaluation dataset contains fuzzy clusters. The simplified version of {\watset} has statistically significantly outperformed all other approaches. Although the LDA-Frames algorithm showed the higher value of F\textsubscript{1} than the original version of {\watset} in this experiment, we found that its sampled F\textsubscript{1}-score is $44.98 \pm 0.04$, while Triframes \watset{[CW\textsubscript{top}, CW\textsubscript{top}]} showed $47.88 \pm 0.01$. Thus, we infer that our method has demonstrated non-significantly lower performance on this verb clustering task. In turn, the NOAC approach showed significantly worse results than both LDA-Frames and our approach ($p \ll 0.01$). Different rankings in Tables~\ref{tab:frames} and \ref{tab:verbs} also suggest that frame induction cannot simply be treated as a verb clustering and requires a separate task.

\begin{table}[t]
\centering
\caption{\label{tab:verbs}Evaluation results on the dataset of polysemous verb classes by \citet{Korhonen:03}. The results are sorted by the descending order of F\textsubscript{1}-score. Best results are boldfaced and statistically significant results are marked with an asterisk ($^\ast$). Simplified {\watset} is denoted as {\watset\S}.}
\begin{tabular}{p{90mm}*{3}{c}}\toprule
\textbf{Method}     & \textbf{nmPU} &  \textbf{niPU} & \textbf{F\textsubscript{1}} \\\midrule
Triframes \watset\S[CW\textsubscript{top}, CW\textsubscript{top}] & $41.21$ & $62.82$ & $\mathbf{49.77}^\ast$ \\
LDA-Frames \citep{Materna:13} & $\mathbf{52.60}$ & $45.84$ & $48.98\hphantom{^\ast}$ \\
Triframes \watset[CW\textsubscript{top}, CW\textsubscript{top}] & $40.05$ & $62.09$ & $48.69^\ast$ \\
NOAC (Egurnov et al. \citeyear{Egurnov:17}) & $36.43$ & $63.68$ & $46.35^\ast$ \\
Triframes \watset[MCL, MCL] & $39.26$ & $54.92$ & $45.78^\ast$ \\
Triframes \watset\S[MCL, MCL] & $36.31$ & $53.81$ & $43.36^\ast$ \\
Triadic Spectral    & $45.70$ & $38.96$ & $42.06\hphantom{^\ast}$ \\
HOSG \citep{Cotterell:17} & $38.22$ & $43.76$ & $40.80^\ast$ \\
Triadic $k$-Means   & $46.76$ & $28.92$ & $35.74^\ast$ \\
Triframes CW        & $18.05$ & $12.72$ & $14.92\hphantom{^\ast}$ \\\midrule
Whole               & $24.14$ & $\mathbf{79.09}$ & $36.99\hphantom{^\ast}$ \\
Singletons          &  $0$    & $27.21$ &  $0\hphantom{^\ast}$ \\\bottomrule
\end{tabular}
\end{table}

\begin{figure}[t]
  \centering\hyphenpenalty=10000
  \begin{tabular}{cp{15mm}p{105mm}}
  \multirow{5}{*}{\rotatebox[origin=c]{90}{\textcolor{ggplotframe}{\textbf{\#~1268}}}} & \textbf{\textcolor{ggplotsubject}{Subjects:}} & expert, scientist, lecturer, engineer, analyst \\
  & \textbf{\textcolor{ggplotverb}{Verbs:}} & study, examine, tell, detect, investigate, do, observe, hold, find, have, predict, claim, notice, give, discover, explore, learn, monitor, check, recognize, demand, look, call, engage, spot, inspect, ask \\
  & \textbf{\textcolor{ggplotobject}{Objects:}} & view, problem, gas, area, change, market \\\midrule
  \multirow{4}{*}{\rotatebox[origin=c]{90}{\textcolor{ggplotframe}{\textbf{\#~1378}}}} & \textbf{\textcolor{ggplotsubject}{Subjects:}} & leader, officer, khan, president, government, member, minister, chief, chairman \\
  & \textbf{\textcolor{ggplotverb}{Verbs:}} & belong, run, head, spearhead, lead \\
  & \textbf{\textcolor{ggplotobject}{Objects:}} & party, people \\\midrule
  \multirow{3}{*}{\rotatebox[origin=c]{90}{\textcolor{ggplotframe}{\textbf{\#~4211}}}} & \textbf{\textcolor{ggplotsubject}{Subjects:}} & evidence, research, report, survey \\
  & \textbf{\textcolor{ggplotverb}{Verbs:}} & prove, reveal, tell, show, suggest, confirm, indicate, demonstrate \\
  & \textbf{\textcolor{ggplotobject}{Objects:}} & method, evidence \\
  \end{tabular}
  \caption{\label{fig:triframes:good}Examples of ``good'' frames produced by the Triframes \watset[CW\textsubscript{top}, CW\textsubscript{top}] method as labeled by our annotators; frame identifiers are present in the first column, pronouns and prepositions are omitted.}
\end{figure}
\begin{figure}[t]
  \centering\hyphenpenalty=10000
  \begin{tabular}{cp{15mm}p{105mm}}
  \multirow{4}{*}{\rotatebox[origin=c]{90}{\textcolor{ggplotframe}{\textbf{\#~8}}}} & \textbf{\textcolor{ggplotsubject}{Subjects:}} & wine, act, power \\
  & \textbf{\textcolor{ggplotverb}{Verbs:}} & hearten, bring, discourage, encumber, \mbox{\dots\textit{432 more verbs}\dots}, build, chew, unsettle, snap \\
  & \textbf{\textcolor{ggplotobject}{Objects:}} & right, good, school, there, thousand \\\midrule
  \multirow{3}{*}{\rotatebox[origin=c]{90}{\textcolor{ggplotframe}{\textbf{\#~1057}}}} & \textbf{\textcolor{ggplotsubject}{Subjects:}} & parent, scientist, officer, event \\
  & \textbf{\textcolor{ggplotverb}{Verbs:}} & promise, pledge \\
  & \textbf{\textcolor{ggplotobject}{Objects:}} & parent, be, good, government, client, minister, people, coach \\\midrule
  \multirow{3}{*}{\rotatebox[origin=c]{90}{\textcolor{ggplotframe}{\textbf{\#~1657}}}} & \textbf{\textcolor{ggplotsubject}{Subjects:}} & people, doctor \\
  & \textbf{\textcolor{ggplotverb}{Verbs:}} & spell, steal, tell, say, know \\
  & \textbf{\textcolor{ggplotobject}{Objects:}} & egg, food, potato \\
  \end{tabular}
  \caption{\label{fig:triframes:bad}Examples of ``bad'' frames produced by the Triframes \watset[CW\textsubscript{top}, CW\textsubscript{top}] method as labeled by our annotators; frame identifiers are present in the first column, pronouns and prepositions are omitted.}
\end{figure}

\paragraph{Manual Evaluation of the Induced Frames} In addition to the to experiments based on gold standard lexical resources, we also performed a  manual evaluation. In particular, we assessed the quality of the frames produced by the Triframes \watset[CW\textsubscript{top}, CW\textsubscript{top}] approach using $n=30$ nearest neighbors for constructing a triple graph, which showed the best performance during automatic evaluation (Tables~\ref{tab:frames} and \ref{tab:verbs}).

To prepare the data for a manual annotation, we sampled 100 random frames and manually annotated them with three different annotators. For the convenience of the annotators, before drawing a sample we removed pronouns and prepositions from the frame elements while keeping them containing at least two different lexical units. This is to remove rather meaningful triples, e.g., $(\textcolor{ggplotsubject}{\text{her}}, \textcolor{ggplotverb}{\text{make}}, \textcolor{ggplotobject}{\text{it}})$, which are however present in large amounts in the FrameNet gold standard dataset.

In this study, annotators were instructed to annotate a frame as ``good'' if its elements (SVO) generally make sense together and each element is a reasonable set of lexical units. In total, the annotators judged 63 frames out of 100 to be good with a Fleiss'~(\citeyear{Fleiss:71}) $\kappa$ agreement of $0.816$.\footnote{We used the DKPro~Agreement tookit by \citet{Meyer:14} to compute the inter-annotator agreement.} While this is a rather general definition, the high agreement rate seems to suggest that it still provides a meaningful definition shared across annotators. \figurename~\ref{fig:triframes:good} presents examples of ``good'' frames, i.e., those which are labeled as semantically plausible by our annotators. \figurename~\ref{fig:triframes:bad} shows examples of ``bad'' frames according to the same criteria. These frames are available for download.\footnote{The examples are from the file {\small\texttt{triw2v-watset-n30-top-top-triples.txt}} is available in the ``Downloads'' section of our GitHub repository at \url{https://github.com/uhh-lt/triframes}.}

\section{\label{sec:classes}Application to Unsupervised Distributional Semantic Class Induction}

In this section, we investigate the applicability of our graph clustering technique in another unsupervised resource induction task. The first two experiments investigated the acquisition of two linguistic symbolic structures from two different types of graphs -- namely, synsets induced from graph of synonyms (Section~\ref{sec:synsets}) and semantic frames induced from graphs of distributionally-related syntactic triples (Section~\ref{sec:triframes}). In this section, we show how {\watset} can be used to induce a third type of structures, namely \textit{semantic classes} from a graph of distributionally-related words, also known as a \textit{distributional thesaurus} (or DT), see~\citep{Lin:98,Biemann:13}. In the context of this article,  \textit{semantic classes} will be considered as semantically plausible groups of words or word senses that have some common semantic feature.

The following sections will provide details of this experiment. In particular, Section~\ref{sub:classes:resources} presents two datasets that are used as gold standard clustering in the experiments. Section~\ref{sub:classes:graph} presents the input graphs that are clustered using our approach to induce semantic structure. Finally, in Section~\ref{sub:classes:evaluation} results of the experiments are presented and discussed comparing them to the baseline clustering algorithms.

\subsection{\label{sub:classes:resources}Semantic Classes in Lexical Semantic Resources}

A \textit{semantic class} is a set of words that share the same semantic feature~\citep{Kozareva:08}. Depending on the definition of the notion of the \textit{semantic feature} the granularity and sizes of semantic classes may vary greatly. Examples of concrete semantic classes include sets of animals (dog, cat, \dots), vehicles (car, motorcycle, \dots), and fruit trees (apple tree, peach tree, \dots). In this experiment, we use a gold standard derived from a reference lexicographical database, namely WordNet~\citep{Fellbaum:98}. This allows us to benchmark the ability of {\watset} to reconstruct the semantic lexicon of such a reliable reference resource that has been widely used in NLP for many decades.

\subsubsection{WordNet Supersenses}

\begin{figure}[t]
  \centering
  \includegraphics[width=\textwidth]{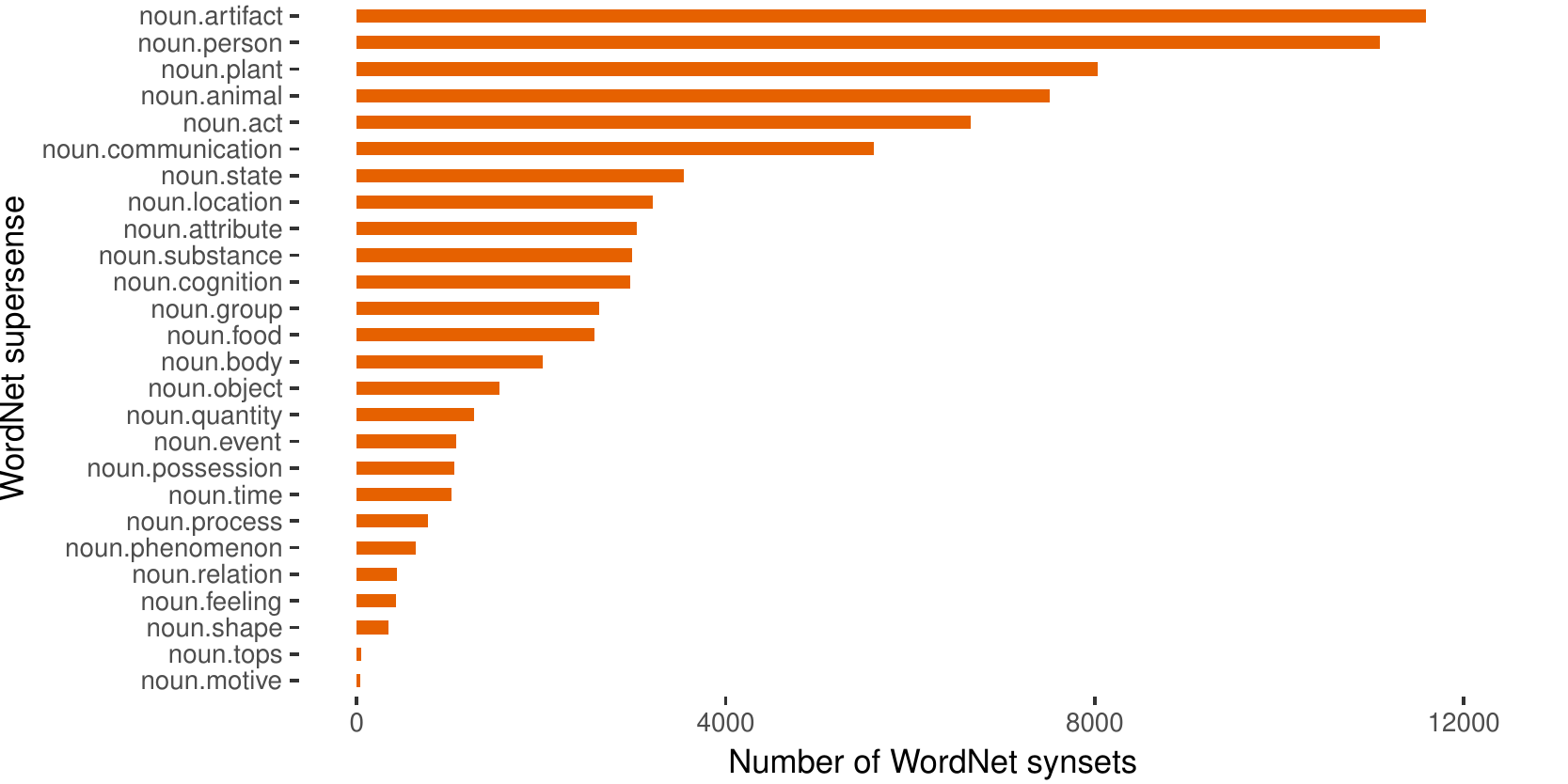} 
  \caption{\label{fig:classes:super}A summary of the noun semantic classes in WordNet supersenses~\citep{Ciaramita:03}.}
\end{figure}

The first dataset used in our experiments consists of 26 broad semantic classes, also known as \textit{supersenses} in the literature \citep{Ciaramita:03}: \textit{person}, \textit{communication}, \textit{artifact}, \textit{act}, \textit{group}, \textit{food}, \textit{cognition}, \textit{possession}, \textit{location}, \textit{substance}, \textit{state}, \textit{time}, \textit{attribute}, \textit{object}, \textit{process}, \textit{process}, \textit{tops}, \textit{phenomenon}, \textit{event}, \textit{quantity}, \textit{motive}, \textit{animal}, \textit{body}, \textit{feeling}, \textit{shape}, \textit{plant}, and \textit{relation}.

This system of broad semantic categories was used by lexicographers who originally constructed WordNet to thematically order the synsets; \figurename~\ref{fig:classes:super} shows the distribution of the 82,115 noun synsets from WordNet~3.1 across the supersenses. In our experiments in this section, these classes are used as gold standard clustering of word senses as recorded in WordNet. One can observe a Zipfian-like power-law~\citep{Zipf:49} distribution with a few clusters, such as \textsf{artifact} and \textsf{person} accounting for a large fraction of all nouns in the resource. Overall, in this experiment we decided to focus on  nouns as the input distributional thesauri used in this experiment (as presented in  Section~\ref{sub:classes:graph}) are most studied for modelling of noun semantics~\citep{Panchenko:16:nsi}.

The WordNet supersenses were applied later also for word sense disambiguation as a system of broad sense labels~\citep{Flekova:16}. For BabelNet, there is a similar dataset called BabelDomains~\citep{CamachoCollados:17} produced by automatically labeling BabelNet synsets with 32 different domains based on the topics of Wikipedia featured articles. Despite the larger size, however, BabelDomains provides only a silver standard (being semi-automatically created). We thus opt in the following to use WordNet supersenses only, since they provide instead a gold standard created by human experts.

\subsubsection{Flat Cuts of the WordNet Taxonomy}

\begin{figure}[t]
  \centering
  \includegraphics[scale=.85]{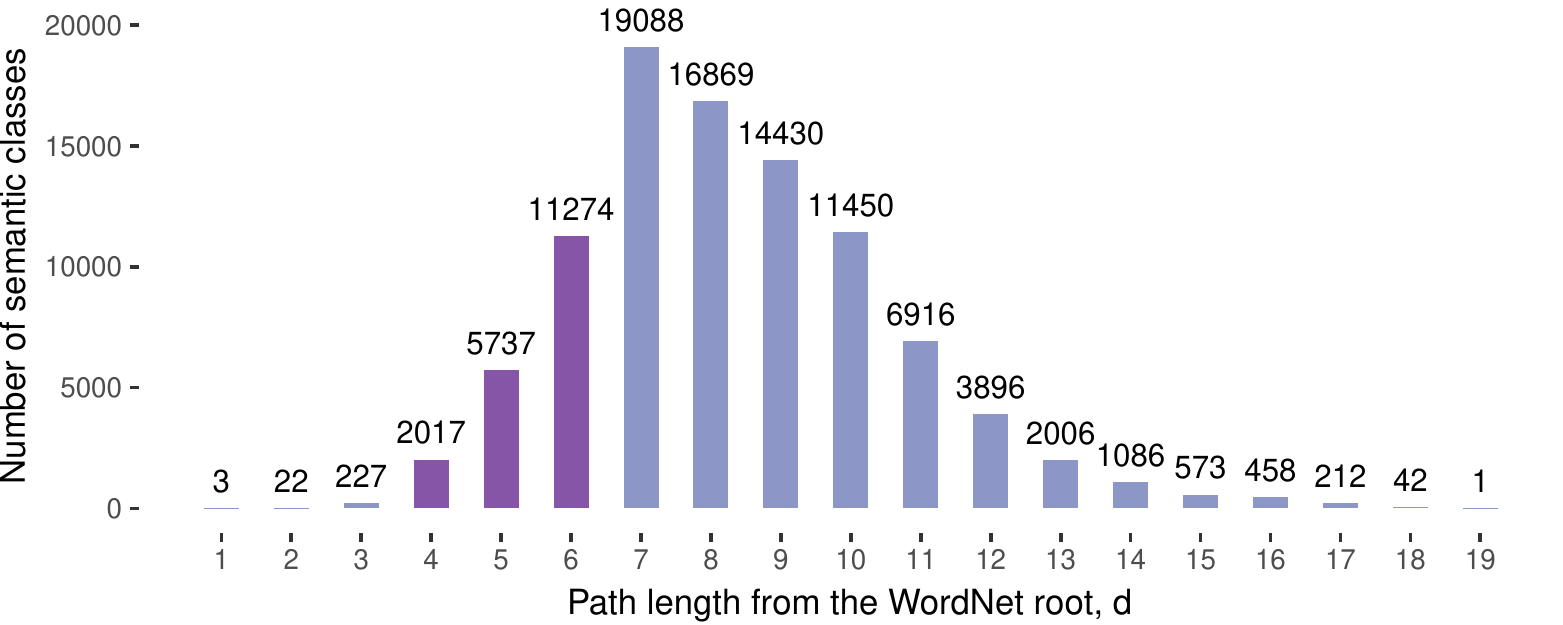} 
  \caption{\label{fig:classes:slices}Relationship between the number of semantic classes and path length from the WordNet~\citep{Fellbaum:98} root. We have chosen $d \in \{4, 5, 6\}$ for our experiments.}
\end{figure}

The second type of semantic classes used in our study are more semantically-specific and defined as subtrees of WordNet at some fixed path length of $d$ steps from the root node. We used the following procedure to gather these semantic classes.

First, we find a set of synsets that are located a exactly distance of $d$ edges from the root node. Each such a starting node, e.g., the synset \textsf{plant\_material.n.01}, identifies one semantic class. This starting node and all its descendants, e.g., \textsf{cork.n.01}, \textsf{coca.n.03}, \textsf{ethyl\_alcohol.n.1}, \textsf{methylated\_spirit.n.01}, and so on, in the case of the \textit{plant material} example, are included into the semantic class. Finally, we remove semantic classes that contain only one element as our goal is to create a gold standard dataset for clustering. \figurename~\ref{fig:classes:slices} illustrates distribution of the number of semantic classes as a function of the path length from the root. As one may observe, the largest number of clusters is obtained for the path length $d$ of 7. In our experiments, we use three versions of these WordNet ``taxonomy cuts'' which correspond to $d \in \{4, 5, 6\}$, since the cluster sizes generated at these levels are already substantially larger than those from the supersense dataset while providing a complementary evaluation at different levels of granularities. Although at some levels, such as $d=2$, the number of semantic classes is similar to the number of supersenses~\citep{Ciaramita:03}, there is no one-to-one relationship between them. As \citet{Richardson:94} points out, this cut-based derivative resource might bias towards the concepts belonging to shallow hierarchies: the node for ``horse'' is 10 levels from the root, while the node for ``cow'' is 13 levels deep. However, we believe that it adds an additional perspective to our evaluation while keeping the interpretability at the same time. Examples of the extracted semantic classes are presented in \tablename~\ref{tab:classes:synsets}.

\begin{table}[t]
\centering
\caption{\label{tab:classes:synsets}Examples of semantic classes extracted from WordNet hierarchy of synsets for the path length $d=5$ from the root synset.}
\begin{tabular}{p{20mm}p{105mm}}\toprule
\textbf{Root Synset} & \textbf{Child Synsets}  \\\midrule
rock.n.02 & aphanite.n.01, caliche.n.02, claystone.n.01, dolomite.n.01, emery\_stone.n.01, fieldstone.n.01, gravel.n.01, ballast.n.02, bank\_gravel.n.01, shingle.n.02, greisen.n.01, igneous\_rock.n.01, adesite.n.01, andesite.n.01, \mbox{\dots\textit{63 more entries}\dots}, tufa.n.01 \\\midrule
toxin.n.01 & animal\_toxin.n.01, venom.n.01, kokoi\_venom.n.01, snake\_venom.n.01, anatoxin.n.01, botulin.n.01,  cytotoxin.n.01, enterotoxin.n.01, nephrotoxin.n.01, endotoxin.n.01, exotoxin.n.01, \mbox{\dots\textit{19 more entries}\dots}, ricin.n.01 \\\midrule
axis.n.01 & coordinate\_axis.n.01, x-axis.n.01, y-axis.n.01, z-axis.n.01, major\_axis.n.01, minor\_axis.n.01, optic\_axis.n.01, principal\_axis.n.01, semimajor\_axis.n.01, semiminor\_axis.n.01 \\\bottomrule
\end{tabular}
\end{table}

\subsection{\label{sub:classes:graph}Construction of a Distributional Thesaurus}

A distributional thesaurus~\citep{Lin:98} is an undirected graph of semantically related words, with edges such as $\{\text{Python}, \text{Perl}\}$. We base our approach on the distributional hypothesis~\citep{Firth:57,Turney:10,Clark:15} to generate graphs of semantically related words for this experiment. The graphs represent $k$ nearest neighbouring of words that are semantically related to each other in a vector space. More specifically, the dimensions of the vector space represent salient syntactic dependencies of each word extracted using a dependency parser. For this, we use the JoBimText framework for computation of count-based distributional models from raw text collections~\citep{Biemann:13}.\footnote{\url{http://www.jobimtext.org}} While similar graphs could be derived also from neural distributional models, such as Word2Vec~\citep{Mikolov:13}, it was shown in~\citet{Riedl:16,Riedl:17} that the quality of syntactically-based graphs is generally superior.

\begin{figure}[t]
  \centering
  \includegraphics[scale=.5]{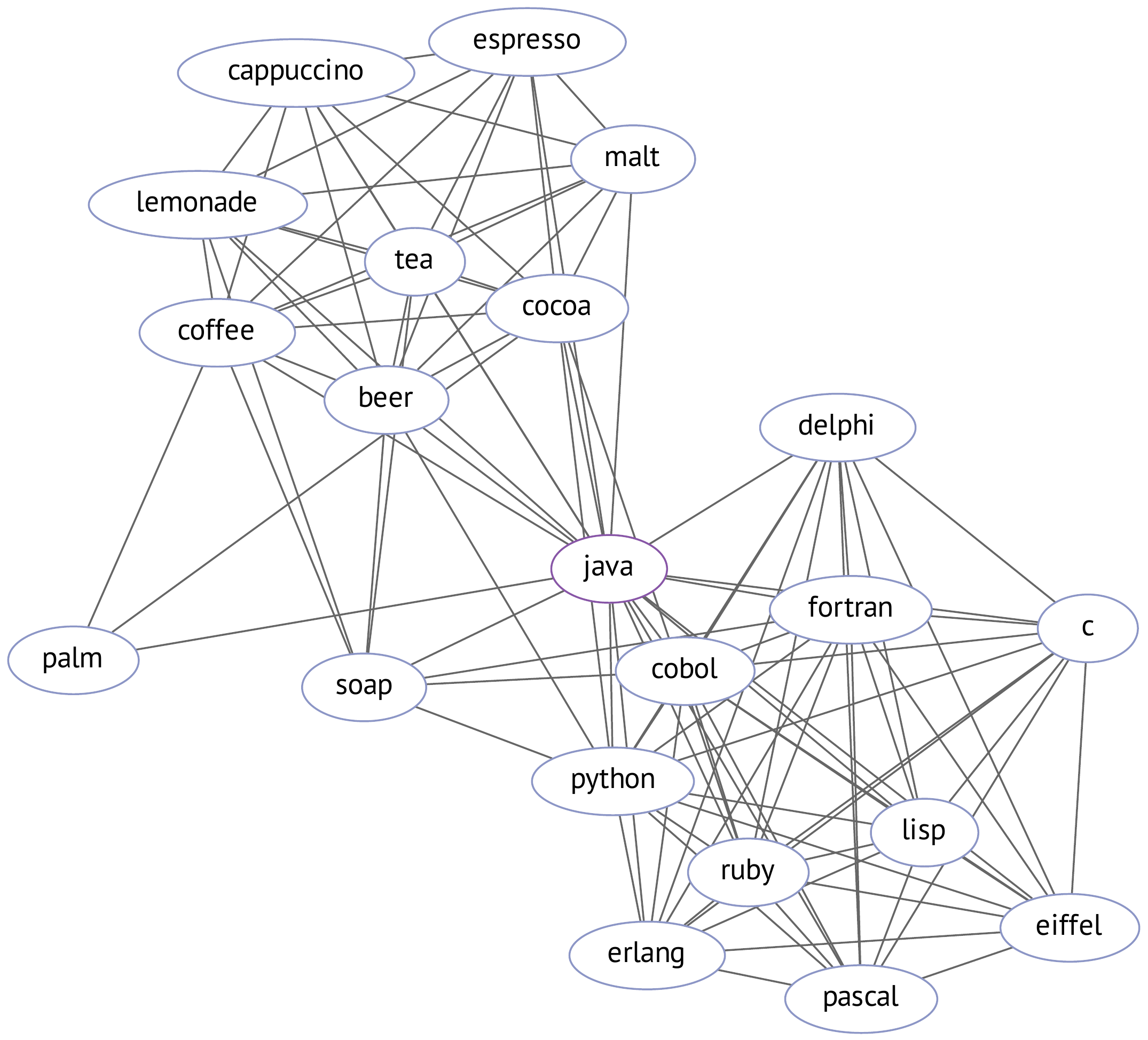}
  \caption{\label{fig:classes:ambigous}An example of the lexical unit ``\textcolor{ggplotsim}{java}'' and a part of its neighborhood in a distributional thesaurus. This polysemous word is not disambiguated, so it acts as a hub between two different senses.}
\end{figure}

The JoBimText framework involves several steps. First, it takes an unlabeled input text corpus and performs dependency parsing so as to extract features representing each word. Each word  is represented by a bag of syntactic dependencies such as $\mathrm{conj\_and}(\text{Ruby}, \cdot)$ or  $\mathrm{prep\_in}(\text{code}, \cdot)$, extracted from the dependencies of MaltParser~\citep{Nivre:06} which are further collapsed using the tool by \citet{Ruppert:15} in the notation of Stanford Dependencies~\citep{deMarneffe:06}.

Next, semantically related words are computed for each word in the input corpus. Features of each word are weighted and ranked using the Local Mutual Information (LMI) measure~\citep{Evert:05}. Subsequently, these word representations are pruned keeping 1000 most salient features per word ($\mathrm{fpw}$) and 1000 most salient words per feature ($\mathrm{wpf}$), where $\mathrm{fpw}$ and $\mathrm{wpf}$ are the parameters specific to the JoBimText framework. The pruning reduces computational complexity and noise. Finally, word similarities are computed as the number of common features for two words. This is, again, followed by a pruning step in which for every word, only the $k$ of 200 most similar terms are kept. The ensemble of all of these words is the distributional thesaurus, which is used in the following experiments. Note that, each word in such a thesaurus (i.e., a graph of semantically related words) is potentially ambiguous.

The last stage of the JoBimText approach performs induction of senses, however, here we do not use output of this stage, but instead apply the {\watset} algorithm to the distributional thesaurus with ambiguous word entries. The process of computation of a distributional thesaurus using the JoBimText framework is described in greater detail in \citet[Section~4.1]{Biemann:18}.

As an input corpus, we use a text collection of about 9.3 billion tokens that consists of a concatenation of Wikipedia,\footnote{\url{https://doi.org/10.5281/zenodo.229904}} ukWaC~\citep{Ferraresi:08}, Gigaword~\citep{Graff:03}, and LCC~\citep{Richter:06} corpora. Given the large size of these corpora, the graphs are built using an implementation of the JoBimText framework in Apache Spark,\footnote{\url{https://spark.apache.org}} which enables efficient distributed computation of large text collection on a distributed computational cluster.\footnote{\url{https://github.com/uhh-lt/josimtext}}

\figurename~\ref{fig:classes:ambigous} shows an example from the obtained distributional thesaurus. As in the experiments described in Sections~\ref{sec:synsets} and \ref{sec:triframes}, we assume that polysemous nodes serve as hubs that connect different unrelated clusters.

\subsection{\label{sub:classes:evaluation}Evaluation}

\begin{figure}[t]
  \centering
  \includegraphics[width=\textwidth]{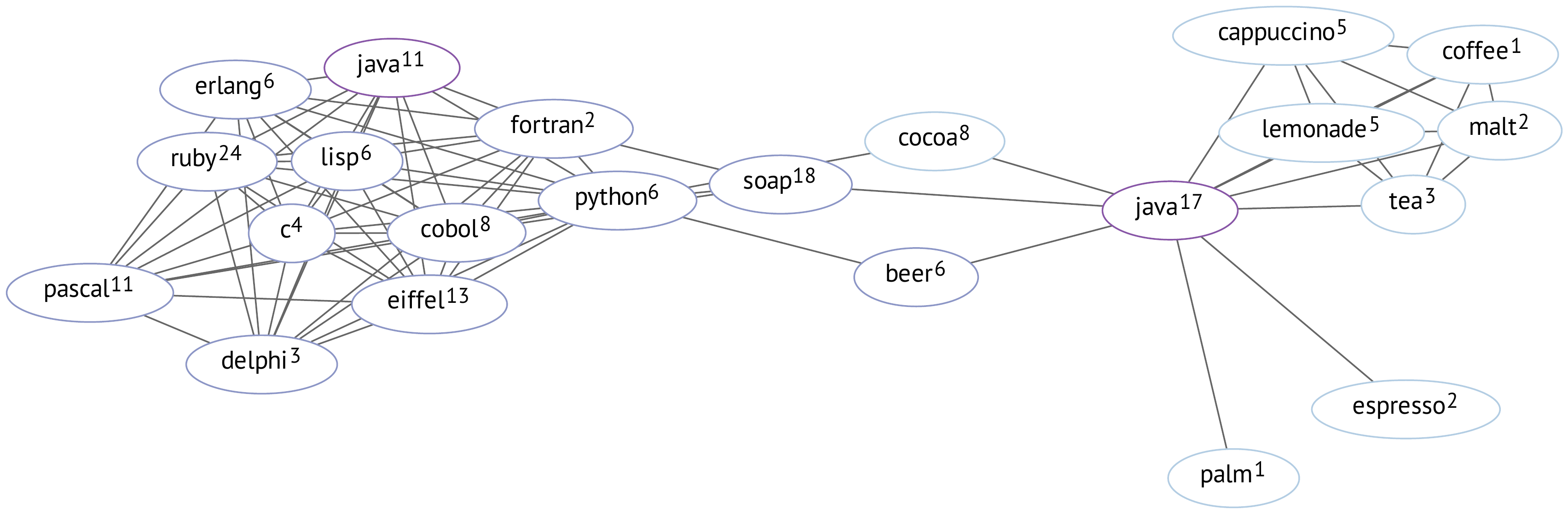}
  \caption{\label{fig:classes:disambiguated}An example of the \textit{sense graph} built by {\watset} for two senses of the lexical unit ``java'' using CW\textsubscript{log} for local clustering. In contrast to \figurename~\ref{fig:classes:ambigous}, in this \textit{disambiguated} distributional thesaurus the node corresponding to the lexical unit ``java'' is split: \sense{java}{11} is connected to \textcolor{ggplotcount}{programming languages} and \sense{java}{17} is connected to \textcolor{ggplotones}{drinks}.}
\end{figure}

We cast the semantic class induction problem as a task of clustering distributionally related graphs of words and word senses, which is conceptually similar to our synset induction task in Section~\ref{sec:synsets}. \figurename~\ref{fig:classes:disambiguated} shows an example of the sense graph (Section~\ref{sub:global}) built by {\watset} before running a global clustering algorithm that induces the sense-aware semantic classes based on a distributional thesaurus example in \figurename~\ref{fig:classes:ambigous}.

\subsubsection{Experimental Setup} Similarly to our synset induction experiment (Section~\ref{sub:synsets:setup}), we study the performance of clustering algorithms by comparing the clustering of the same input distributional thesaurus to a gold standard clustering. We used the same implementations and algorithms as all other experiments reported in this paper, such as Markov Clustering (MCL) by \citet{vanDongen:00}, Chinese Whispers (CW) by \citet{Biemann:06}, and MaxMax~\citep{Hope:13:maxmax}. We did not evaluate such algorithms as CPM~\citep{Palla:05} and ECO~\citep{GoncaloOliveira:14} due to their poor performance shown on the synset induction task.

\begin{table}[t]
\centering
\caption{\label{tab:classes:input}Properties of the input datasets used in the semantic class induction experiment compared to the original distributional thesaurus (DT) by \citet{Biemann:13}.}
\begin{tabular}{p{90mm}rr}\toprule
\textbf{DT Pruning Method} & \textbf{\# of nodes} & \textbf{\# of edges} \\\midrule
Unpruned \citep{Biemann:13} & $4{,}430{,}170$ & $595{,}916{,}414$ \\
Supersenses~(Ciaramita~\citeyear{Ciaramita:03}) & $37{,}937$ & $6{,}944{,}731$ \\
Path Length of $d=4$ & $33{,}213$ & $5{,}841{,}359$ \\
Path Length of $d=5$ & $32{,}048$ & $5{,}478{,}110$ \\
Path Length of $d=6$ & $29{,}515$ & $4{,}814{,}132$ \\\bottomrule
\end{tabular}
\end{table}

\paragraph{Input Data} We use the distributional thesaurus as described in Section~\ref{sub:classes:graph}. Since the original distributional thesaurus graph has approximately 600 million edges, we pruned it by removing all the edges having the minimal weight, i.e., $0.001$ in our case. Also, due to the difference in lexicons between the gold standards and the input graph, we performed additional pruning by removing all the edges connecting words missing the gold standard lexicons. As the result, we obtained four different pruned input graphs (\tablename~\ref{tab:classes:input}). We performed no parameter tuning in this experiment, so we report the best-performing configuration of each method among other ones.

\paragraph{Gold Standard} We use two different kinds of semantic classes for evaluation purposes. Both of the used semantic class types are based on the WordNet lexical database \citep{Fellbaum:98} yet they have widely different granularities. First, we use the WordNet supersenses dataset by \citet{Ciaramita:03}. Second, we use our path-based gold standards of lengths 4, 5 and 6 as described in Section~\ref{sub:classes:resources}.

\paragraph{Quality Measure} In the synset induction experiment (Section~\ref{sub:synsets:setup}) we use the pairwise F\textsubscript{1}-score \citep{Manandhar:10} as the performance indicator. However, since the average size of a cluster in this experiment is much higher (\tablename~\ref{tab:classes:input} and \figurename~\ref{fig:classes:super}), we found that the enumeration of $2$-combinations of semantic class elements is not computationally tractable in reasonable time on relatively large datasets like the ones we use in this experiment. For example, a cluster of $10{,}000$ elements needs to be transformed into a sufficiently large set of $\frac{1}{2} \times 10^5\times(10^5-1) \approx 5 \times 10^9$ pairs, which is inconvenient for processing. Therefore, we used the same quality measure as in our unsupervised lexical semantic frame induction experiment (Section~\ref{sub:triframes:setup}), namely normalized modified purity ($\nmpu$) and normalized inverse purity ($\nipu$) as defined by \citet{Kawahara:14}.

\paragraph{Statistical Testing} Since the chosen quality measure does not allow the computation of a contingency table, we use exactly the same procedure for statistical testing as in the experiment on lexical semantic frame induction (Section~\ref{sub:triframes:setup}). Due to a high computational complexity of the bootstrapping statistical testing procedure~\citep{Dror:18}, we limited the number of samples $N$ to $5000$ in this experiment.

\subsubsection{Results and Discussion}

\paragraph{Comparison to Baselines}
\tablename~\ref{tab:classes:supersenses} shows the evaluation results on the WordNet supersenses dataset. We found that our approach, \watset[CW\textsubscript{lin}, CW\textsubscript{log}], shows statistically significantly better results with respect to F\textsubscript{1}-score ($p \ll 0.01$) than all the methods apart from Simplified {\watset} in the same configuration. The experimental results in \tablename~\ref{tab:classes:d456} obtained on different variations of our WordNet-based gold standard as described in Section~\ref{sub:classes:resources} confirm a high performance of {\watset} on all the evaluation datasets. Thus, results of experiments on these four types of semantic classes of greatly variable granularity (from 26 classes for the supersenses to 11,274 classes for the flat cut with $d=6$) lead to similar conclusions about the advantage of the {\watset} approach as compared to the baseline clustering algorithms.

\begin{table}[t]
\centering
\caption{\label{tab:classes:supersenses}Comparison of the graph clustering methods against the WordNet supersenses dataset by \citet{Ciaramita:03}; best configurations of each method in terms of F\textsubscript{1}-scores are shown. Results are sorted by F\textsubscript{1}-score, top values of each measure are boldfaced and statistically significant results are marked with an asterisk ($^\ast$). Simplified {\watset} is denoted as {\watset\S}.}
\begin{tabular}{p{70mm}cccc}\toprule
\textbf{Method} & \textbf{\# clusters} & \textbf{nmPU} & \textbf{niPU} & \textbf{F\textsubscript{1}} \\\midrule
\watset[CW\textsubscript{lin}, CW\textsubscript{log}] & $47{,}054$ & $57.20$ & $40.52$ & $\mathbf{47.44}\hphantom{^\ast}$ \\
\watset\S[CW\textsubscript{lin}, CW\textsubscript{log}] & $47{,}797$ & $58.16$ & $39.86$ & $47.30^\ast$ \\
CW\textsubscript{log} & $108$ & $35.03$ & $\mathbf{46.17}$ & $39.84^\ast$ \\
MCL & $368$ & $61.34$ & $15.31$ & $24.50^\ast$ \\
MaxMax & $4050$ & $\mathbf{68.48}$ & $4.15$ & $7.82\hphantom{^\ast}$ \\\bottomrule
\end{tabular}
\end{table}

\begin{table}[t]
\centering
\caption{\label{tab:classes:d456}Evaluation results on path-limited versions of WordNet by $4$, $5$ and $6$; best configurations of each method in terms of F\textsubscript{1}-scores are shown. Results are sorted by F\textsubscript{1}-score on the $d=6$ WordNet slice, top values of each measure are boldfaced. Simplified {\watset} is denoted as {\watset\S}.}
\resizebox{1.0\linewidth}{!}{
\begin{tabular}{l*{3}{|ccc}}\toprule
\multirow{2}{*}{\textbf{Method}} & \multicolumn{3}{c}{$\mathbf{d=4}$} & \multicolumn{3}{|c}{$\mathbf{d=5}$} & \multicolumn{3}{|c}{$\mathbf{d=6}$} \\\cmidrule{2-10}
& \textbf{nmPU} & \textbf{niPU} & \textbf{F\textsubscript{1}} & \textbf{nmPU} & \textbf{niPU} & \textbf{F\textsubscript{1}} & \textbf{nmPU} & \textbf{niPU} & \textbf{F\textsubscript{1}} \\\midrule
\watset\S[CW\textsubscript{lin}, CW\textsubscript{top}] &
$47.43$ & $42.63$ & $\mathbf{44.90}$ &
$45.26$ & $42.67$ & $\mathbf{43.93}$ &
$40.20$ & $44.37$ & $\mathbf{42.18}$ \\
\watset[CW\textsubscript{lin}, CW\textsubscript{top}] &
$47.38$ & $42.65$ & $44.89$ &
$44.86$ & $43.03$ & $\mathbf{43.93}$ &
$40.07$ & $44.14$ & $42.01$ \\
CW\textsubscript{lin} &
$34.09$ & $40.98$ & $37.22$ &
$34.92$ & $40.65$ & $37.57$ &
$31.84$ & $41.89$ & $36.18$ \\
CW\textsubscript{log} &
$29.00$ & $\mathbf{44.85}$ & $35.23$ &
$29.63$ & $\mathbf{44.72}$ & $35.64$ &
$26.00$ & $\mathbf{46.36}$ & $33.31$ \\
MCL &
$54.90$ & $19.63$ & $28.92$ &
$45.32$ & $22.59$ & $30.15$ &
$38.38$ & $26.96$ & $31.67$ \\
MaxMax &
$\mathbf{59.29}$ & $6.93$ & $12.42$ &
$\mathbf{52.65}$ & $10.14$ & $17.01$ &
$47.28$ & $13.69$ & $21.23$ \\\bottomrule
\end{tabular}
}
\end{table}

\tablename~\ref{tab:classes:sample} shows examples of the obtained semantic classes of various sizes for the best {\watset} configuration on the WordNet supersenses dataset. During error analysis we found two primary causes of errors: incorrectly identified edges and overly specific sense contexts.

Since we performed only a minimal pruning of the input distributional thesaurus, this contains many edges with low weights that typically represent mistakenly recognized relationships between words. Such edges, when appearing between two disjoint meaningful clusters, act as hubs, which {\watset} puts in both clusters. For example, a sense graph in \figurename~\ref{fig:classes:disambiguated} has a node \sense{soap}{18} incorrectly connected to a drinks-related node \sense{java}{17} instead of the node \sense{java}{11} that is more related to programming languages.\footnote{Strictly speaking, SOAP (Simple Object Access Protocol) is not a programming language, so the presence of this node in the graphs demonstrated in Figures~\ref{fig:classes:ambigous} and \ref{fig:classes:disambiguated} is a mistake.} Reliable distinction between ``legitimate'' polysemous nodes and incorrectly placed hubs is a direction for future work.

The node sense induction approach of {\watset}, as described in Section~\ref{sub:wsi}, takes into account only the neighborhood of the target node which is a first-order ego network~\citep{Everett:05}. As we observe throughout all the experiments in this article, {\watset} tends to produce more fine-grained senses than one might expect. These fine-grained senses, in turn, lead to the global clustering algorithm to include incoherent nodes to clusters as in \tablename~\ref{tab:classes:sample}. We believe that taking into account additional features, such as second-order ego networks, to induce coarse-grained senses could potentially improve the overall performance of our algorithm (at a higher computational cost).

We found a generally poor performance of MCL in this experiment due to its tendency to produce fine-grained clusters by isolating hubs from their neighborhoods. Although this behavior improved the results on the synset induction task (Section~\ref{sub:watset:results}), our distributional thesaurus is a more complex resource as it expresses semantic relationships other than synonymity, so the incorrectly identified edges affect MCL as well as {\watset}.

\begin{table}[t]
\centering
\caption{\label{tab:classes:sample}Sample semantic classes induced by the \watset[CW\textsubscript{lin}, CW\textsubscript{log}] method as according to the WordNet supersenses dataset by \citet{Ciaramita:03}.}
\begin{tabular}{cp{120mm}}\toprule
\textbf{Size} & \textbf{Semantic Class}\\\midrule
7 & dye, switch-hitter, dimaggio, hitter, gwynn, three-hitter, muser \\\midrule
13 & worm, octopus, pike, anguillidae, congridae, conger, anguilliformes, eel, marine, grouper, muraenidae, moray, elver \\\midrule
16 & gothic, excelsior, roman, microgramma, stymie, dingbat, italic, century, trajan, outline, twentieth, bodoni, serif, lydian, headline, goudy \\\midrule
20 & nickel, steel, alloy, chrome, titanium, cent, farthing, cobalt, brass, denomination, fineness, paisa, copperware, dime, cupronickel, centavo, avo, threepence, coin, centime \\\midrule
23 & prochlorperazine, nicotine, tadalafil, billionth, ricin, pravastatin, multivitamin, milligram, anticoagulation, carcinogen, microgram, niacin, l-dopa, lowering, arsenic, morp
hine, nevirapine, caffeine, ritonavir, aspirin, neostigmine, rem, milliwatt \\\midrule
54 & integer, calculus, theta, pyx, curvature, saturation, predicate, \mbox{\dots\textit{40 more words}\dots}, viscosity, brightness, variance, lattice, polynomial, rho, determinant \\\midrule
369 & electronics, siren, dinky, banjo, luo, shawm, shaker, helicon, rhodes, conducting, \mbox{\dots\textit{349 more words}\dots}, narrator, paradiddle, clavichord, chord, consonance,sextet, zither, cantor, viscera, axiom \\\midrule
1093 & egg, pinworm, forager, decidua, psittacus, chimera, coursing, silkworm, spirochete, radicle, \mbox{\dots\textit{1073 more words}\dots}, earthworm, annelida, integument, pisum, biter, wilt, heartwood, shellfish, swarm, cryptomonad \\\bottomrule
\end{tabular}
\end{table}

\paragraph{Impact of Distributional Thesaurus Pruning on Ambiguity} In order to study the effect of pruning, we performed another experiment on a DT that was pruned using a relatively high edge weight threshold of $0.01$, which is 10 times larger than the minimal threshold we used in the experiment described in Section~\ref{sub:classes:evaluation}. A manual inspection of the pruned graph showed that most, if not all, nodes were either monosemeous words or proper nouns, so hard clustering algorithms should have an advantage in this scenario. \tablename~\ref{tab:classes:supersenses:pruned} confirms that in this setup soft clustering algorithms, such as {\watset} and MaxMax, are clearly outperformed by hard clustering algorithms that are more suitable for processing monosemous word graphs. Since our algorithm explicitly performs node sense induction to produce fine-grained clusters, we found that an average semantic class produced by \watset[CW\textsubscript{top}, CW\textsubscript{top}] has $10.77 \pm 187.37$ words, while CW\textsubscript{log} produced semantic classes of $133.46 \pm 1317.97$ words in average.

\begin{table}[t]
\centering
\caption{\label{tab:classes:supersenses:pruned}Comparison of the graph clustering methods on the pruned DT with an edge threshold of $0.01$ against the WordNet supersenses dataset by \citet{Ciaramita:03}; best configurations of each method in terms of F\textsubscript{1}-scores are shown. Results are sorted by F\textsubscript{1}-score, top values of each measure are boldfaced. Simplified {\watset} is denoted as {\watset\S}.}
\begin{tabular}{p{70mm}rrrr}\toprule
\textbf{Method} & \textbf{\# clusters} & \textbf{nmPU} & \textbf{niPU} & \textbf{F\textsubscript{1}} \\\midrule
CW\textsubscript{log} & $183$ & $39.72$ & $\mathbf{28.46}$ & $\mathbf{33.16}$ \\
\watset\S[CW\textsubscript{top}, CW\textsubscript{top}] & $3944$ & $57.22$ & $20.21$ & $29.87$ \\
\watset[CW\textsubscript{top}, CW\textsubscript{top}] & $3954$ & $57.38$ & $19.91$ & $29.56$ \\
MCL & $526$ & $65.12$ & $8.46$ & $14.98$ \\
MaxMax & $3671$ & $\mathbf{72.17}$ & $2.00$ & $3.88$ \\\bottomrule
\end{tabular}
\end{table}

To summarize, in contrast to synonymy dictionaries, whose completeness and availability are limited (Section~\ref{sub:watset:results}), a distributional thesaurus can be constructed for any language provided with a relatively large text corpus. However, we found that they need to be carefully pruned to reduce the error rate of clustering algorithms~\citep{Panchenko:18:mangosteen}.

\section{\label{sec:conclusion}Conclusion}

In this article, we presented {\watset}, a generic meta-algorithm for fuzzy graph clustering. This algorithm creates an intermediate representation of the input graph that naturally reflects the ``ambiguity'' of its nodes. Then, it uses hard clustering to discover clusters in this ``disambiguated'' intermediate graph. This enables straightforward semantic-aware grouping of relevant objects together. We refer to {\watset} as a meta-algorithm because it does not perform graph clustering \textit{per se}. Instead, it encapsulates the existing clustering algorithms and builds a sense-aware representation of the input graph that we call a \textit{sense graph}. Although we use the sense graph in this article exclusively for clustering, we believe that it can be useful for more applications.

The experiments show that our algorithm performs fuzzy graph clustering with a high accuracy. This is empirically confirmed by successfully applying {\watset} to complex language processing, such as tasks as unsupervised induction of synsets from a synonymy graph, semantic frames from dependency triples, as well as semantic class induction from a distributional thesaurus. In all cases, the algorithm successfully handled the ambiguity of underlying linguistic objects, yielding the state-of-the-art results in the respective tasks. {\watset} is computationally tractable and its local steps can easily be run in parallel.

As future work we plan to apply {\watset} to other types of linguistic networks to address more natural language processing tasks, such as taxonomy induction based on networks of noisy hypernyms extracted from text~\citep{Panchenko:16:semeval}. Besides, an interesting future challenge is the development of a scalable graph clustering algorithm that can natively run in a parallel distributed manner, e.g., on a large distributed computational cluster. The currently available algorithms, such as MCL~\citep{vanDongen:00} and CW~\citep{Biemann:06}, cannot be trivially implemented in such a fully distributed environment, limiting the scale of language graph they can be applied to. Another direction of future work is using {\watset} in downstream applications. We believe that our algorithm can successfully detect structure in a wide range of different linguistic and non-linguistic datasets, which can help in processing out-of-vocabulary items or resource-poor languages or domains without explicit supervision.

\subsubsection*{\label{sub:implementation}Implementation} We offer an efficient open source multi-threaded implementation of {\watset} (Algorithm~\ref{alg:watset}) in the Java programming language.\footnote{\url{https://github.com/nlpub/watset-java}} It uses a thread pool to simultaneously perform \textit{local} steps, such as node sense induction (lines~\ref{alg:watset:wsi:begin}--\ref{alg:watset:wsi:end}, one word per thread) and context disambiguation (lines~\ref{alg:watset:wsd:begin}--\ref{alg:watset:wsd:end}, one sense per thread). Our implementation includes Simplified {\watset} (Algorithm~\ref{alg:watset:simplified}) and also features both a command-line interface and an application programming interface for integration into other graph and language processing pipelines in a generic way. Additionally, we bundle with it our own implementations of Markov Clustering~\citep{vanDongen:00}, Chinese Whispers~\citep{Biemann:06}, and MaxMax~\citep{Hope:13:maxmax} algorithms. Also, we offer an implementation of the Triframes frame induction approach\footnote{\url{https://github.com/uhh-lt/triframes}} and an implementation of the semantic class induction approach.\footnote{\url{https://github.com/umanlp/watset-classes}} The datasets produced during this study are available on Zenodo.\footnote{\url{https://doi.org/10.5281/zenodo.2621579}}

\begin{acknowledgments}
We acknowledge the support of the Deutsche Forschungsgemeinschaft (DFG) foundation under the ``JOIN-T'' and ``ACQuA'' projects, the Deutscher Akademischer Austauschdienst (DAAD), and the Russian Foundation for Basic Research (RFBR) under the project no.~16-37-00354~\foreignlanguage{russian}{мол\_а}. We also thank Andrew Krizhanovsky for providing a parsed Wiktionary, Natalia Loukachevitch for the provided RuWordNet dataset, Mikhail Chernoskutov for early discussions on computational complexity of {\watset}, and Denis Shirgin who actually suggested the {\watset} name. Furthermore, we thank Dmitry Egurnov, Dmitry Ignatov, and Dmitry Gnatyshak for help in operating the NOAC method using the multimodal clustering toolbox. Besides, we are grateful to Ryan Cotterell and Adam Poliak for a discussion and an implementation of the High-Order Skip Gram (HOSG) method. We thank Bonaventura Coppolla for discussions and preliminary work on graph-based frame induction and Andrei Kutuzov, who conducted experiments with the HOSG-based baseline related to the frame induction experiment. We thank Stefano Faralli for early work on graph-based sense disambiguation. We thank Rotem Dror for discussion of the theoretical background underpinning the statistical testing approach that we use in this paper. We are grateful to Federico Nanni and Gregor Wiedemann for proofreading this paper. Finally, we thank three anonymous reviewers for insightful comments on the present article.
\end{acknowledgments}

\clearpage
\starttwocolumn
\bibliography{watset.cl}

\end{document}